\documentclass[conference]{IEEEtran}
\IEEEoverridecommandlockouts
\usepackage{cite}
\usepackage{amsmath,amssymb,amsfonts}
\usepackage{algorithmic}
\usepackage{graphicx}
\usepackage{textcomp}
\usepackage{xcolor}
\def\BibTeX{{\rm B\kern-.05em{\sc i\kern-.025em b}\kern-.08em
    T\kern-.1667em\lower.7ex\hbox{E}\kern-.125emX}}

\usepackage{subfigure}
\graphicspath{ {Figures/} }
\begin{document}

\title{Variance Tolerance Factors For Interpreting All Neural Networks}

%\author{\IEEEauthorblockN{Anonymous Authors}}

\author{\IEEEauthorblockN{ Sichao Li}
\IEEEauthorblockA{\textit{School of Computing} \\
\textit{Australian National University}\\
Acton ACT 2601, Australia \\
Sichao.Li@anu.edu.au}
\and
\IEEEauthorblockN{Amanda Barnard}
\IEEEauthorblockA{\textit{School of Computing} \\
\textit{Australian National University}\\
Acton ACT 2601, Australia \\
Amanda.S.Barnard@anu.edu.au}}
\maketitle

\begin{abstract}
Black box models only provide results for deep learning tasks, and lack informative details about how these results were obtained. Knowing how input variables are related to outputs, in addition to why they are related, can be critical to translating predictions into laboratory experiments, or defending a model prediction under scrutiny. In this paper, we propose a general theory that defines a variance tolerance factor (VTF) inspired by influence function, to interpret features in the context of black box neural networks by ranking the importance of features, and construct a novel architecture consisting of a base model and feature model to explore the feature importance in a Rashomon set that contains all well-performing neural networks. Two feature importance ranking methods in the Rashomon set and a feature selection method based on the VTF are created and explored. A thorough evaluation on synthetic and benchmark datasets is provided, and the method is applied to two real world examples predicting the formation of noncrystalline gold nanoparticles and the chemical toxicity 1793 aromatic compounds exposed to a protozoan ciliate for 40 hours.
\end{abstract}


\section{Introduction}

Machine learning is a powerful tool that can learn and describe the patterns behind data, enabling researchers to make predictions and improve decisions \cite{hastie2009elements}. This advanced technology has been widely employed in  scientific disciplines, e.g. psychology, physics and genomics, but many problems are solved using box machine learning models and big datasets that are not representative \cite{rudin2019stop}. Knowing the prediction results or a single metric, such as classification accuracy, is insufficient for describing tasks \cite{doshi2017towards} and inappropriate for studies in the sciences where predictions must be accountable. For example, the effectiveness of a medical treatment for a patient may be predictable, but how and why the prediction was made is also an important part of the patient record. Interpretability and unambiguous explanations of the model can be crucial.

The definition of interpretability is hard to be established. Miller \cite{miller2019explanation} and Kim \cite{kim2016examples} described the interpretability ``the degree to which a human can comprehend the cause of a decision''. Humans can understand the behavior and predictions of machine learning by using transparent methods or interpretable methods. Feature importance rankings that measure the contribution of the input variables to the prediction hava become an important tool in interpretable machine learning \cite{samek2017explainable}. These ranking methods are often associated with feature selection, as features can be selected or eliminated based on the rankings to provide an optimal subset \cite{guyon2003introduction}.
Intuitively, if a model can ``tell'' which features are important to itself, the machine learning system would benefit from a subset of optimal features by reducing the dimensionality (improving efficiency), improving the generalizability of the learning system \cite{guyon2002gene, song2007supervised, wojtas2020feature} and simplifying accountability. It was argued that focusing solely on explaining a pre-specific black box model is unjustifiable according to recent research \cite{rudin2019stop, fisher2019all}, and it would be more appropriate to explore the feature importance in all well-performing models. These are the primary motivations of this study.

In this paper, we defined the variance tolerance factor (VTF) that evaluates model's tolerance to features' variance in a Rashomon set that contains accurate neural networks, and proposed a novel model-agnostic framework based on it to address the lack of interpretability in neural networks \cite{ibrahim2013comparison, olden2004accurate}, where the contribution of the input variables (features) in predicting the output (target) is opaque. Two model-agnostic methods are devised; one from data-driven perspective and one based on mathematical principles. Our architecture consists of a ``base model'' and a ``feature model'', where the base model can be constructed in any format based on the type of task (classification or regression). The two models are trained sequentially and the feature model is systematically re-trained until it achieves the same performance as the base task. After training, the base model is fixed and the weights of a mask layer in the feature model provides the general feature importance recognised by the model. 
These weights can be further processed as the basis for fast feature selection and interpretable ranking methods using domain knowledge.
Each method is thoroughly evaluated on synthetic, benchmark and real datasets via representative machine learning models (e.g. Linear Regression, Logistic Regression, Multilayer Perceptron, Convolutional Neural Network and Recurrent Neural Network) and the study demonstrates the flexibility and interpretability of our framework
\footnote{The Supporting Information with code can be found in https://github.com/Sichao-Li/Variance-Tolerance-Factors-for-Interpreting-All-Good-Neural-Networks}.

\section{Related Work}

Interpretable machine learning methods are normally categorized as model-specific interpretation methods and model-agnostic interpretation methods \cite{molnar2018guide}.
It's easy to achieve interpretability by using algorithms that create interpretable models, e.g. linear regression, logistic regression and decision trees. The great advantage of model-agnostic interpretation methods that separate the explanations from the machine learning mode is their flexibility \cite{liu2018improving}. Model-agnostic interpretation methods can be further categorized as global interpretation methods and local interpretation methods. 
Global methods describe the average behavior of a machine learning model, such as permutation feature importance \cite{ho1995random}, the partial dependence plot \cite{friedman2001greedy}, functional decomposition and global surrogate. In contrast, local interpretation methods explain individual predictions. Shapley values from game theory are an example of a method used to explain individual predictions \cite{lundberg2017unified}. These methods report models' interpretations through feature importance ranking. 
The importance of features are defined differently based on various algorithms: filter-based methods, e.g. SPEC \cite{gallagher2007diabetic} and Fisher Score \cite{gu2012generalized}, feature discriminative capability, e.g. ReliefF \cite{robnik2003theoretical} and statistics-based methods , e.g. T-Score \cite{davis1986statistics}. The wrapper-based methods depend on the accuracy of specific models and evaluate all the possible combinations of features against the evaluation criterion, which usually suffer the problem of high computation complexity \cite{tang2014data}. 
Embedded methods are model-specific and use their own built-in feature selection methods, similar to model-specific methods discussed above. Learning attention used to indicate the feature importance is proved successful in the past \cite{gui2019afs}. 
To obtain a more complete understanding of feature importance for a set of different models with equally accurate predictive power, defined as Rashomon set \cite{breiman2001statistical}, researchers started exploring the set where all models with loss below a threshold \cite{dong2020exploring, fisher2019all}. In this paper, our aim is to propose a new view of exploring the Rashomon set for neural networks and interpret the feature importance using information from the set. 

\section{Theorems and methods}

\subsubsection{Rashomon set}
Given a model class F, the black box model with the minimum loss $L(f^{*})$ and the pre-defined tolerance $ \epsilon > 0$, the Rashomon set is defined as:
\begin{equation}
R(\epsilon, f, F) = \{f \in F | L(f) \leq L(f^{*}) + \epsilon \}
\end{equation}

\subsubsection{Variance Tolerance Factor (VTF)} 
VTF measures the sensitivity of a model to changes in the features space by exploring the model class of the Rashomon set $\lim_{\epsilon\to0} R(\epsilon, f, F)$.
The idea behind the searching method is to explore the feature space by modifying the distributions, instead of model space like conventional methods, as the structure and parameters of black box remain unknown in this context. Details are discussed in section \ref{sec: md}.

By scaling all features simultaneously and quantifying the impact without influencing the model performance, we can find a Rashomon set without prior knowledge of the model information.
Then we defined VTF, inspired by the influence function, as the influence of scaling training feature $X_{i}$ on the parameters. 
\begin{equation}
t_{w_{X_{i}}} = \lim_{\epsilon\to0}\frac{\Delta w_{X_{i}, \epsilon}}{\epsilon}
\end{equation}
Where $t$ is the notation of VTF and $w_{X_{i}}$ is the scaling factor. 
The logic behind why influence function can be used in features is roughly reasoned here: researchers \cite{koh2019accuracy, basu2020second} have proved the success in application of influence function on a group of instances. When the group contains all instances in the dataset and perturbing all instances in the same way, the process can be seen as perturbing features, evidence provided in the Supporting Information.

Then we can conclude that if the model performance is unaffected by large scaling factors, the model can't tolerate much variance in such features and then these features are consequentially important. As the parameters are always changing from $||1||$, the VTF can be further simplified as $t_{w_{X_{i}}} = \left \| t_{w_{X_{i}}}  -1 \right \|$.
If the VTF tends to 0, the feature importance tends to 1; i.e. the model is less tolerant to change in this dimension. 

\subsubsection{Feature Selection}
Based on the VTF, unimportant features can be selected such that the deviation between the factor of a feature and 1 is greater than 1,  $t_{w_{X_{i}}} > 1$, and eliminated in linear time, $O(n)$. Specific tasks might apply different thresholds rather than 1.
\begin{equation}
 \begin{split}
& F_{all} = \{\text{All Features}\} \\
& F_{unimportant} = \forall xF_{all}(t_{w_{X_{i}}}>1)
\end{split}
\end{equation}

\subsubsection{Auxiliary mask} To find the VTF, an auxiliary mask $m\in \mathbb{R}^{d}$ is introduced, which enables the feature set $ \chi \in \mathbb{R}^{n\times d}$ to achieve the similar performance on the same model $Q$, where $ m = \{w_{X_{1}}, w_{X_{2}} ... w_{X_{i}}\}$ and $ \chi = \{X_{1}, X_{2} ... X_{i}\}$ is a dictionary of features. Each $Q$ can be seen as a neural network in $R(\epsilon, Q, F)$. Then the following performance equivalence equation can be obtained:
\begin{equation}
L_{Q(m \otimes \chi)} \leq L_{Q(\chi)} + \epsilon , Q \in \lim_{\epsilon\to0} R(\epsilon, Q, F)
\label{eqn:coefficient_equal}
\end{equation}
where $\otimes$ denotes the element-wise product.

\subsection{Problem Formulation}
The problem then becomes ranking the importance of features in $\chi$ through an auxiliary mask $m$. 
Suppose $\chi$ is a feature set of $d$ features and $y$ is the corresponding target set.
The performance equivalence equation Eqn.\ref{eqn:coefficient_equal} can be solved as an optimization problem. 
\begin{equation}
 \label{eqn:coefficient_transfer}
argmin_{m}L_{Q(m \otimes \chi)} + \lambda l(\theta), Q \in \lim_{\epsilon\to0} R(\epsilon, Q, F)
\end{equation}
$l(\cdot)$ is an L2-norm that helps to speed up the optimization process and $\lambda$ controls the strength of penalty.

\subsection{Model Description}
\label{sec: md}
\begin{figure}[h]
\centering
\includegraphics[width=.5\textwidth]{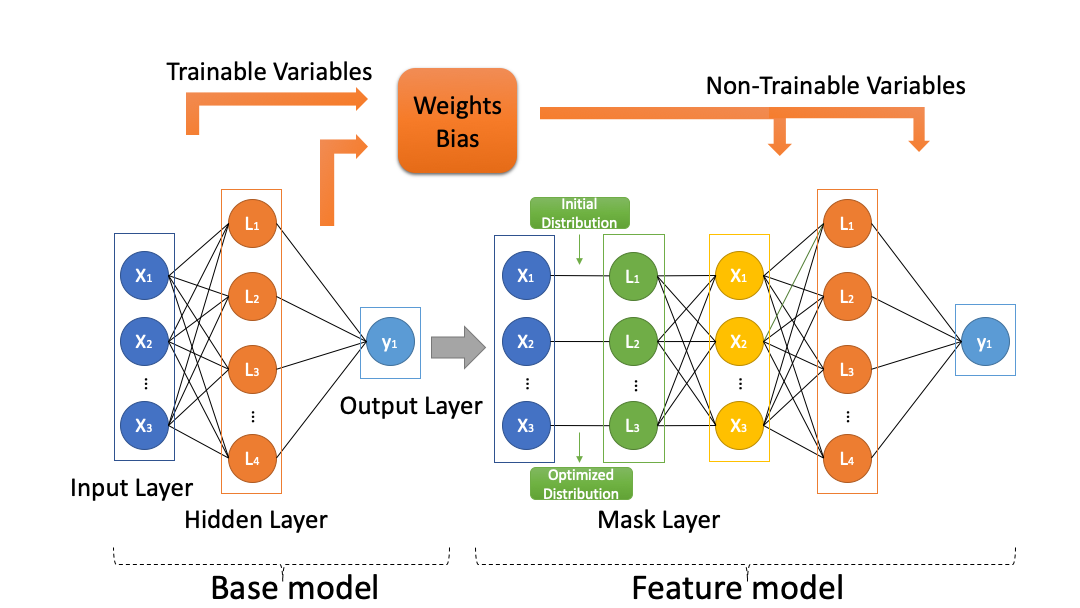}
\caption{The structure of the framework}

\label{fig:structure}
\end{figure}

To obtain the mask $m$ stated in Eqn.\ref{eqn:coefficient_transfer}, we propose a framework consisting of a base model and a feature model, as shown in Fig.\ref{fig:structure}. The base model is selected and employed to accomplish a supervised learning task, e.g., classification or regression, while the feature model is designated to discover the feature importance through a systematic retraining process. The feature model follows the trained base model and fixes the model parameters, while adding an extra mask layer to learn. The feature model is retrained to achieve the loss retrieved from base model and the weights of mask layer are extracted as the output. 

Due to the randomness of weights during initialisation, learning rate and other factors, the value of $m$ is contingent to a certain extent. Weights can also be misleading with regards to performance. In the case where an unimportant feature is assigned to a high weight, the overall performance is not affected because its contribution to the outcome is virtually insignificant.  
The weight list in $m$ is the final parameter used for the feature selection and feature importance ranking.

\subsubsection{The choice of initial mask}
It is observed that these final mask values depend on the initials used for starting the training process in Supporting Information Proof of learning algorithm. The choice of the initial weights follows 3 principles.
\begin{itemize}
    \item The weight of each feature must be approximately 0, making sure no initial feature contributions.
    \item Each feature equally contributes to the model.
    \item The weight is randomly chosen so that each iteration commences differently.
\end{itemize}
In summation, the initial weights are distributed as $U(-0.05, 0.05)$, so that the contribution of each feature is not identical but minor differences can be ignored and the final weight accumulates the effect of all connection weights across all hidden neurons in the network.

Two model-agnostic methods are devised based on VTF for interpreting the model with domain knowledge in the Rashomon set.

\subsection{Recursive Variance Tolerance Weight (RVTW)}
From a VTF point of view, the scaling factors (weights) of the important features $w_{X_{i}}$ tends to 1 as the model restricts the change of the feature instances in training at global level. 
On the other hand, important features statistically generate greater weights, for example in linear regression, where increasing the value of a feature by one unit changes the estimated outcome by its weight. 

To be more convincing we use a data processing strategy that averages a list of weights from repeatedly retraining of the feature model until the average of the weights reaches a stable point. 
The weight is also scaled with its standard error to mitigate the uncertainty, because if the model performance is unaffected by large variations in the scaling factors, then the feature is consequentially unimportant.

 \begin{equation}
v_{X_{i}} = \frac{mean(w_{X_{i}})}{mean(t_{w_{X_{i}}})e_{w_{X_{i}}}} = \frac{\frac{1}{N}\sum_{j=0}^{N}w_{X_{ij}}}{e_{w_{X_{i}}}\frac{1}{N}\sum_{j=0}^{N}t_{w_{X_{ij}}}}\\
\label{eqn:tolerance_var}
\end{equation}
where the value of $v_{X_{i}}$ is the measurement of feature importance, $N$ is the number of training times, $w_{X_{ij}}$ is the coefficient of the feature $X_{i}$ at training time $j$.
 $e_{w_{X_{i}}}$ is the standard error $STD(w_{X_{i}})$.

\subsection{Contribution Factor (CF)}
Considering the equivalence of Eqn.\ref{eqn:coefficient_equal}, another way to approach $m$ it is to solve this as an equation system.
For any feature set $\chi$ $(d \geq 1)$ (number of features is greater or equal to 1) contributing to a model, the contribution can be accumulated, such that:
\begin{equation}
\mathbb{C}(Q, \chi) = \sum_{i=0}^{d}\mathbb{C}(Q,X_{i})
\label{eqn:additivity}
\end{equation}
where $\mathbb{C(\cdot)}$ is a contribution function.

The idea of CF is to treat each feature contribution as an unknown in an equation and construct a solvable equation system.
After the feature retraining, we can immediately derive Eqn.\ref{eqn:contribution_additivity} based on Eqn.\ref{eqn:additivity}. A factor $\mu$ is introduced to represent the relationship between the weight $w$ and the contribution $\mathbb{C}(X)$, and can be derived through experiments. The general case is formulated in Eqn.\ref{eqn:general_system}. A detailed proof of algorithm is shown in Supporting Information Derive of Contribution Factor.

\begin{equation}
\begin{split}
\mathbb{C}(X_{1}) + \mathbb{C}(X_{2}) + ... + \mathbb{C}(X_{i}) = \\
\mathbb{C}(w_{X_{1}}X_{1}) +  \mathbb{C}(w_{X_{2}}X_{2}) + ...+  \mathbb{C}(w_{X_{i}}X_{i})
\label{eqn:contribution_additivity}
\end{split}
\end{equation}

To construct the equation system the number of retrain attempts must be equal or greater than the number of features, which might result in an overdetermined system. Theoretically, it is enough to solve unknowns in an equation system, given the exact same number of equations, but we need more equations to reduce the effect of bias introduced from unimportant features (discussed above), so that with $d$ unknowns and $N$ equations: $(N>d)$. Matrix reduction is necessary to solve this overdetermined system and Gaussian elimination can be applied. With reduced row echelon form, the equation system can be solved, and the contribution of each feature to the performance can be derived.

\begin{equation}
\left\{\begin{matrix}
\mathbb{C}(X_{1}) + \mathbb{C}(X_{2}) + ... + \mathbb{C}(X_{i}) = 1\\
\mu_{1} \mathbb{C}(X_{1}) + \mu_{2}  \mathbb{C}(X_{2}) + ...+  \mu_{i}  \mathbb{C}(X_{i}) = 1\\
\vdots
 \\
\mu_{1}' \mathbb{C}(X_{1}) +  \mu_{2}' \mathbb{C}(X_{2}) + ...+   \mu_{3}' \mathbb{C}(X_{i}) = 1\\ \end{matrix}\right.
\label{eqn:general_system}
\end{equation}

%\subsection{Workflow}\label{sect:work}
%
%Given the raw dataset, preprocessing is undertaken by splitting the datasets to training, testing and validation sets. This process is accompanied by the selection of a machine learning model that is suitable for the specific task (either regression or classification). The base model is trained and optimized using $k$-fold cross validation, in which all layers and weights are saved for further processing. To extract the target mask, it is necessary to utilize another model with the same structure but containing the feature model. The feature model is built upon the base model, with one extra $1\times N$ mask layer, where $N$ corresponds to the number of features in the dataset. In particular, the mask layer is trainable while all layers with weights from base model are fixed. After set up, the feature model is trained and optimized using $k$-fold cross validation until the performance, measured by the loss, matches the base model. Once the training is complete, the weights from the mask layer can be extracted. To reduce the effect of noise on the weights, the feature model is retrained repeatedly, and the number of training times depends on the number of features and specific requirements.

\section{Proof of Concept}

To demonstrate the basic feasibility and reliability of the theory, a linear regressor and logistic classifier are selected as a baseline test. The framework is applied to benchmarks and a real-world dataset to evaluate it against practical problems. All datasets used herein are publicly available and were split into 80\% training set and 20\% testing set prior to optimization with the same random seed. All models and experiments are conducted using TensorFlow software and their details are tabulated to provide the greatest opportunity for experimental control and reproductivity. Training results are tabulated, or included in learning curves which also quantify any under-fitting or over-fitting. Feature importances are ranked using histograms to give intuitive insights of their effects. 
% add baseline
To test the reliability and accuracy of the proposed methods, we compare their performance with other well-established methods.
\subsubsection{Model-agnostic methods}
\begin{itemize}
\item Fisher Score: selects features based on similarities.
\item ReliefF: selects features based on the identification of near-hit and near-miss instances.
\item Permutation method: gives another feature importance profile by measuring the increase in the base model prediction error after permuting a feature.
\item AFS: attention-based supervised feature selection.
\item KernelShap: computes the Shapley values that fairly distributes the prediction to each individual feature.
\end{itemize}
\subsubsection{Model-specific methods}
\begin{itemize}
\item Random Forest (RF): regressor provides feature importance based on Gini impurity.
\item Connection weights: calculates the sum of products over all hidden layers and ranks the relative importance by the outputs.
\end{itemize}

\subsubsection{Evaluation Protocols}
To evaluate the performance of feature selection method, the traditional recursive feature elimination (RFE) method is applied to the same model and dataset, with the same feature importance rankings.

Unimportant features are selected according to their importance by percentage, from 10\% to 90\% and a new model independent from previous training and outputs is used to fit the selected subset. The loss or accuracy of the feature subset indicates the importance of dropped features as the performance metric. For MNIST dataset, we followed the evaluation protocols of \cite{gui2019afs} to provide a fair comparison.

\subsection{Linear Regression and Logistic Regression}

\begin{figure*}[ht!]
\centering
\subfigure[]{\includegraphics[width=0.28\textwidth]{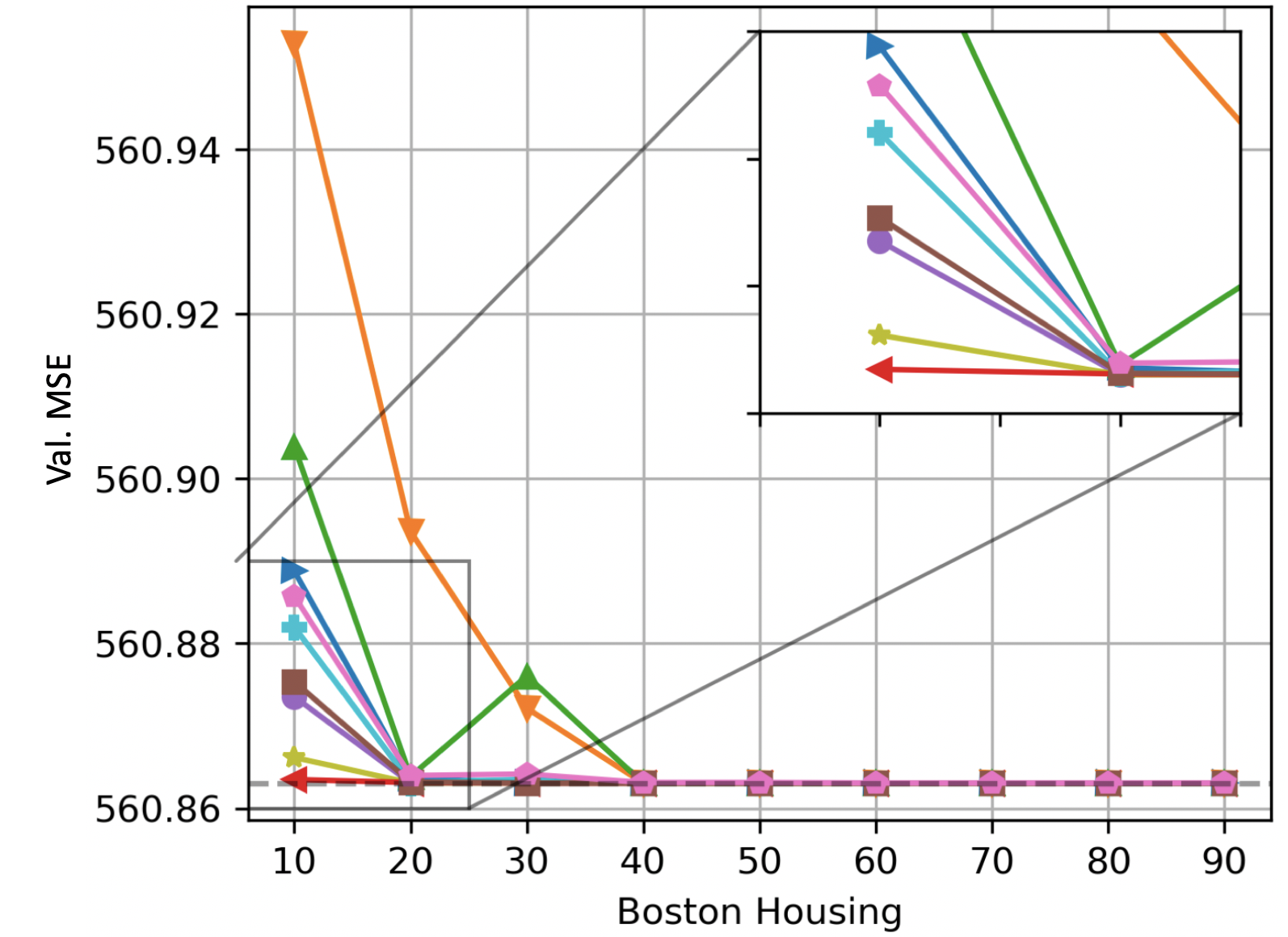}}
\subfigure[]{\includegraphics[width=0.275\textwidth]{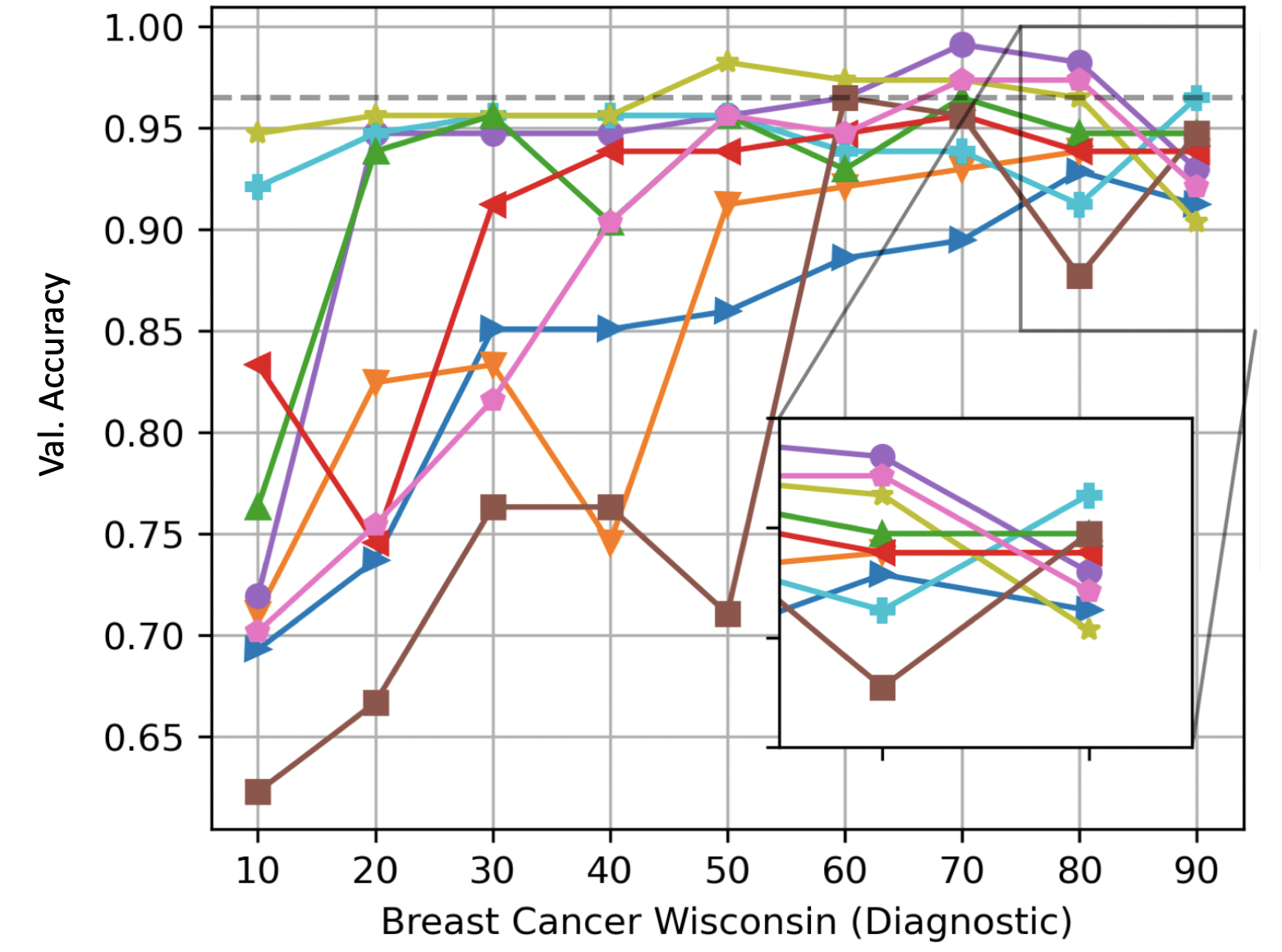}}
\subfigure[]{\includegraphics[width=0.375\textwidth]{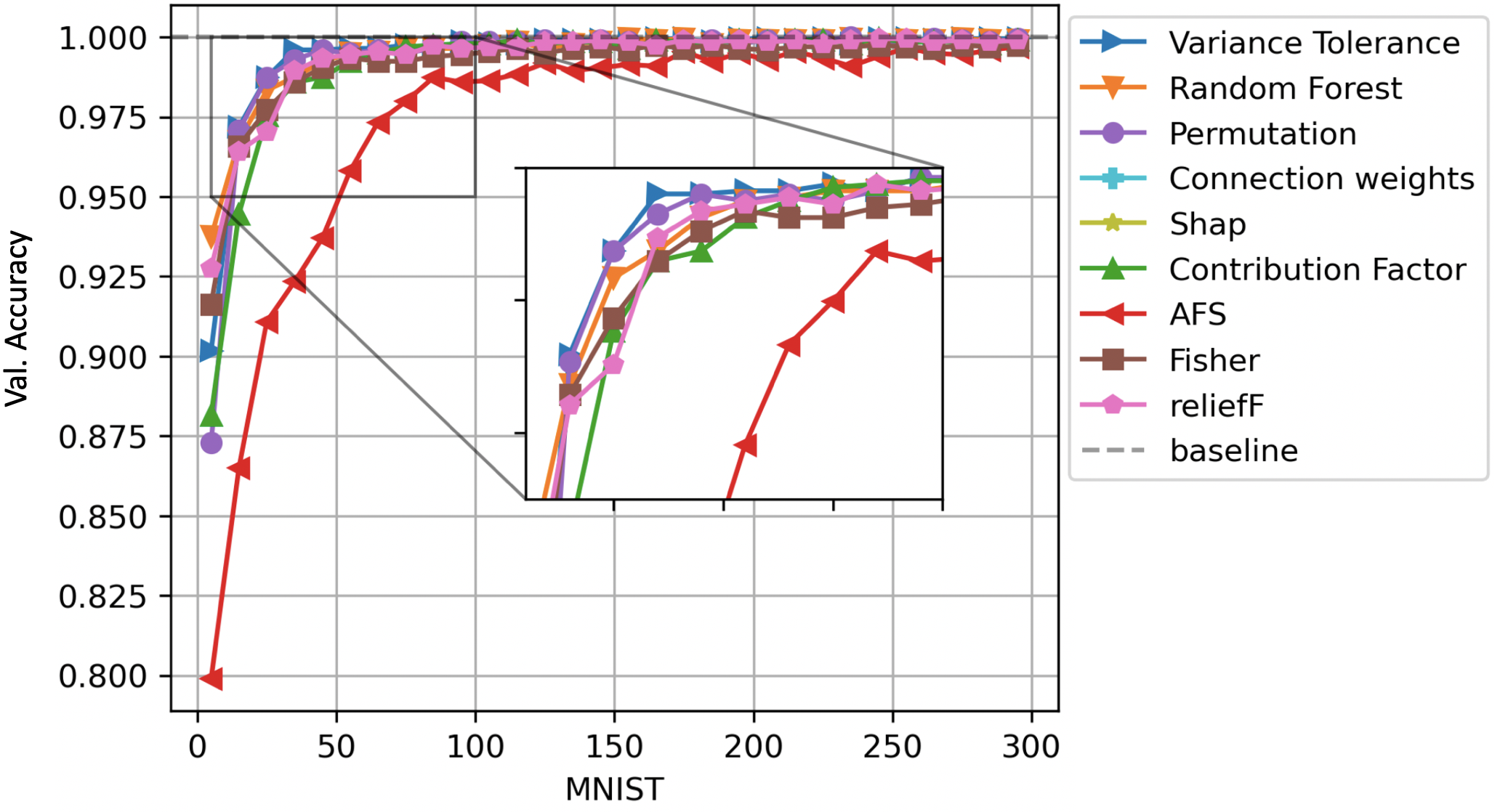}}
\caption{Feature selection methods' reliability and accuracy comparison using dataset (a) Boston house price and (b) Breast Cancer Wisconsin (c) MNIST. In the first two datasets, the percentage of selected unimportant features are set from 10\% to 90\% with an interval of 10\%, while the numbers of selected features are set from 3 to 295 with an interval of 10 in MNIST, following the protocol \cite{gui2019afs} }
\label{fig:baseline_results}
\end{figure*}

The dataset used in the linear regression is The Boston Housing Dataset \cite{asuncion2007uci}, which was collected in 1978 and contains information about houses from various suburbs in Boston, Massachusetts. The dataset can be accessed from the scikit-learn library \cite{scikit-learn}, including 506 samples and 13 feature variables.
The dataset used in the logistic regression is Breast Cancer Wisconsin (Diagnostic) Dataset \cite{wolberg1992breast}, which was collected in 1955 and contains information about characteristics of the cell nuclei present in the image of a fine needle aspirate (FNA) of a breast mass. The dataset can be accessed from the scikit-learn library \cite{scikit-learn}, including 569 samples and 30 features. 

A linear regressor and a logistic classifier are built to model the relationship between independent variables and one dependent variable. The model and training details are in Supporting Information Table.2. The base model was first trained with all features to make the decision for 1000 epochs, achieving low testing errors and high accuracies, with no over-fitting and no under-fitting, shown in Supporting Information Fig.2. 

\subsubsection{Baseline Recursive Variance Tolerance Weight}
Based on the outputs of the base model, including its structure and weights, the feature model was constructed with the exact same settings except for one mask layer on the top of the base model. Then the feature model was trained and optimised using all features to predict the target until the loss achieved the same number as base model, which training and details are summarized in Supporting Information Fig.2. 
According to the theory, the retraining process was conducted numerous times and visualised in the learning curve to provide an insight into the process, from which it is noted that to achieve the same performance the feature model requires a relatively small number of epochs, illustrated in Supporting Information Fig.2. The linear regressor achieves the loss target in 175 epochs and the logistic regressor achieves the loss target in 300 epochs with no over-fitting and no under-fitting.

In both cases, the feature model is retrained 250 times and the weight information (weight and variance) retrieved from training iterations are visualized in a weight plot in Supporting Information Fig.1. The feature importance is ranked based on the RVTW and to compare the difference with the effect of the VTF, we applied both methods in the feature set, resulting a feature importance profile and a feature unimportance profile, as depicted in the histogram in Fig.\ref{fig:linear_base_model_hist_TV}.  The feature importance profile varies as the number of retrain times increasing and becomes stable after some iterations, shown in Supporting Information Fig.4.

\subsubsection{Baseline Contribution Factor}
Given the results from feature model retraining, an equation system was constructed as described in Eqn.\ref{eqn:general_system}, where $\mu$ remains unknown. To obtain this parameter, the base model was used to evaluate various settings of the feature set according to the procedure in Supporting Information and the loss change was then used to simulate the relationship between weights and contributions $\mu$.

\begin{figure}[h!]
\centering
\subfigure[]{\includegraphics[width=0.25\textwidth]{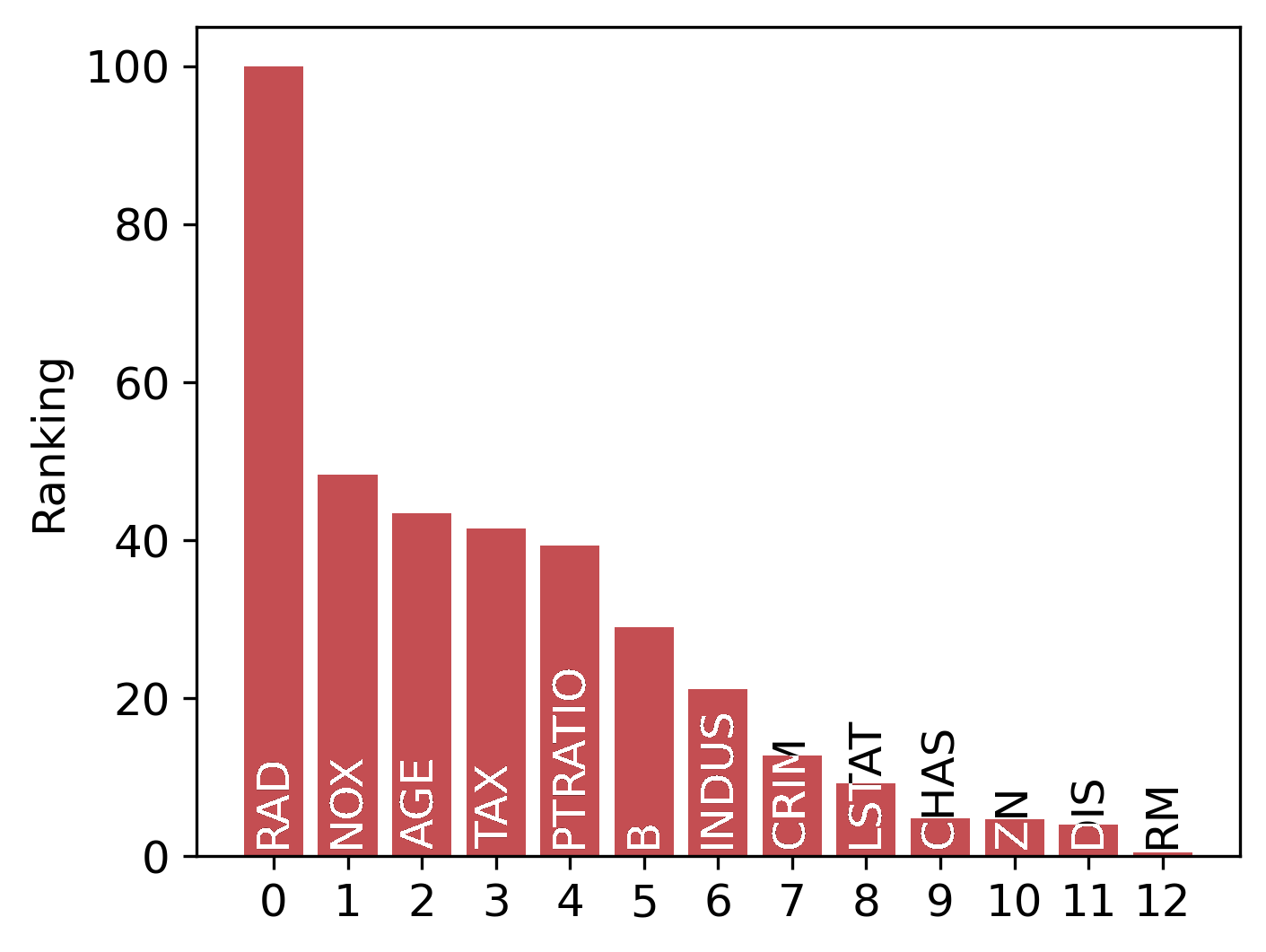}
\includegraphics[width=0.25\textwidth]{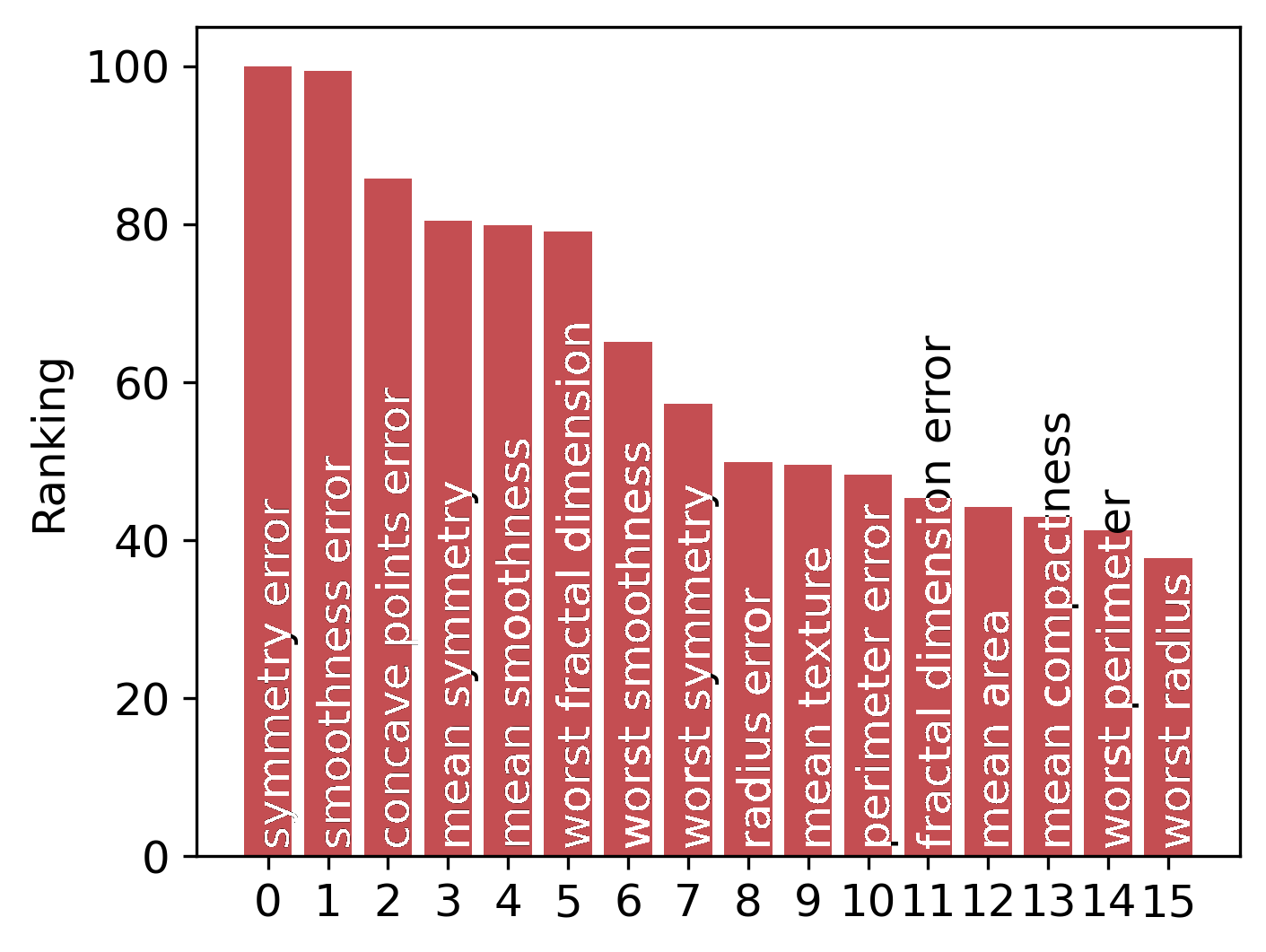}}
\subfigure[]{\includegraphics[width=0.25\textwidth]{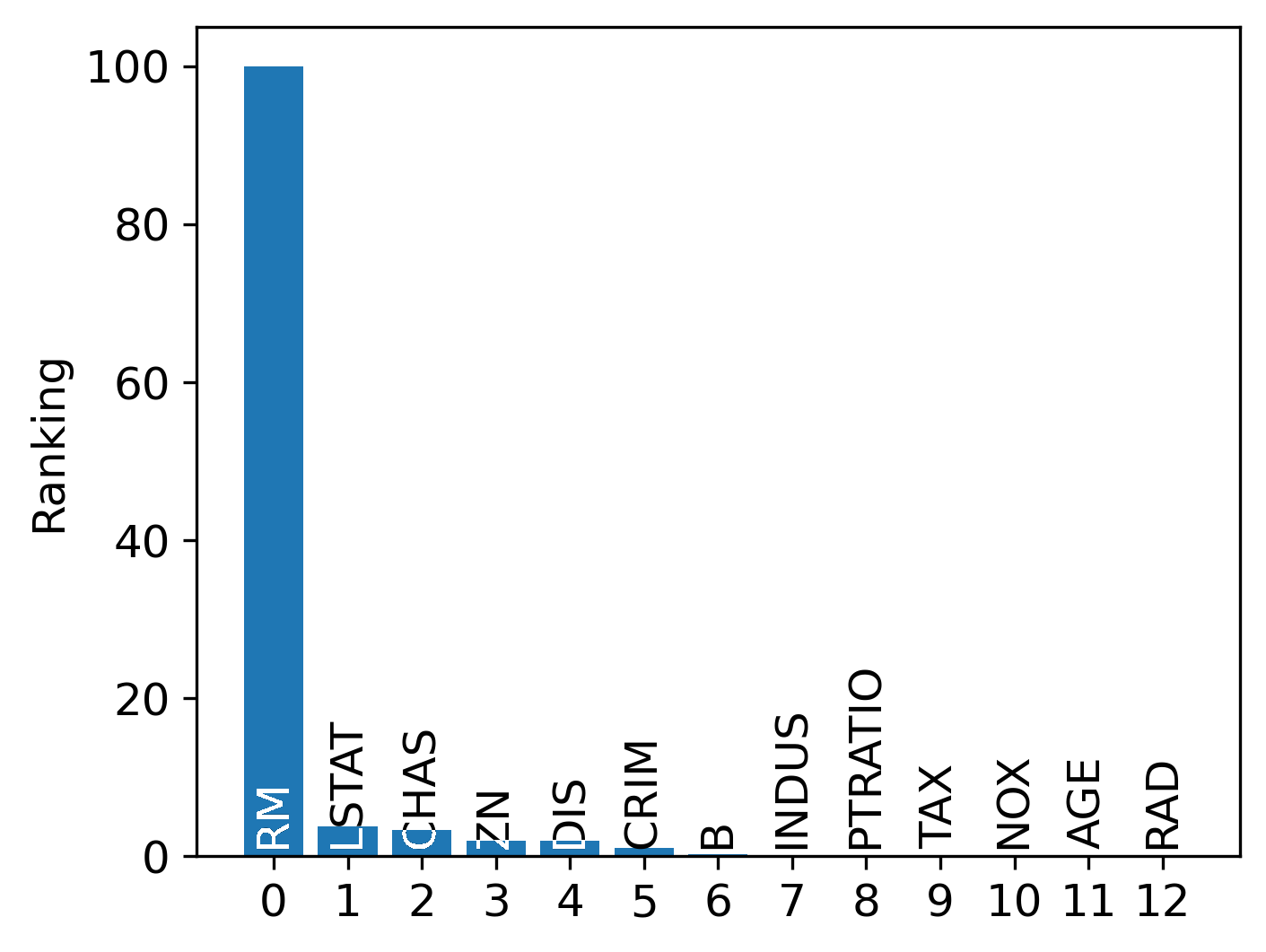}
\includegraphics[width=0.25\textwidth]{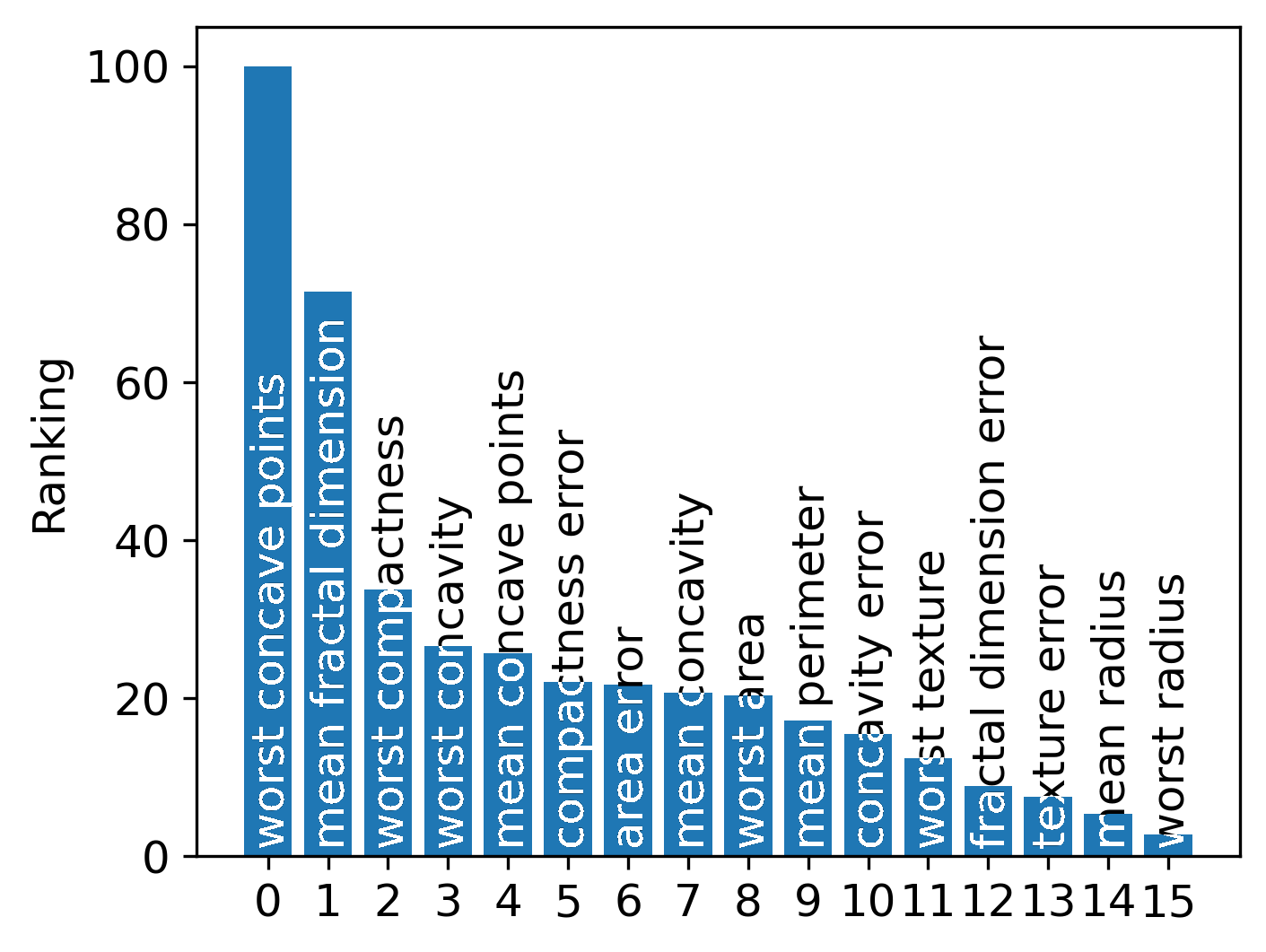}}
\subfigure[]{\includegraphics[width=0.25\textwidth]{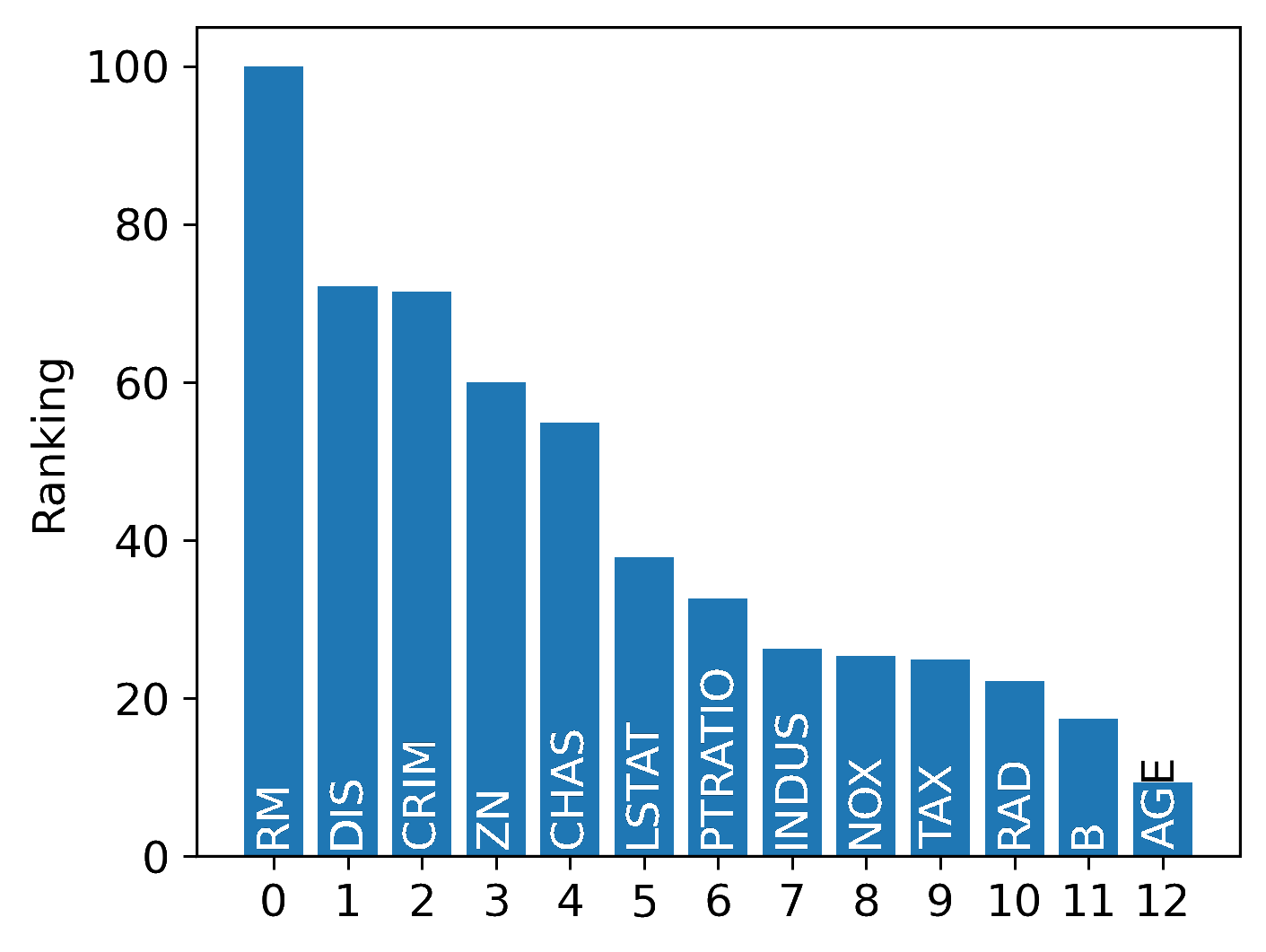}
\includegraphics[width=0.25\textwidth]{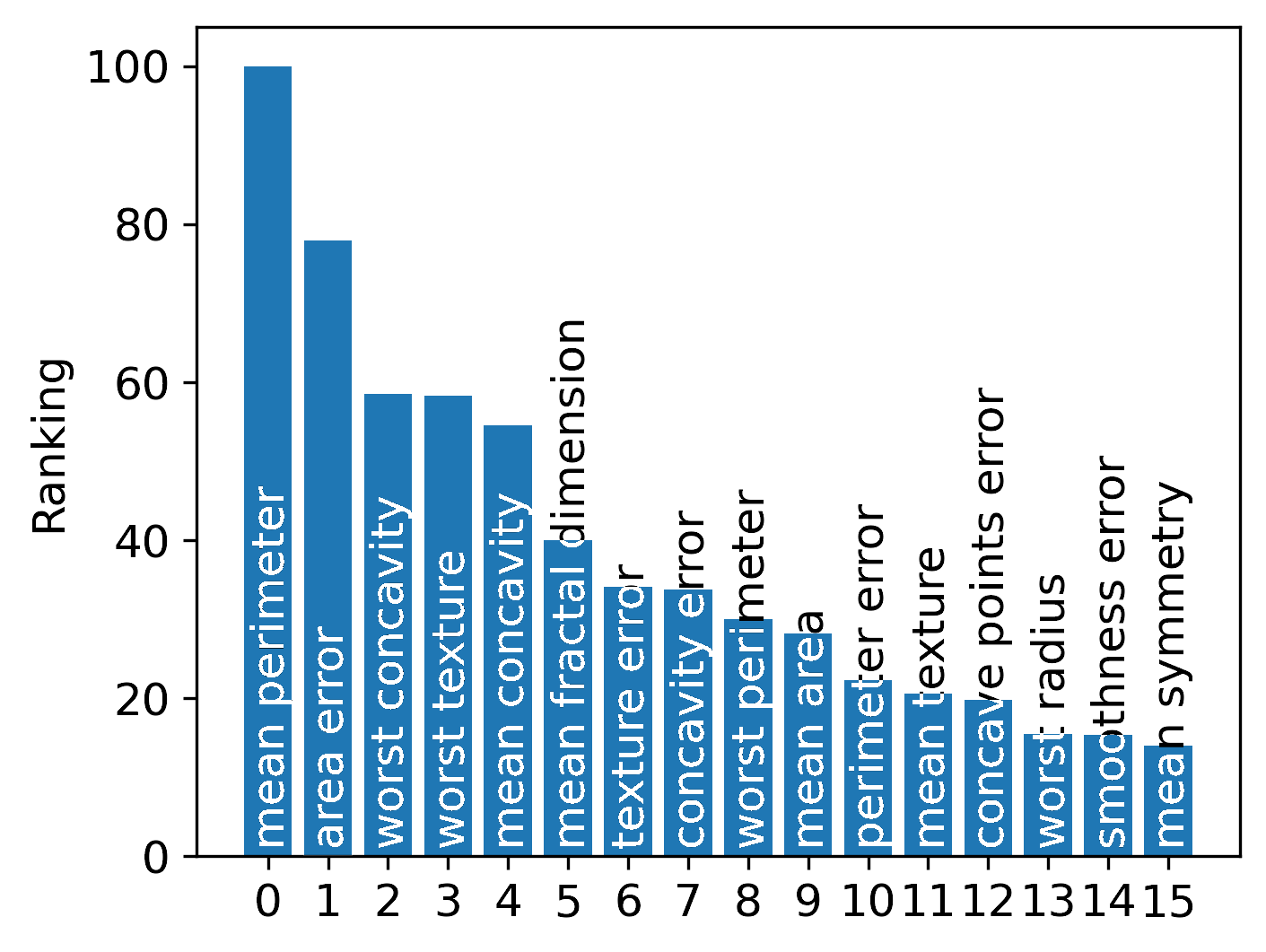}}

\caption{The feature importance rankings of Boston house price (left column) and Breast Cancer Wisconsin (right column) measured by (a) VTF (importance from right to left), (b) RVTW (importance from left to right) and (c) CF (importance from left to right).}
\label{fig:linear_base_model_hist_TV}
\end{figure}

Two equation systems of more than 13 equations with 13 unknowns, and more than 30 equations with 30 unknowns were solved with Gaussian reduction algorithm, where the  unknowns represent the contribution of each single feature in the feature set. To capture the relationship of the distribution of each feature, the results of the equation system are divided by one of the features' distribution.
 Fig.\ref{fig:linear_base_model_hist_TV} (c) presents an overview of the contribution of each feature and detailed contributions are also provided in Supporting Information Fig.5. 
 With the more equations included, the change in the contribution distribution is shown in Supporting Information Fig.6.
 
 % baseline results
 \subsubsection{Baseline Evaluation}
 From Table.\ref{tb:app_model_feature_selection}, it shows that our feature selection method can achieve similar performance in much shorter time.
 The permutation method, connection weights, RF, KernelSHAP, Fisher Score, ReliefF and AFS are used as a benchmark to evaluate the performance of the theory applied in the linear regressor. All feature importance profiles are provided in Supporting Information Fig.7. 
From the importance profile, we can see that  \texttt{LSTAT} and \texttt{RM} occurs most among important features in all rankings, and \texttt{AGE} and \texttt{NOX} occurs most among least features in all rankings for Boston Housing Dataset. From Fig.\ref{fig:baseline_results} (a) we can observe that the model performance drops more with 80\% and 90\% important features extracted from RVTW and CF than others except RF, while our methods still accurately detect the important features. In the case of 70\% important features removed, CF outperforms others and from 60\% important features removal, all methods perform similarly. In terms of logistic regression case, \texttt{mean perimeter}, \texttt{worst area} and \texttt{worst concave points} occurs most among important features in all rankings, and \texttt{symmetry error}, \texttt{worst error} and \texttt{mean fractal dimension} occurs most among least features in all rankings. Fig.\ref{fig:baseline_results} (b) shows that accuracy gaps between baseline and results from RVTW are wider than most methods, while CF performs in average.
%
%\begin{table}[h!]
%\caption{Comparison of feature selection methods}
%\label{tb:linear_model_feature_selection}
%\centering
%\begin{tabular}{llll}
% \multicolumn{4}{c}{Linear regression} \\ \hline
% & Our method & RFE & Base model \\
%Time complexity & $O(n)$ & $O(n^2)$ & - \\
%MSE & 3.4014 & 3.4097 & 3.4003 \\ \hline
%
% \multicolumn{4}{c}{Logistic regression}  \\ \hline
% & Our method & RFE & Base model \\
%Time complexity & $O(n)$ & $O(n^2)$ & - \\
%Loss & 0.0866 & 0.0834 & 0.0812 \\
%\end{tabular}
%\end{table}

\section{Application}

The framework is also applied to widely-used neural networks, convolutional neural network (CNN), recurrent neural network (RNN) and multilayer perceptron (MLP).
Training details and network structures are tabulated in Supporting Information Table.10.

\subsection{Convolutional Neural Network}
To demonstrate the interpretability of the methods via visual inspection, we employ a CNN to classify the MNIST well-known benchmark dataset \cite{deng2012mnist}, focusing on the hard-to-distinguish digits ``3'' and ``8''. The subset is a dataset of 13966 $28\times28$ grayscale images, and can be accessed from TensorFlow library \cite{tensorflow2015-whitepaper}, with 784 features and 2 classes, where each pixel can be treated as a feature. The base model was first trained for 1000 epochs, achieving 100\% accuracy with low loss.

\begin{figure}[]
\centering
\subfigure[]{\includegraphics[width=0.11\textwidth]{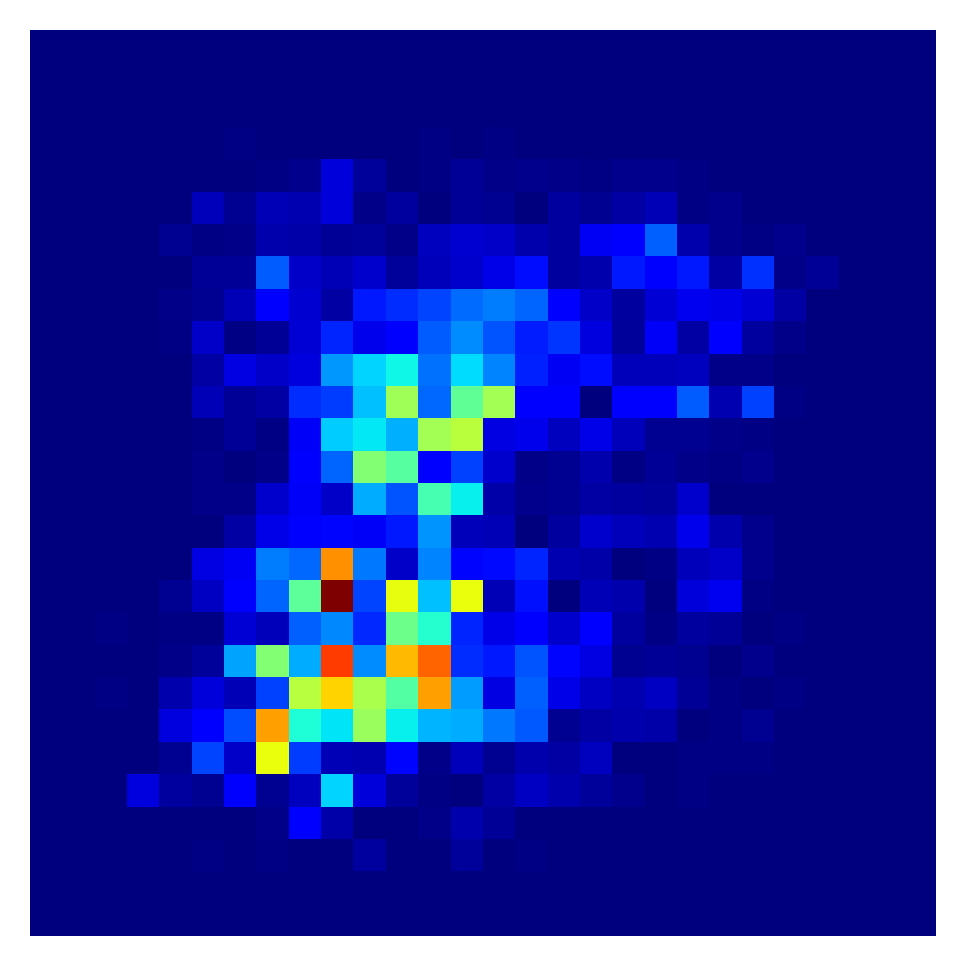}}
\subfigure[]{\includegraphics[width=0.11\textwidth]{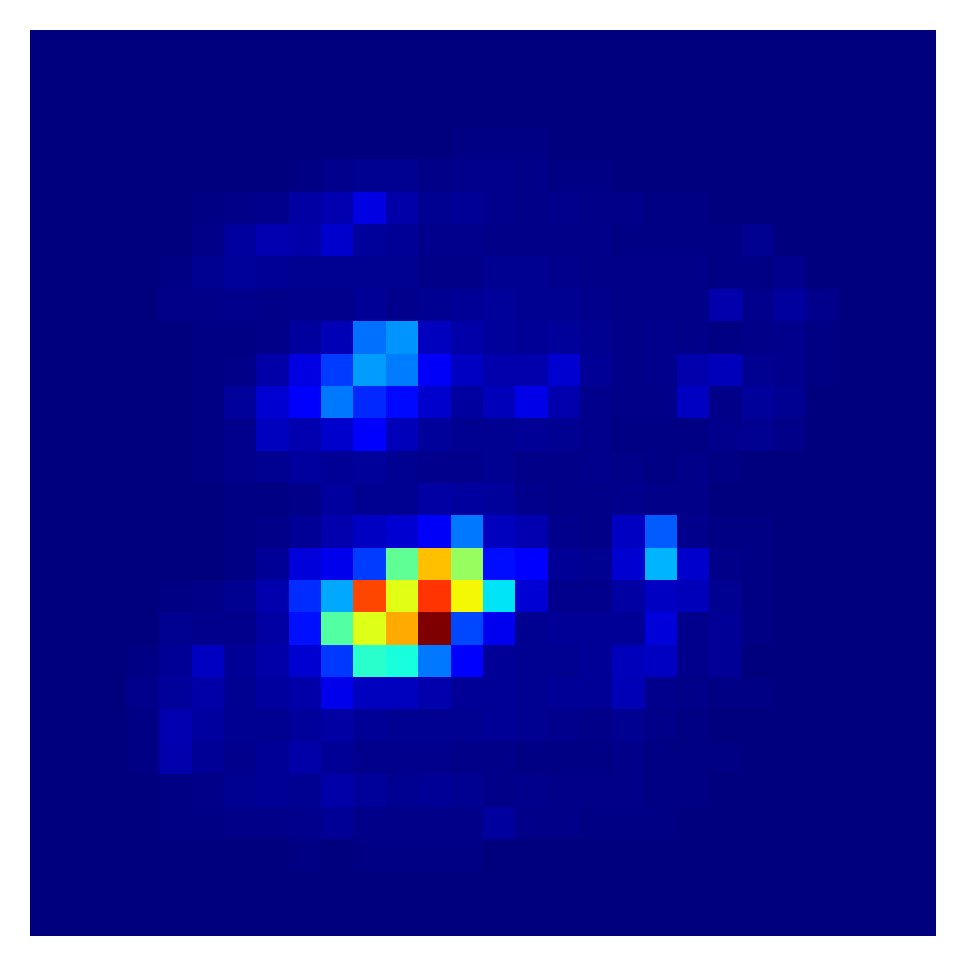}}
\subfigure[]{\includegraphics[width=0.11\textwidth]{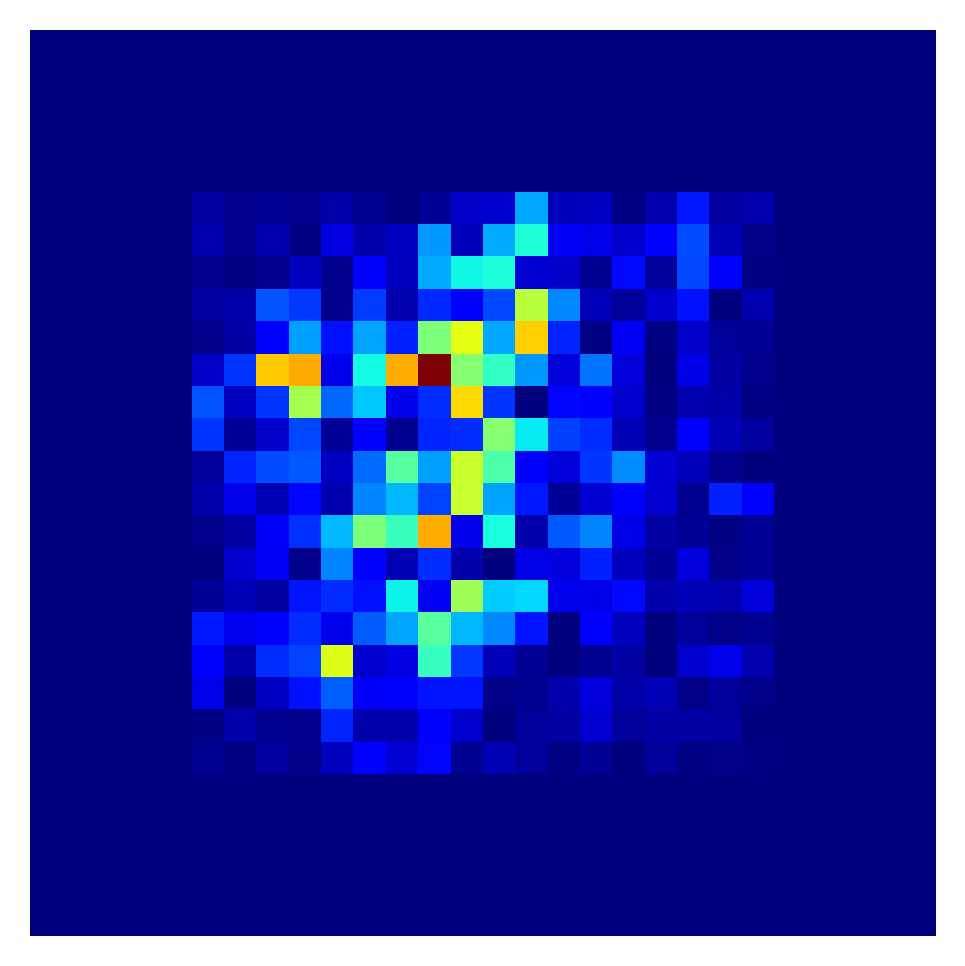}}
\subfigure[]{\includegraphics[width=0.11\textwidth]{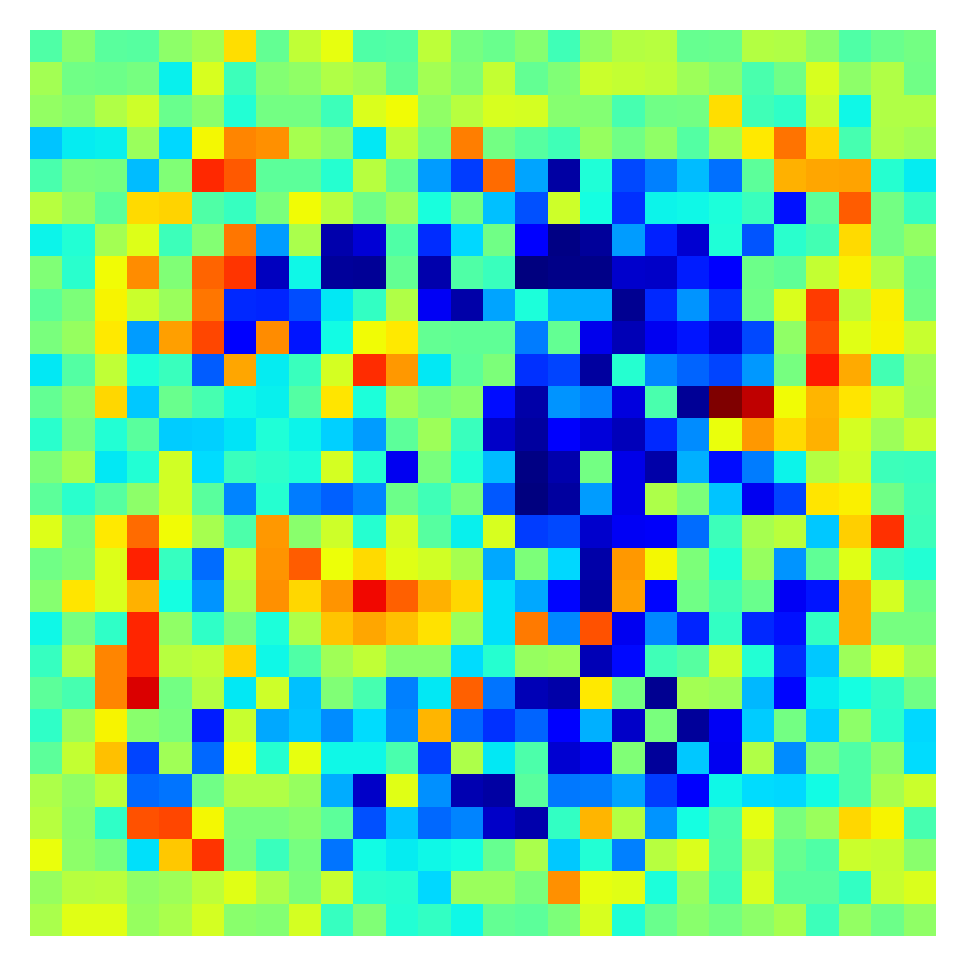}} 
\subfigure[]{\includegraphics[width=0.11\textwidth]{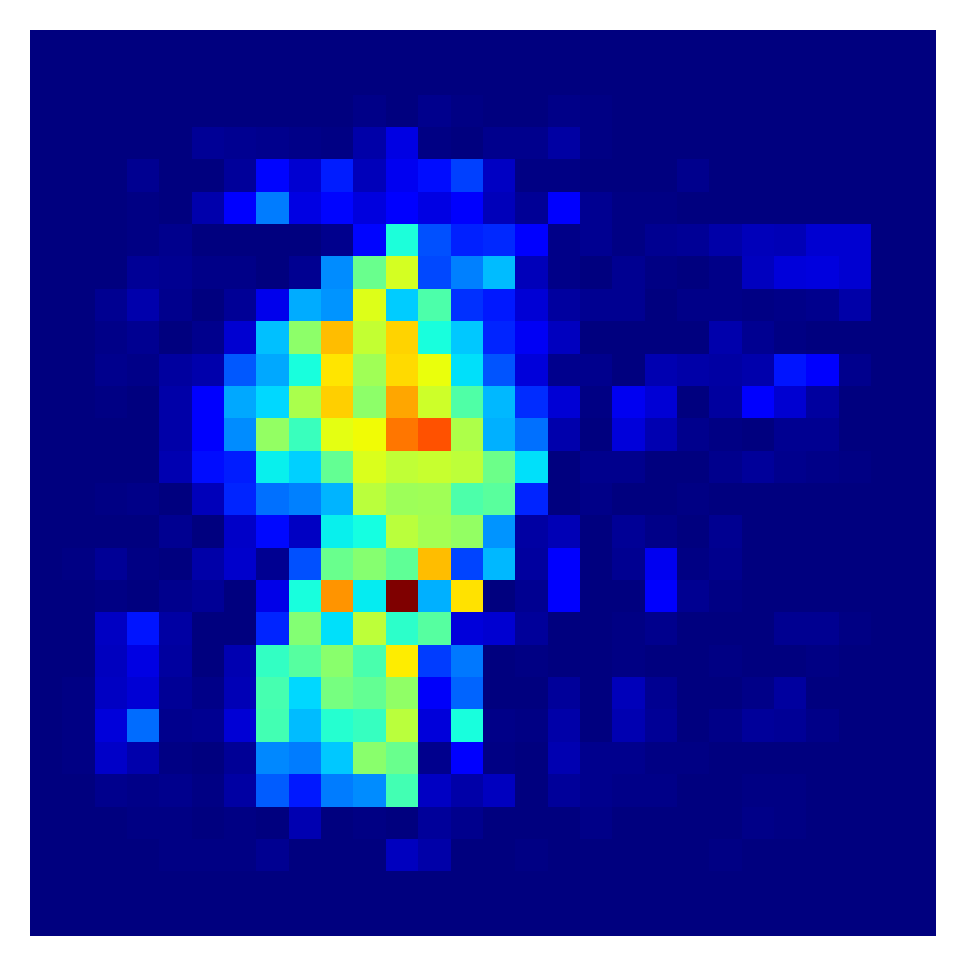}}
\subfigure[]{\includegraphics[width=0.11\textwidth]{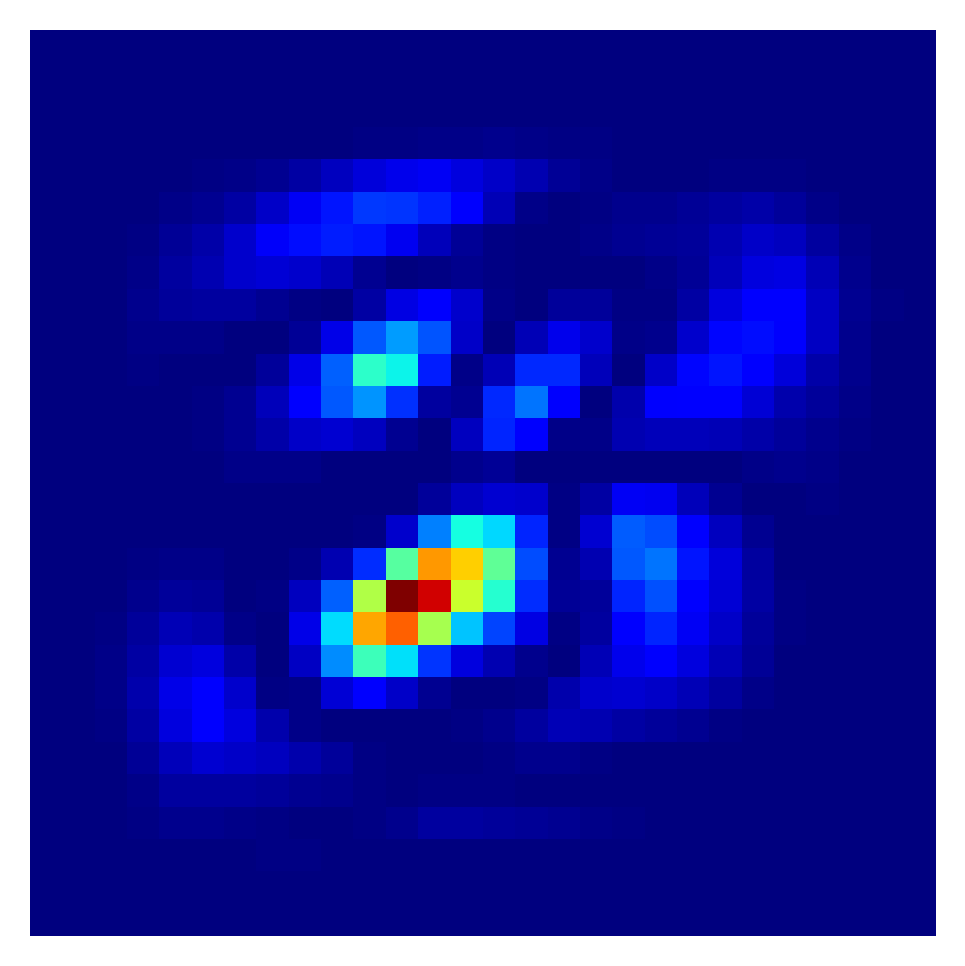}} 
\subfigure[]{\includegraphics[width=0.11\textwidth]{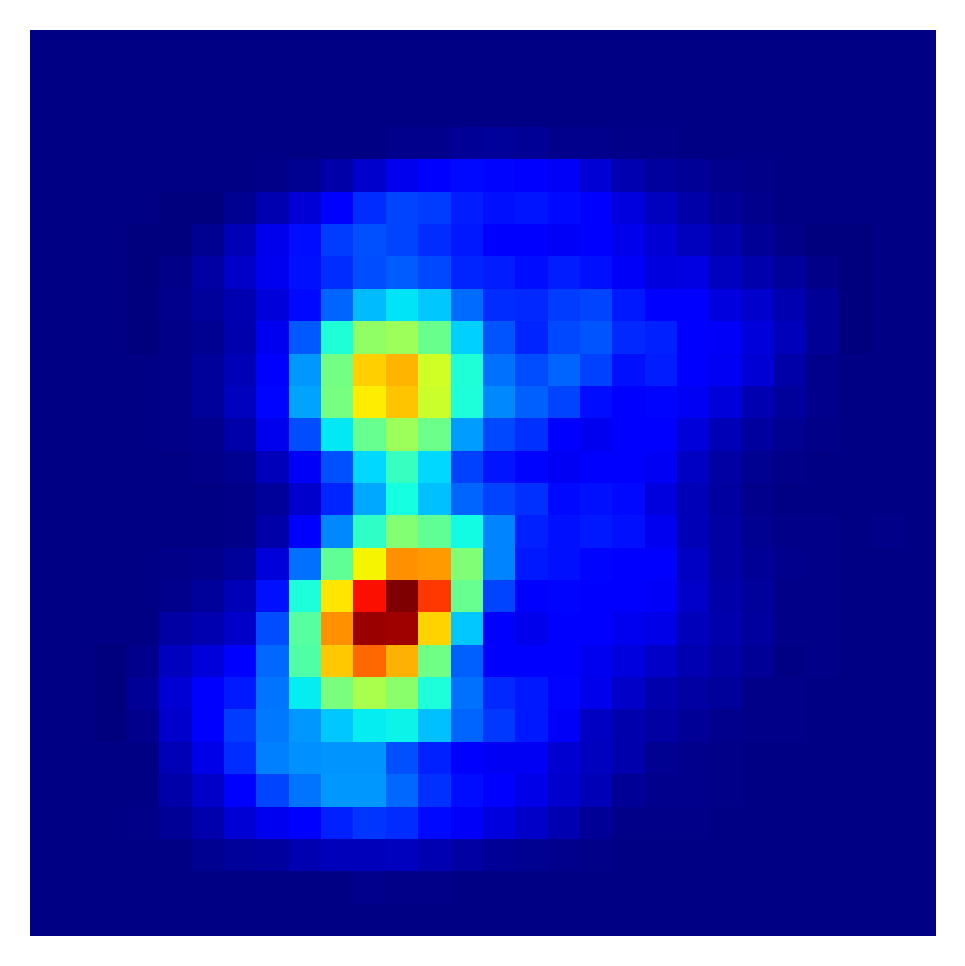}}
\subfigure[]{\includegraphics[width=0.11\textwidth]{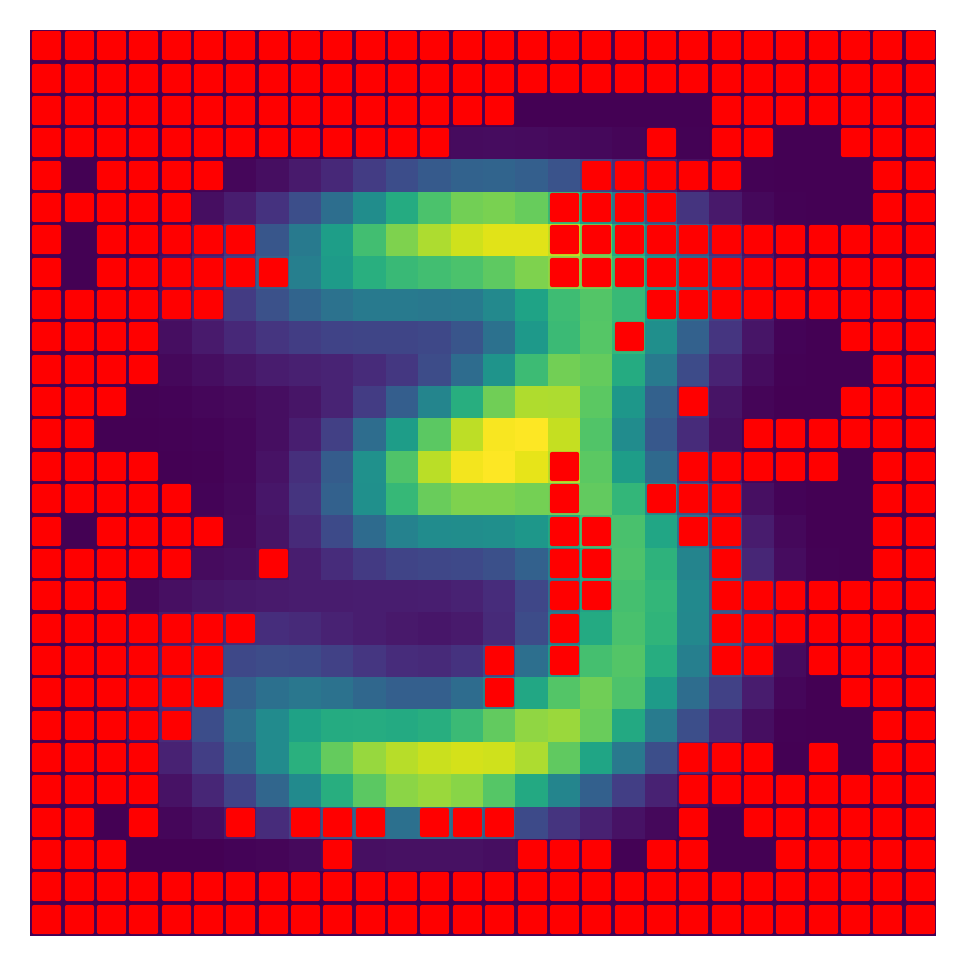}}

\caption{Feature importance maps of MNIST generated from different methods. (a) Permutation method, (b) Random forest (RF), (c) Contribution factor, (d) AFS, (e) RVTW, (f) Fisher Score, (g) ReliefF, (h) Unimportant feature profile}
\label{fig:CNN_results}
\end{figure}

The feature model was trained and optimised using all images to classify the output until the loss achieved the same number as base model. The retraining process was conducted 800 times, which is greater than the number of features (784) needed for the CF. The feature model also achieves the same loss with 100\% accuracy. The feature importance is ranked based on RVTW and CF, the pixel importances as shown in Fig.\ref{fig:CNN_results}. Fig.\ref{fig:CNN_results}(c) presents an overview of the contribution of each feature.

The AFS, Fisher Score, ReliefF, permutation method and RF are used as a benchmark.
The permutation method results in a feature importance map after permuting pixels across all images in Fig.\ref{fig:CNN_results}(a). The RF regressor fitted the relationship between 784 features and the class, with 98.89\% accuracy, and the feature importance based on Gini impurity is ranked in Fig.\ref{fig:CNN_results}(b). We can see from Fig.\ref{fig:CNN_results}(a-g) that the RVTW and ReliefF produce more interpretable and meaningful maps than other methods for human. From the Fig.\ref{fig:baseline_results} (c), 
we can observe that RVTW achieves the best accuracy on almost all feature selection ranges. 
The CF method yields a map similar to the permutation method. Unimportant pixels are marked Fig.\ref{fig:CNN_results} (h). Overall, our proposed methods can produce promising outcomes.

\subsection{Recurrent Neural Network}
To demonstrate the interpretability of the methods in sentiment analysis, we employ the RVTW and CF in a RNN on IMDB movie reviews \cite{maas2011learning}, to classify the overall review as positive or negative, where each word is considered as a feature. For the purpose of demonstration, the dataset is loaded from TensorFlow library \cite{tensorflow2015-whitepaper}, where each review is encoded as a list of integers. For convenience, only the top 2,000 most common words are considered and actual words are indexed from 3. An RNN base model is constructed following the structure and trained for 500 epochs, achieving 88.24\% accuracy with low loss.

\begin{figure*}[]
\centering
\subfigure[]{\includegraphics[width=0.24\textwidth]{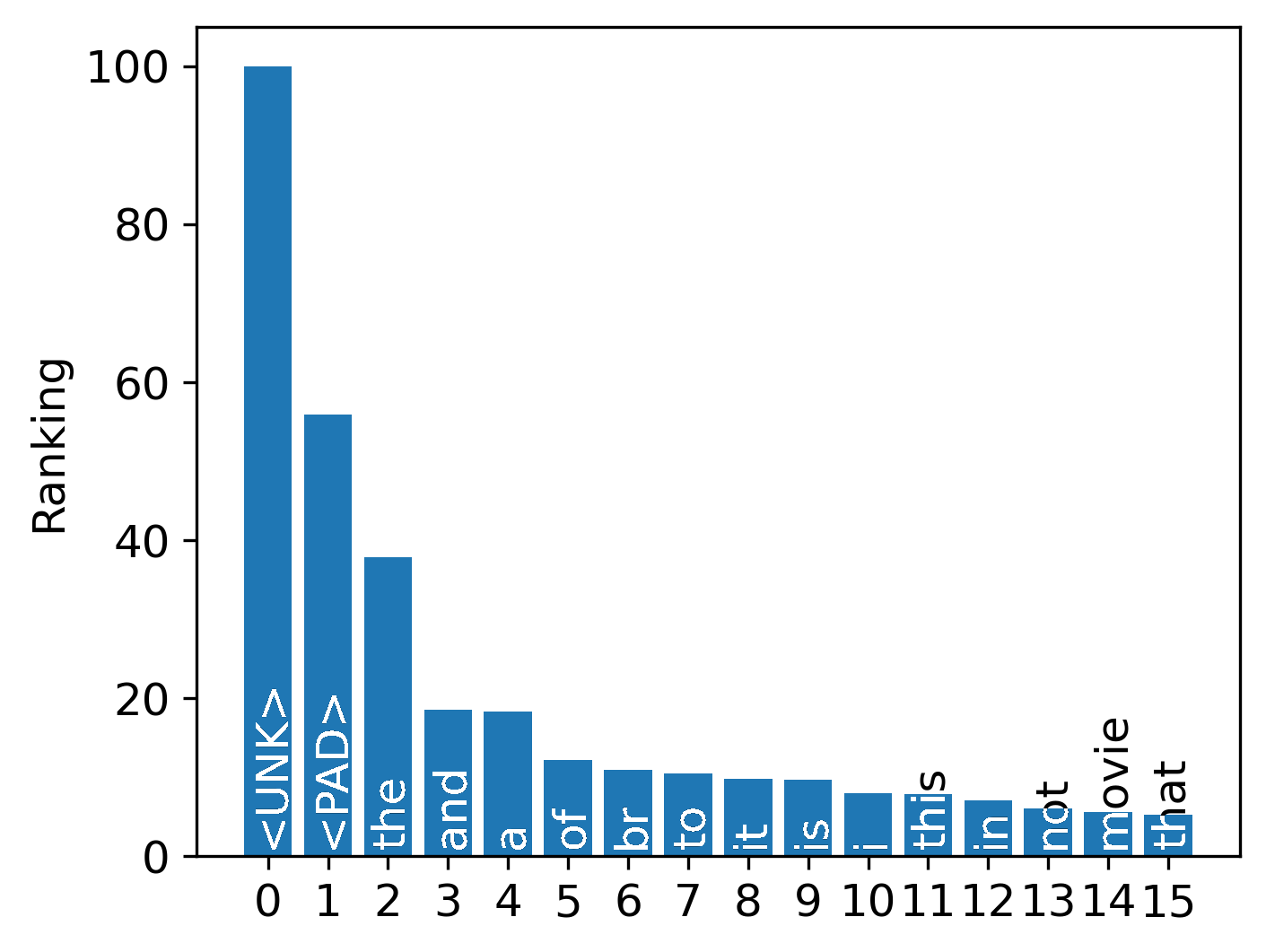}}
\subfigure[]{\includegraphics[width=0.24\textwidth]{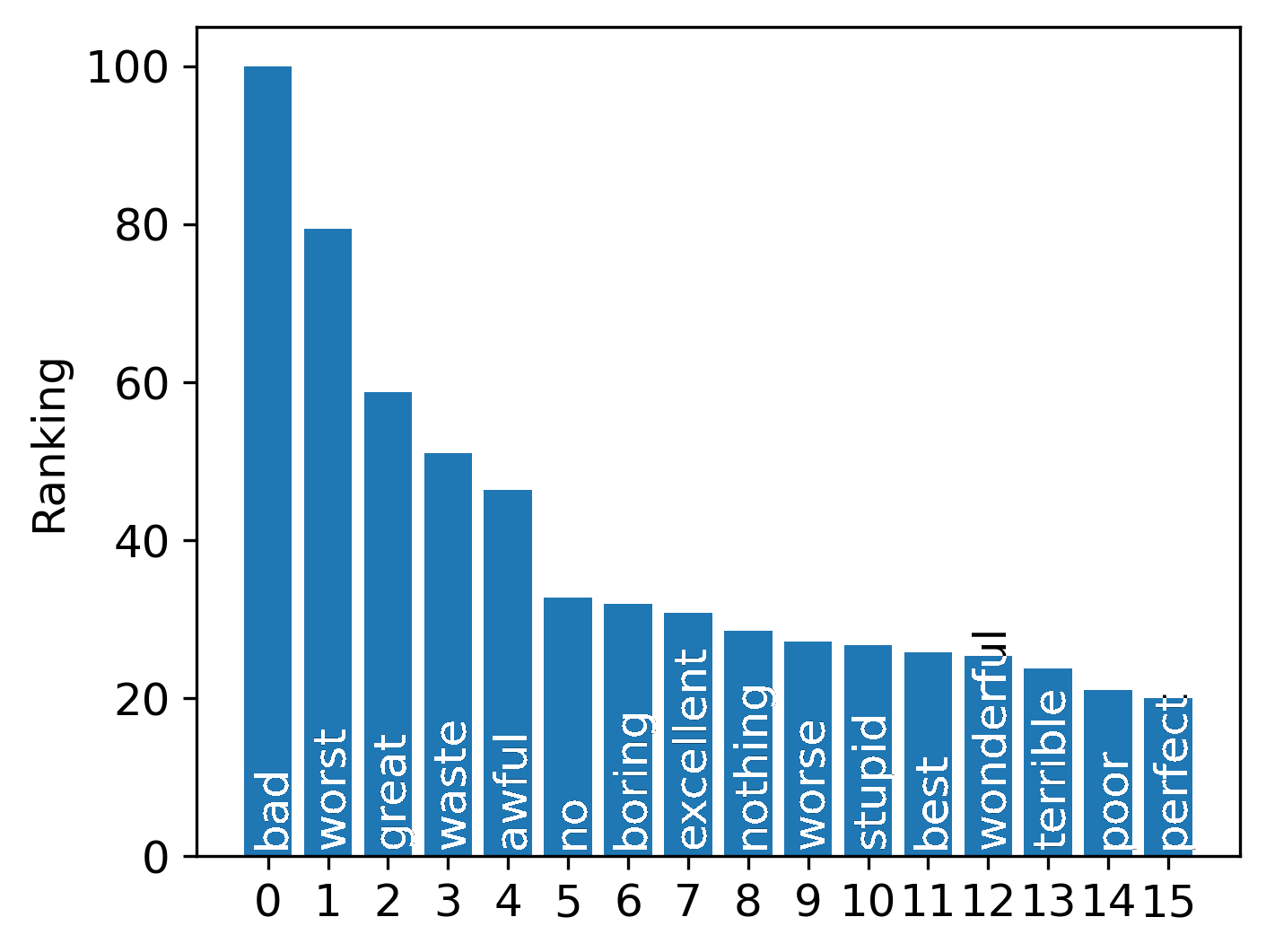}}
\subfigure[]{\includegraphics[width=0.24\textwidth]{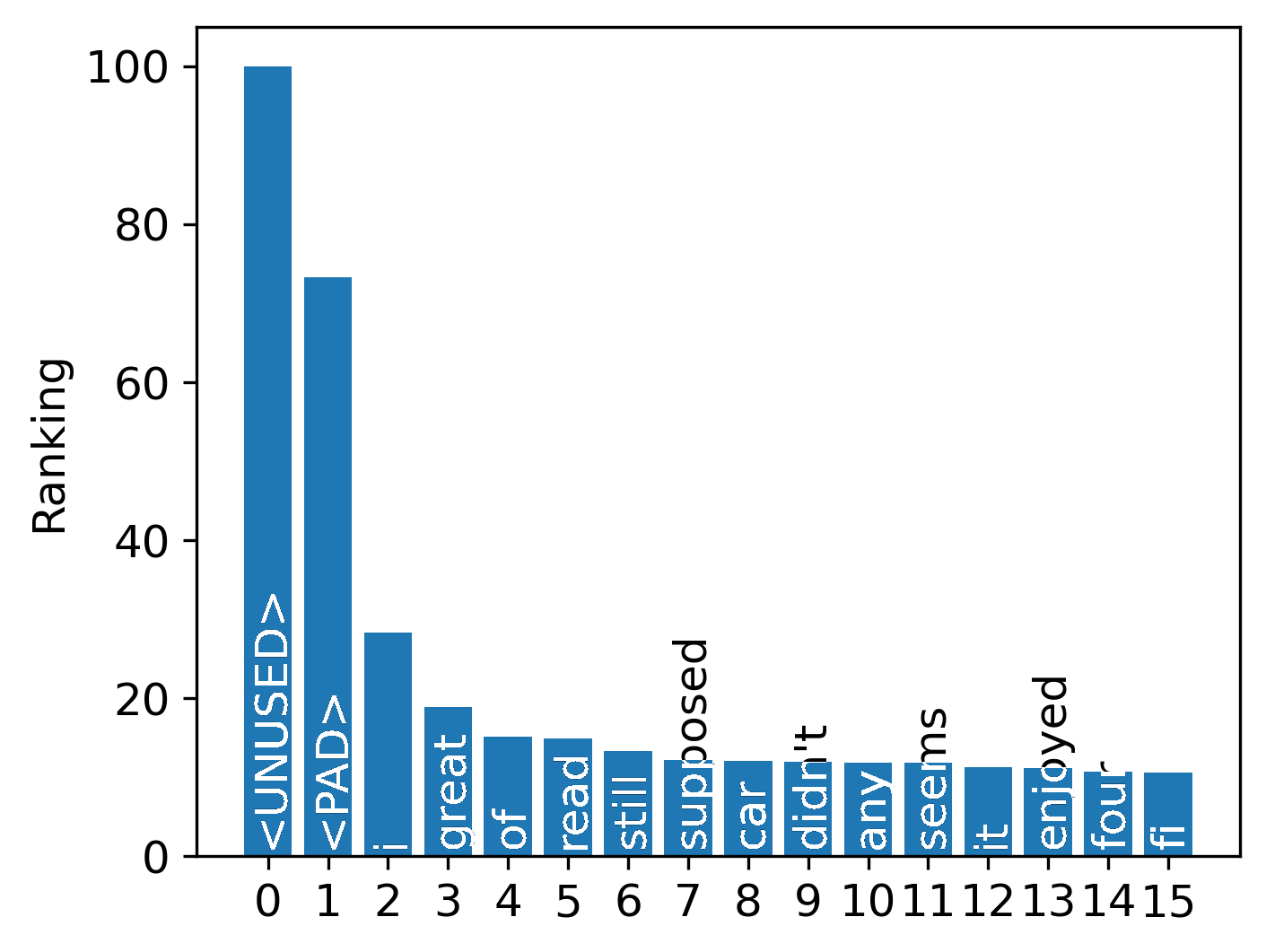}}
\subfigure[]{\includegraphics[width=0.24\textwidth]{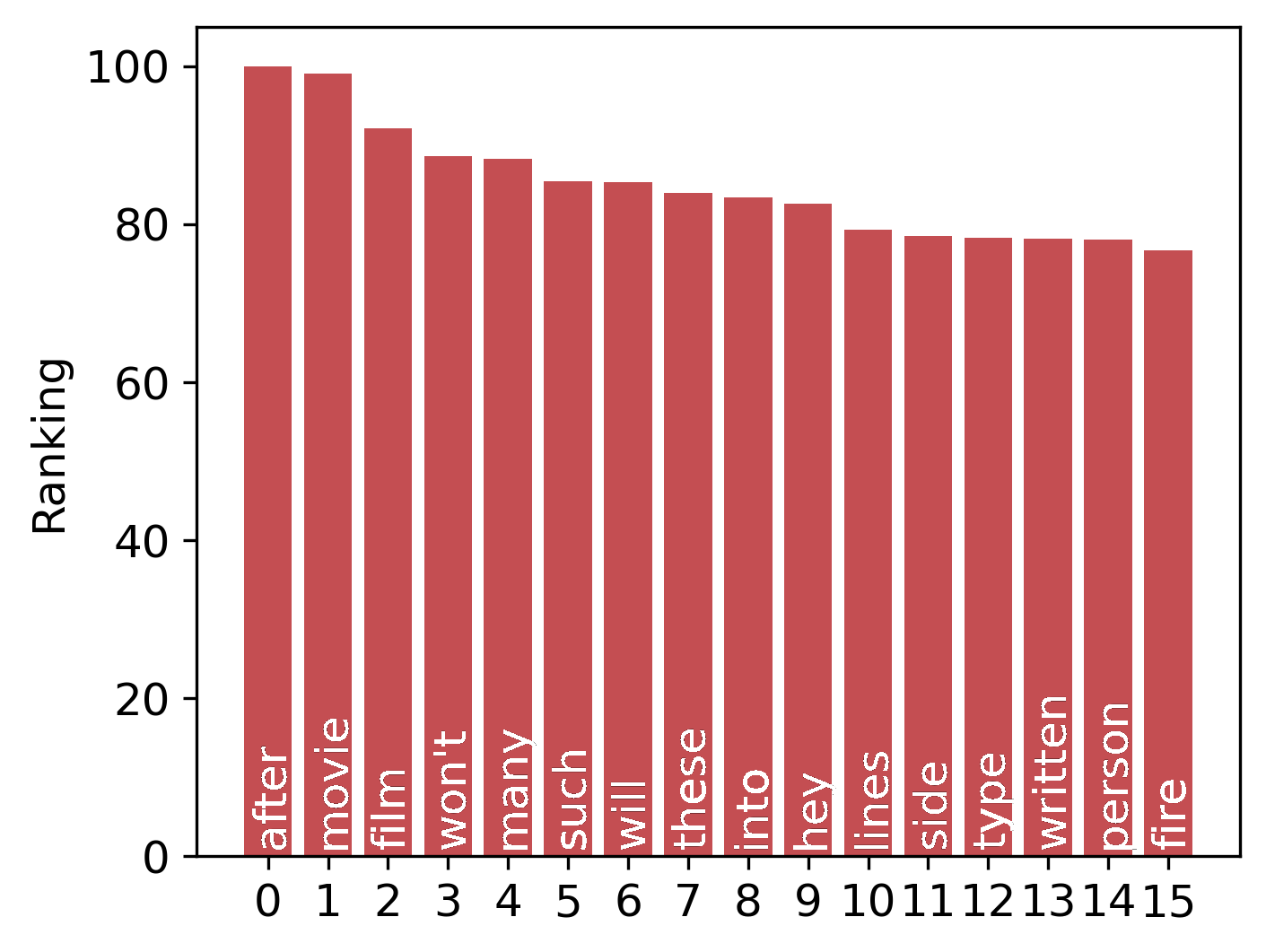}}
\subfigure[]{\includegraphics[width=0.24\textwidth]{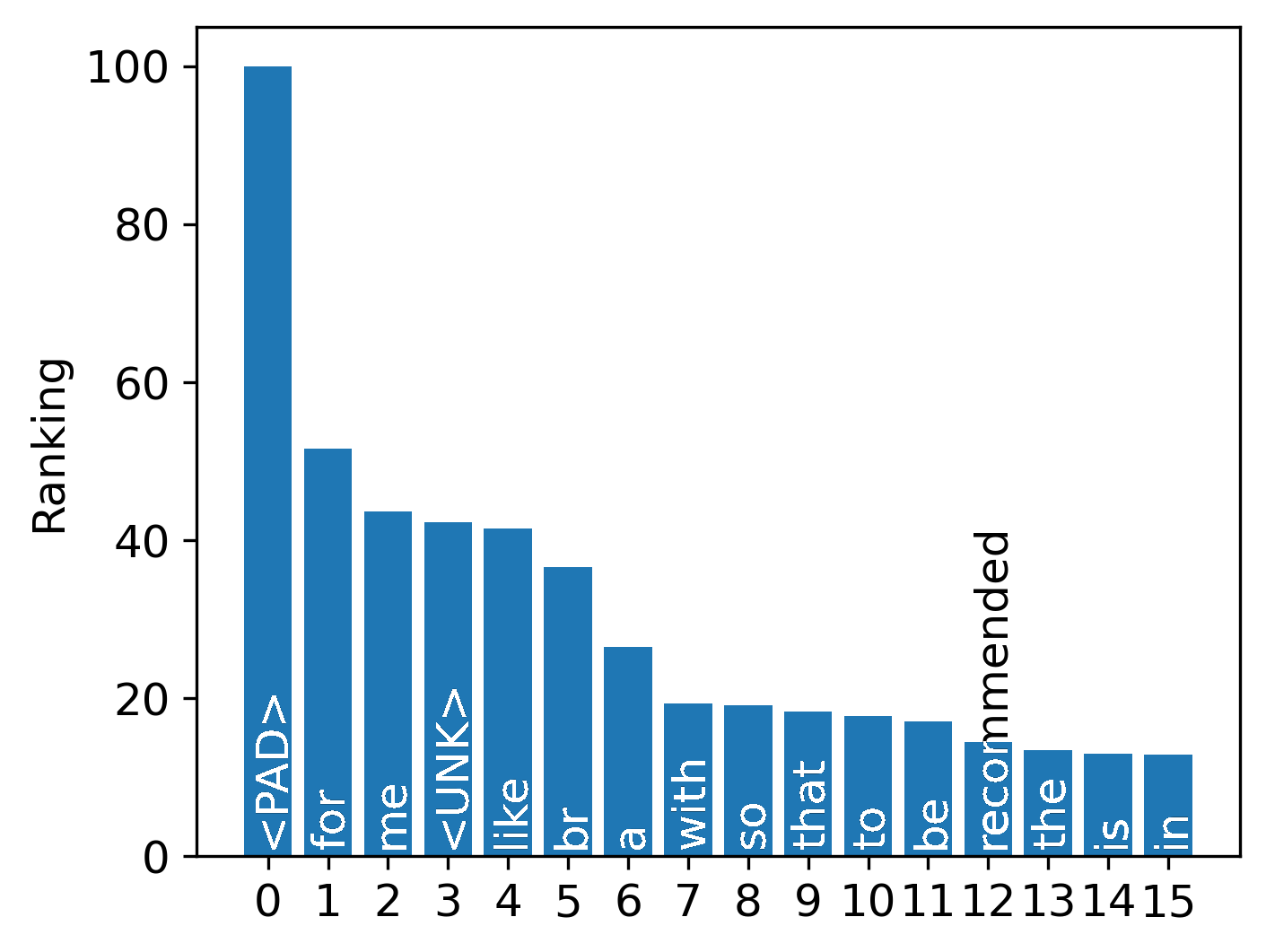}}
\subfigure[]{\includegraphics[width=0.24\textwidth]{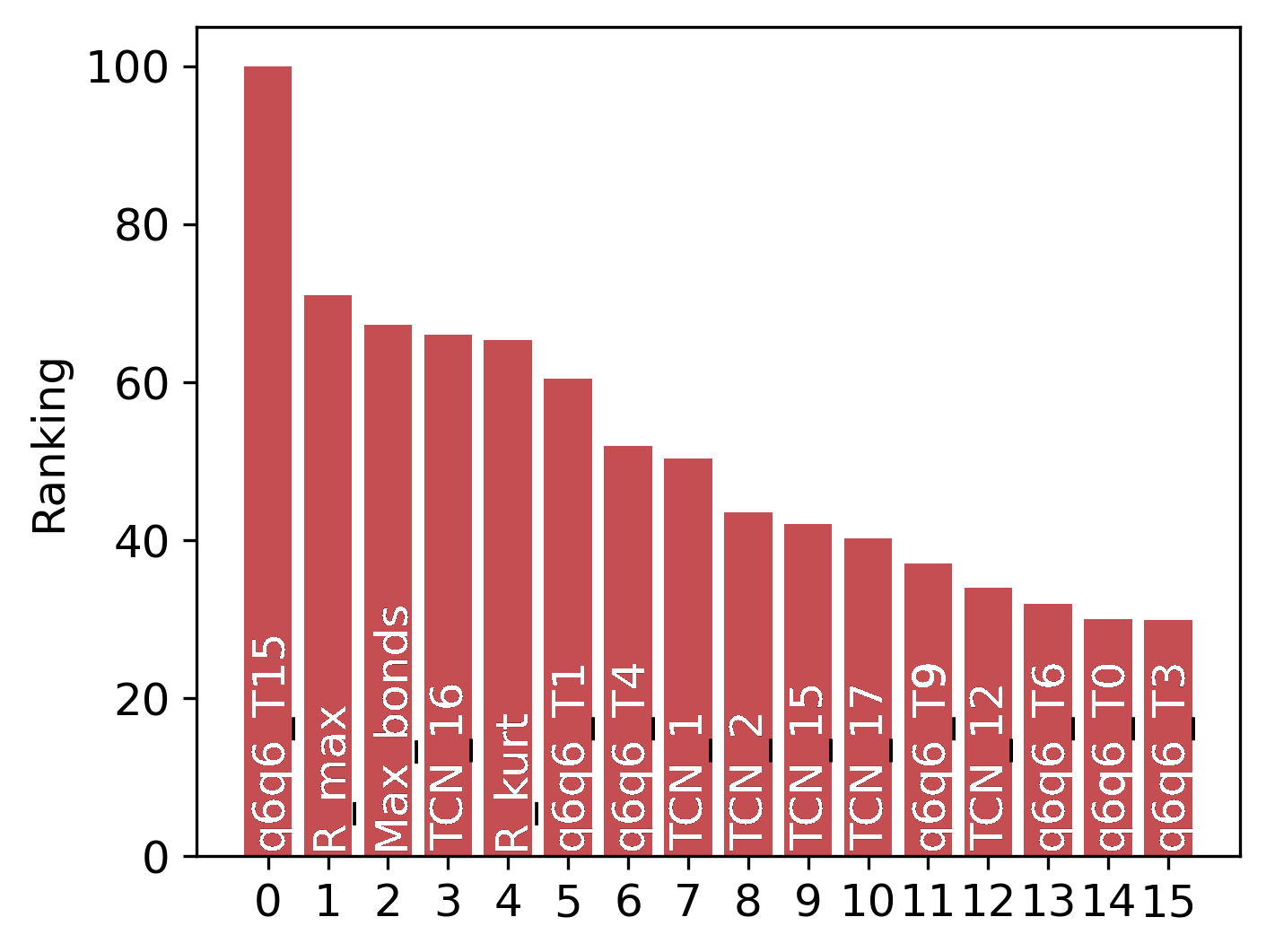}}
\subfigure[]{\includegraphics[width=0.24\textwidth]{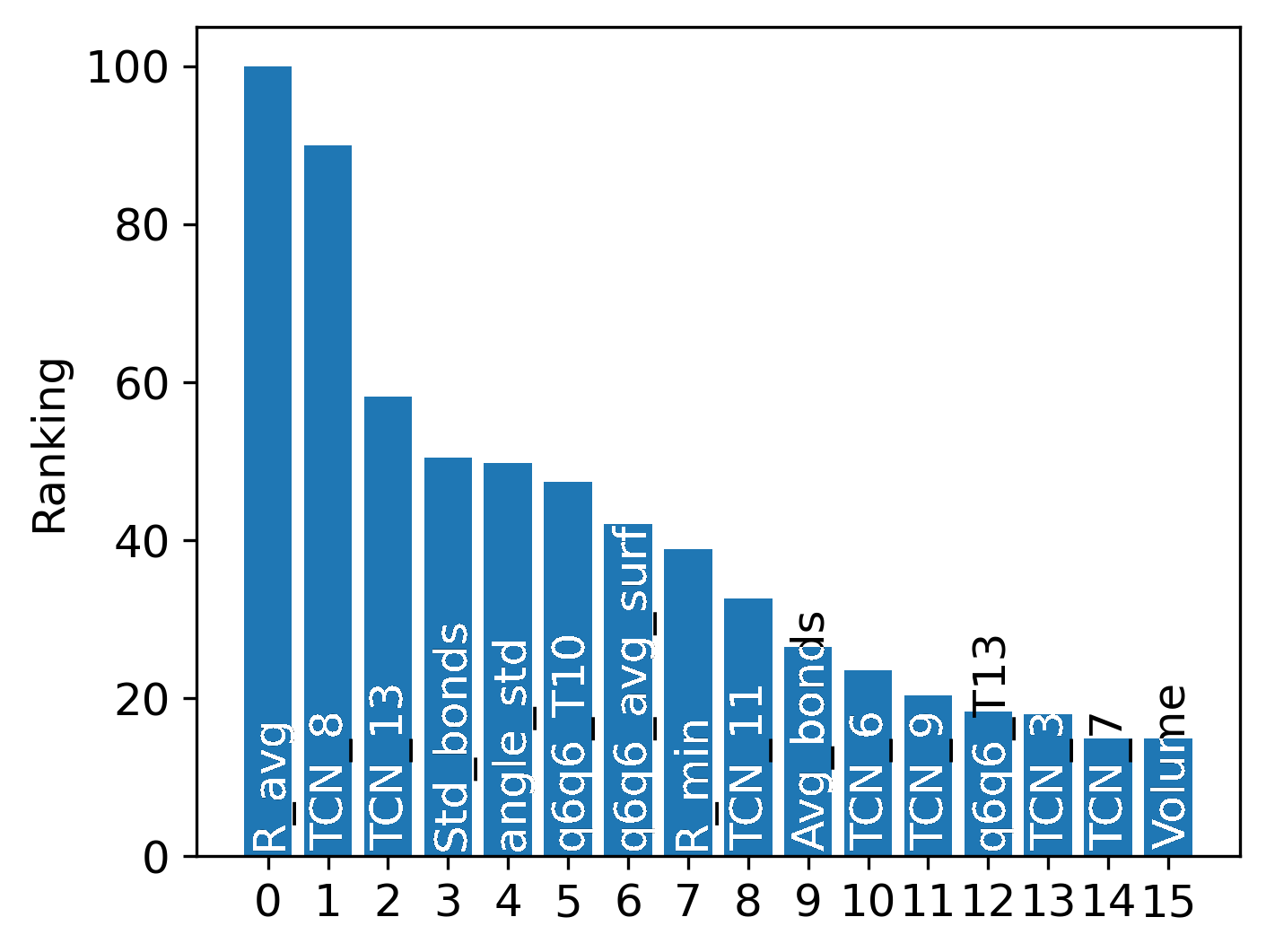}}
\subfigure[]{\includegraphics[width=0.24\textwidth]{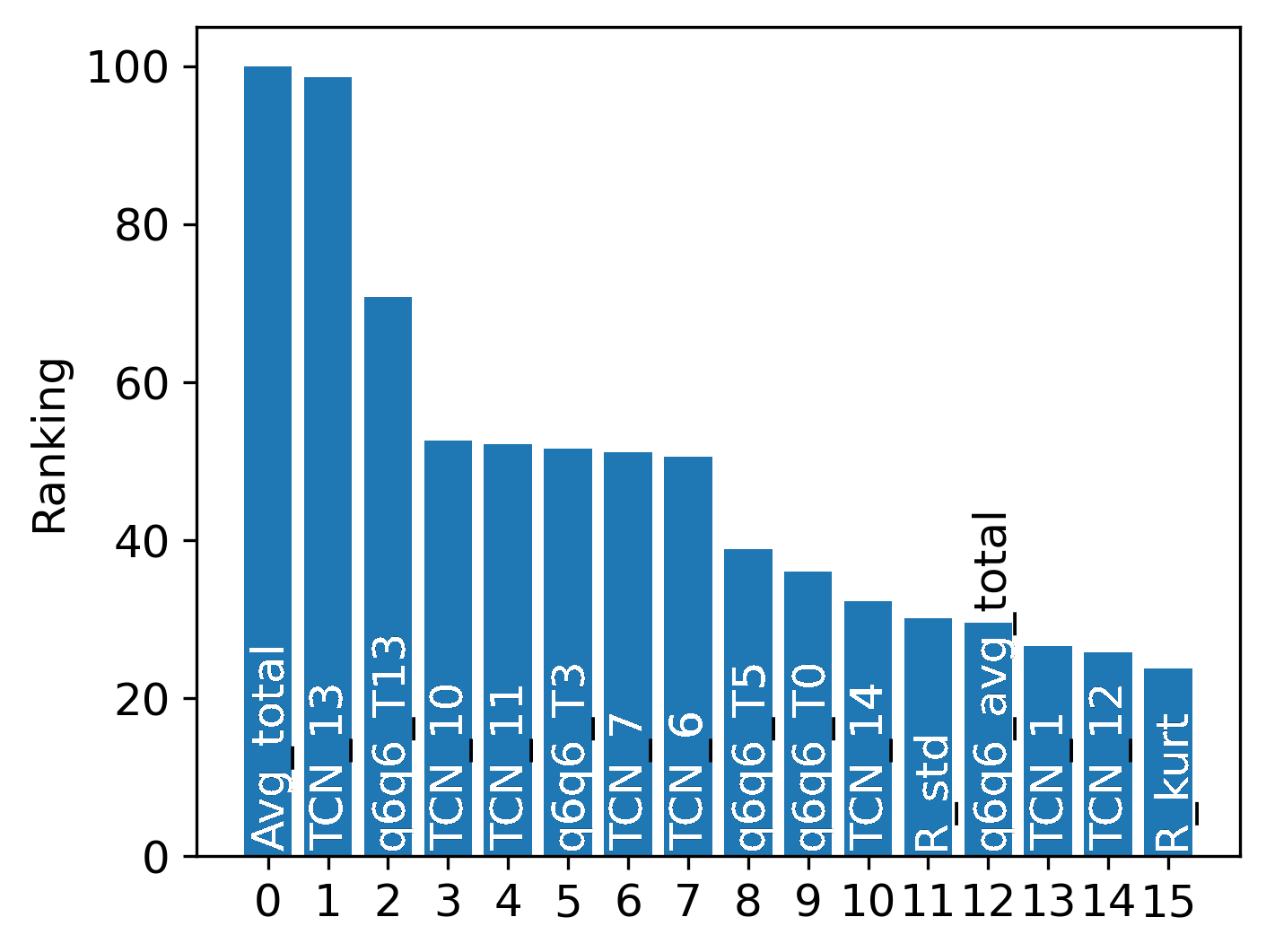}}

\caption{Feature unimportance and importance maps of words (a-e) and Gold dataset (f-h) generated from different methods and models. IMDB dataset: (a) Permutation method, (b) RF, (c) Contribution factor, (d) VTF, (e) RVTW. Gold dataset: (f) Variance Tolerance Factor, (g) RVTW and (h) CF.}
\label{fig:RNN_results}
\end{figure*}

The feature model was constructed based on the outputs of the base model, however, the length of sentence is fixed and a sentence can't contain all words. To update the weight of each word in each iteration and rank all words in a library, the mask layer is located between a one-hot layer and an LSTM layer. The feature model was retrained 2000 times since 2000 features were needed to demonstrate the CF method. The feature model also achieves the similar loss with 88\% accuracy. The feature importance is ranked and visualised in Fig.\ref{fig:RNN_results}(a-e).

As mentioned above, due to the large scale of the word bank, we constructed an equation system with 2000 unknowns and 2000 equations to demonstrate the success in RNN. The contribution of top 30 words are ranked and shown in Fig.\ref{fig:RNN_results}(c), where is impossible to present all words.

Considering the computing complexity, only permutation method and RF are used as a benchmark to evaluate the performance of the VTF-based methods applied to an RNN. To fit all sentences and rank their feature importance, all sentences are encoded using one-hot format and fitted using a RF regressor based on Gini impurity. The accuracy is 82.94 \% results and importances are illustrated in Fig.\ref{fig:RNN_results}(a). By permuting each word in the library in each sentence, the feature importance profile can be obtained according to their loss change, shown in Fig.\ref{fig:RNN_results}(b).

Semantic analysis using RNN is an example showing the feasibility and flexibility of the framework, but when the corpus (each representing a feature) is large, it becomes impractical. Unimportant feature profile from the VTF treats most of the unimportant features equally, so the CF and RVTW capture the most frequently occurring words, like the permutation method, even if some of them are uninformative.

\begin{table}[h!]
\caption{Comparison of feature selection methods in computational time and performance}
\small
\centering
\begin{tabular}{llll}
\hline 
 \multicolumn{4}{c}{Linear regression} \\ \hline 
 & Our method & RFE & Base model \\
Time complexity & $O(n)$ & $O(n^2)$ & - \\
MSE & 3.4014 & 3.4097 & 3.4003 \\ \hline

 \multicolumn{4}{c}{Logistic regression}  \\ \hline
Time complexity & $O(n)$ & $O(n^2)$ & - \\ 
Loss & 0.0866 & 0.0834 & 0.0812 \\ \hline
 \multicolumn{4}{c}{Convolutional Neural Network} \\ \hline
Time complexity & $O(n)$ &  $O(n^2)$ & - \\
Accuracy & 99.76\% & - & 100\% \\ \hline

 \multicolumn{4}{c}{Recurrent Neural Network}  \\ \hline
Time complexity & $O(n)$ &  $O(n^2)$ & - \\
Accuracy & 87.96\% & - & 88\% \\ \hline 
\end{tabular}
\label{tb:app_model_feature_selection}
\end{table}

\subsection{Real-world Applications}
In this study, we implemented our proposed framework on real-world datasets, tackling two difficult tasks. The first task involved predicting the formation of highly disordered nanoparticles based on their physical properties, while the second task involved predicting the $IGC_{50}$ chemical property from aromatic compounds \cite{karim2021quantitative}. Further information regarding the datasets and experimental procedures are provided in the subsequent sections.

\subsubsection{Gold Nanoparticles}
 The dataset used for this study contains 4000 gold (Au) nanoparticles generated using molecular dynamics at various temperatures, growth rates and times, which is available online ~\cite{goldset}. Atomistic, molecular, topological and crystallographic descriptors are characterised as features and the formation energies are assigned as the target property labels. Based on different types of information about the nanoparticles the features are grouped into 5 different feature sets, referred as Bulk (B), Surface (S), Totals (T), Condensed (C) and Expanded (E). A comprehensive evaluation of different machine learning models on these feature sets has been previously reported \cite{barnard2020selecting}. In the present study T feature set is used to train a MLP and all 5 feature sets are used to train individually linear regressors. The model structures are listed in Supporting Information Table.9. A total of 5 linear regressors reproduced the reported LR results \cite{barnard2020selecting} in Supporting Information Table.1, while MLP achieved better performance in T set, with a MAE of 0.0282.

According to the number of features, the feature model is retrained from 35 to 150 times. The feature importance is ranked based on the VTF, RVTW and CF, and the feature unimportance and importance maps for the MLP are visualised in Fig.\ref{fig:RNN_results} (f-h). Feature importance profiles extracted from 5 linear regressors are provided in Supporting Information Fig.8.
We can observe our histograms differs from the reported Extra-trees outcomes \cite{barnard2020selecting}, which is to be expected. 
Different machine learning models trained on same task lead to different rankings.

\subsubsection{Chemical Toxicity}
The $IGC_{50}$ dataset shows the 50\% growth inhibitory concentration of the Tetrahymena pyriformis organism (a protozoan ciliate) when exposed for 40 hours and contains 1793 aromatic compounds. The SMILES2vec model \cite{goh2017deep} is a mixed CNN-GRU architecture that automatically learns features from SMILES strings \cite{weininger1988smiles} to predict chemical properties \cite{goh2017smiles2vec}. The main architecture consists of an embedding layer, convolutional layer with RELU activation and Gated Recurrent Units (GRU). The model is trained for 200 epochs and achieved a faithful validation MSE of 0.31 based on the previous report \cite{karim2021quantitative}. The training setting can be found in the Appendix. The explanation mask takes the embedded SMILES representation as input and generates a character-level importance profile. The feature model was trained and optimised until the MSE achieved the same number as base model, or slightly better.

\begin{figure}[]
\centering
\includegraphics[width=0.4\textwidth]{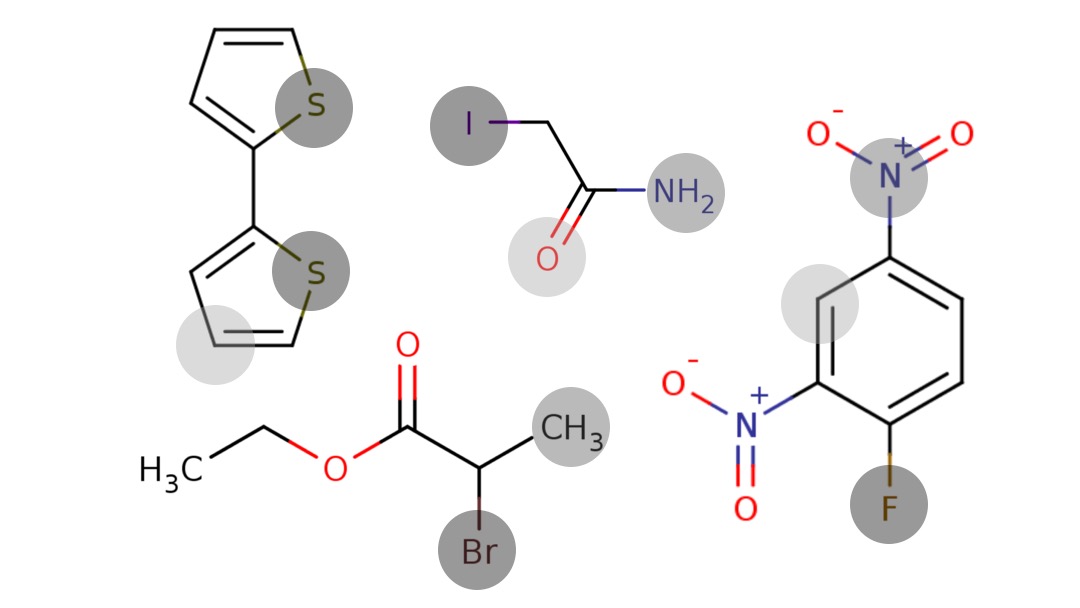}
\caption{Atom feature importance extracted from RVTW and CF, colored circles of increasing darkness indicate increasing importance.}
\label{fig:smiles_results}
\end{figure}

To test the model interpretability, we calculated RVTW and CF in this application. Some SMILES samples are converted to structures for visualizing. In Fig \ref{fig:smiles_results}, we highlight the important character based on the extracted feature importance profile, darker circles indicating more importance from the mask. From the result, we can observe that the model is more sensitive to atoms from hydrophilic groups, such as F, I, Br and S, deeming them more important. Considering the fact that many halogenated compounds are toxic, these atoms play a critical role in such compounds. The results from data-driven are consistent domain knowledge.

\section{Conclusion}

In general, the aim of present work is to provide human-readable interpretability to black box models in the Rashomon set by measuring the tolerance of the model to the change of features. 
The present approach is motivated by linear regression \cite{goodfellow2016machine} and Shapley values, and the way to accomplish the idea comes from transformer \cite{ji2021show}, fixing the parameters from the base model and training a feature model. Weights of the mask layer on feature model are completely data-driven feature importances and we proposed two methods based on VTF, RVTW and CF, to endow interpretability at a global level in the Rashomon set, which can be combined with human understanding. More importantly, other possibilities for interpretability still exist. The universality of this approach is demonstrated by deploying in linear regressor, logistic regressor for general cases; CNN for image-related tasks; RNN for semantic analysis and two real-world applications are provided to show the importance and usage of interpretability for physicochemical systems.

Our method is fast and flexible in model selection, feature selection and feature representation, such as visualising individual words and image pixels in the experiments. It can easily make use of any advanced neural networks to be our component-base model in the framework by separating interpretability from models \cite{ribeiro2016should}, meaning it is not limited to certain models that might lose predictive performance. The model is applicable to all neural networks, irrespective of their performance, however, neural networks with poor performance hold no value.
In the language processing case, the random forest classifies the words with accuracy 82.94\%, while RNN can achieve 88\% accuracy. The feature unimportance profile is not exactly the reverse of important feature profile extracted from VTF, same features ranked at different positions, like \texttt{AGE} and \texttt{NOX} in Fig.\ref{fig:linear_base_model_hist_TV}. With more factors taken into consideration, the outcome of the RVFW ranking is different from other feature importance profiles, but top important features are always selected, although their positions vary as different approaches being applied. It is not fair to judge the performance of a ranking over another, as the interpretability is subjective, but if a feature is selected as important by different approaches, it informs the importance of that feature. The performance of feature selection methods based on the result from Fig.\ref{fig:baseline_results} demonstrate the reliability and accuracy of our methods.
CF provides more detailed interpretability for machine learning models by calculating contribution of each feature through solving equation systems. In addition to ranking the feature importance profile, the method also calculates specific contribution values for features. More importantly, the method opens a new field for interpretable methods for further investigation and the potential for directly measuring the impact of "domain-driven" interventions by researchers.

Several comparative and classic methods are used to provide insights from different perspectives, including model-agnostic methods and model-specific methods: connection weights, random forest, KernelShap, permutation method, AFS, ReliefF and Fisher Score. Olden \cite{olden2004accurate} found that the connection weight approach achieves the best performance for accurately quantifying variable importance among 8 approaches. The performance of the connection weight approach highly depends on the initial weights used for starting the training process \cite{ibrahim2013comparison}, whereas the retraining strategy in our method mitigates the dependence on initial weights. Tree-based methods, such as random forests, are interpretable models but fail to deal with linear relationships and are impossible to extend to neural networks, while our new method is extensible to all machine learning methods and data types. Unlike permuting instances for a feature, all instances are perturbed in a constant way to compare the effect of a feature. The two well-accepted ReliefF and Fisher Score have comparably stable performance, but not accurate as RVFW. AFS performed poorly in MNIST setting and the reason might be the choice of CNN.

There are other machine learning models that can theoretically be used to extend to this work, such as support vector machines or Gaussian process models that are widely used in science and technology, and even unsupervised models. Further application to complex models and huge datasets would help address the scalability, but the above experiments demonstrate the general feasibility and opportunities.  As the number of features and model layers increase, it is still feasible to re-train the feature models as suggested, but the computational cost rises correspondingly. Mathematically, the time complexity of a forward pass of a trained MLP depends on the input, number of layers and the size of each layer, and the feature model is trainable as long as the base model can be trained. Discovering the specific relationship between the size, complexity and feature model training will be the topic of future work.

\bibliographystyle{IEEEtran}
\bibliography{ref}

\end{document}

\title{ \bf SUPPORTING INFORMATION \\}
%\author[1]{Sichao Li}
%\author[1]{Amanda S. Barnard\footnote{Corresponding author: amanda.s.barnard@anu.edu.au}}
%\affil[1]{School of Computing, Australian National University, Acton 2601, Australia} 
%\email[4] {Amanda.S.Barnard@anu.edu.au}
%\date{}
%\setcounter{Maxaffil}{0}
%\renewcommand\Affilfont{\itshape\small}
\maketitle

\subsection*{Proof of learning algorithm}
\label{sec:proof}
If we add an extra mask layer in the base model, the feedforward propagation becomes:
\begin{eqnarray}
\label{eqn:coefficient_feedforward}
a_j^{l+1} = a_j^lw_j^{l, l+1},&\mathrm{if}\quad l=0 \\
a_j^{l+1} = \sigma (\sum_i a_i^lw_{ij}^{l, l+1} + b_j^{l+1}),&\mathrm{if}\quad l>0
\end{eqnarray}
where $w_{ij}^{k}$ is weight for node $j$ in layer $l_{k}$ for incoming node $i$; $\sigma$ is the activation function; $b_{j}^{k}$ is bias for node $j$ in layer $l_{k}$ and $a_{j}^{l}$ is product sum plus bias (activation) $j$ in layer $l_{k}$.

An error function $E(X, \theta)$ defines error between the desired output $y$ and the calculated output $\hat{y}$ of the neural network with the parameter $\theta$ with respect to the weights $w_{ij}^{k}$ and biases $b_{j}^{k}$. In the training process, weights are updated according to the gradient descent of each iteration. Based on the learning rate $\eta$, each iteration of gradient descent updates as:
 \begin{equation}
\theta^{t+1} = \theta^{t} - \eta \frac{\partial E(X, \theta^{t})}{\partial \theta}
\label{eqn:coefficient_update}
\end{equation}
 where $\theta^{t}$ is parameters of the network at iteration $t$. To illustrate the calculation details of the model with a mask layer the mean squared error function is selected, defined as
 \begin{equation}
E(X, \theta) = \frac{1}{2N}\sum_{i=1}^{N}(\hat{y}-y)^{2}
\label{eqn:coefficient_error_func}
\end{equation}

According to the backpropagation algorithm and chain rule, the derivation of gradients is straightforward. By deriving $ \frac{\partial E(X, \theta)}{\partial w_{ij}^{k}} = \frac{1}{N}\sum_{d=1}^{N}\frac{\partial E_{d}}{\partial w_{ij}^{k}}$,
the general form for all input in $X$ can be summarized by adding all individual gradients. The individual gradient with respect to a general weight is $\frac{\partial E_{d}}{\partial w_{ij}^{k}}$, where the chain rule applies.
The error function partial derivative becomes $\frac{\partial E_{d}}{\partial \alpha _{j}^{k}}\frac{\partial \alpha _{j}^{k}}{\partial w_{ij}^{k}}$. The first term is usually called the error, denoted $\delta _{j}^{k}$, and the second term can be simplified as $o _{i}^{k-1}$. The first term can be transferred to the following equation by applying the chain rule:
 \begin{equation}
\delta _{j}^{k} = \frac{\partial E_{d}}{\partial \alpha _{j}^{k}}= \sum_{l=1}^{r^{k+1}}\frac{\partial E_{d}}{\partial \alpha _{l}^{k+1}}\frac{\partial \alpha _{l}^{k+1}}{\partial \alpha _{j}^{k}}=\sum_{l=1}^{r^{k+1}}\delta _{l}^{k+1}\frac{\partial \alpha _{l}^{k+1}}{\partial \alpha _{j}^{k}}
\label{eqn:coefficient_error_term}
\end{equation}
where $r^{k+1}$ is the number of nodes in the next layer. Because all the weights (except for those between the input layer and the mask layer) are not trainable, they can be considered as constants.

According to the linear relationship shown in Fig. \ref{eqn:coefficient_feedforward}, the partial derivative between the input layer and the mask layer can be derived as:
 \begin{equation}
 \small
\frac{\partial E}{\partial \alpha _{j}^{1}} = \delta _{j}^{1} o_{i}^{0} \\
\delta _{j}^{1} = \sum_{l=1}^{r^{2}}\delta _{j}^{2} \frac{\partial \alpha _{l}^{2}}{\partial \alpha _{j}^{1}} = \sum_{l=1}^{r^{2}}\delta _{j}^{2} w_{jl}^{2} \\
\frac{\partial E}{\partial \alpha _{j}^{1}} = o_{i}^{0} \sum_{l=1}^{r^{2}}\delta _{j}^{2} w_{jl}^{2}
\label{eqn:coefficient_mask_partial}
\end{equation}
where $o_{i}^{0}$ is input data directly from $X$. In this equation Fig. \ref{eqn:coefficient_mask_partial}, we can observe that $o_{i}^{0}$ and $w_{jl}^{2}$ are constant and the only variable is $\delta _{j}^{2}$. If we trace this down to the final layer $\delta _{j}^{m}$, in which $m$ is the final layer and the formula defined as $(\hat{y} - y)g{}'(\alpha _{j}^{m})$, where $g$ is the activation function, the only factor that affects the weights between the input layer and the mask layer is the prediction error. Thus, if a feature is important to the outcome, causing the substantial change in prediction error, the coefficient (weight in the model) of the feature would be updated accordingly. However, the coefficient (weight in the model) of unimportant features will not deviate significantly from the initialization value. Overall, the $\theta$ accumulates the effect of all connection weights across all hidden neurons.

The effect of the prediction error $(\hat{y} - y)$ is same in each iteration to all neurons in the mask layer and the factor that controls the $\theta$ is the weight $w_{jl}^{2}$ in Fig. \ref{eqn:coefficient_mask_partial} obtained from the base model. Considering that the weight $w_{jl}^{2}$ is calculated based on all connection weights between its connected neurons in hidden layers, the weights imply the importance of the variance.

\subsection*{Derivation of Contribution Factors} \label{sec:CF}
Considering the contribution of each feature in the feature set and the linear relationship between mask and feature set, we can derive $\mathbb{C}(Q, (X_{0}, X_{1}, X_{2},\cdots, X_{i})) = \\ \mathbb{C}(Q, (w_{X_{0}} {0}X_{0},  w_{X_{1}}X_{1}, w_{X_{2}}X_{2},\cdots,  w_{X_{i}}X_{i}))$. As the model is fixed, the function can be simplified into:
\begin{equation}
\sum_{i=0}^{N}\mathbb{C}(Q,X_{i}) = \sum_{i=0}^{N}\mathbb{C}(Q,w_{X_{i}}X_{i})
\label{eqn:additivity_extension}
\end{equation}

To solve the equation with unknowns, the exact corresponding number of equations have to be provided, offered by retraining the system. The $w_{X_{i}}$ in the function also needs to be moved outside the parentheses, where the $w_{X_{i}}$ represents the relationship between the weight and the contribution.

\subsubsection*{Linear regression case explanation}
Thinking of the most straightforward model, a linear regressor, the coefficient of a feature is linear to its contribution, as increasing the weight by one unit changes the estimated outcome by 100\%, which doubles the contribution of the feature. So the following function can be derived: $\mathbb{C}(X_{1}) + \mathbb{C}(X_{2}) +\cdots+ \mathbb{C}(X_{i}) = w_{X_{1}}\mathbb{C}(X_{1}) +  w_{X_{2}}\mathbb{C}(X_{2}) +\cdots+  w_{X_{i}}\mathbb{C}(X_{i})$ from Eqn. \ref{eqn:additivity_extension}. If the equation system can be solved, then the contribution of each feature to the model performance can be decided.

Fortunately, the previous retrain strategy provides a chance out! Given the number of unknown parameters, we can solve the equation system using the exact corresponding number of equations, offered by retraining the system. The following equation system can be solved easily. By setting the output of each equation to 1, the relationship among all features can be derived. The system can be solved using its associated augmented matrix, as it is empirically more suitable for computer manipulations: $(A|I) = \left( \begin{smallmatrix} 1&1&\cdots&1&|&1 \\ w_{X_{1}}&w_{X_{2}}&\cdots&w_{X_{i}}& | & 1 \\ w'_{X_{1}}&w'_{X_{2}}&\cdots&w'_{X_{i}}&|&1 \end{smallmatrix} \right)$,
with: 
\begin{equation}
\left\{\begin{matrix}
\mathbb{C}(X_{1}) + \mathbb{C}(X_{2}) +\cdots+ \mathbb{C}(X_{i}) = 1\\
w_{X_{1}}\mathbb{C}(X_{1}) + w_{X_{2}}\mathbb{C}(X_{2}) +\cdots+  w_{X_{i}}\mathbb{C}(X_{i}) = 1\\
\vdots
 \\
w\prime_{X_{1}}\mathbb{C}(X_{1}) +  w\prime_{X_{2}}\mathbb{C}(X_{2}) +\cdots+  w\prime_{X_{i}}\mathbb{C}(X_{i}) = 1
\end{matrix}\right.
\label{eqn:linear_system}
\end{equation}

\subsubsection*{Introduction and derive of $\mu$} However, in practice the model usually is much more complex and the factor $w_{X_{i}}$ can not be applied to the contribution directly. To deal with the coefficient in general cases, the model can be run with different input settings to determine the relationship between function $\mathbb(\cdot)$ and $w$ by $\mu$ experimentally.

For a specific coefficient $w_{X_{0}}$ of a feature $X_{0}$, all other coefficients are fixed except for the one that is being calculated and we compare the performance between the model with input $\chi (X_{0}, X_{1}, X_{2},\cdots, X_{i})$ and
$\chi (w_{X_{0}}X_{0}, X_{1}, X_{2},\cdots, X_{i})$ by multiplying a mask $(w_{X_{0}}, 1,\cdots, 1) $ to $\chi$. The change in $\Delta P$ between these two outcomes $\Delta P_{partial} = \mathbb{P}(\mathbb{Q}(X_{0}, X_{1},\cdots, X_{i})) - \mathbb{P}(\mathbb{Q}(w_{X_{0}}X_{0}, X_{1},\cdots, X_{i}))$ is from $X_{0} \mapsto w_{X_{0}}X_{0}$ and represents the effect of part of the $X_{0}$. The total contribution of feature $X_{0}$ can be found by setting $\Delta P_{total} = \mathbb{P}(\mathbb{Q}(X_{0}, X_{1},\cdots, X_{i})) - \mathbb{P}(\mathbb{Q}(0 \times X_{0}, X_{1},\cdots, X_{i}))$.

According to the $\Delta P_{partial}$ and $\Delta P_{total}$ above, the $\mathbb{C}(w_{X_{0}}X_{0})$ can be represented by $ \mu \times \mathbb{C}(X_{0})$, where the term $\mu$ is introduced to represent $\frac{\Delta P_{partial}}{\Delta P_{total}}$ for simplicity, and the simple linear case can be extended to the general case based on Fig. \ref{eqn:linear_system}. The system can then be solved with row operations from Gaussian elimination, including swapping rows positions, add one row onto another and multiplying a row by a non-zero scaler.
%In practice, it is likely that $\mathbb{E}_{aux}(Q, \theta_{1} X_{1})$ is close to 0, which results in $0 \times \mathbb{C}(X_{1}) = 0$ and thus the determinant of augmented matrix is 0, $det(A) = 0$. In terms of this case, the weight remains the same $\theta$.

Taking a linear regressor as an example, $Q = X_{1} + X_{2} + X_{3}$, given an input instance (0.1, 0.3, 0.6), the output is $1 = 0.1+0.3+0.6$, in which $X_{1}$(0.1), $X_{2}$(0.3) and $X_{3}$(0.6) can be seen providing 10\%, 30\% and 60\% contribution to the model $Q$. However, the contribution is not always obvious in real cases. In this simple case, to meet the output requirement, the following functions Fig. \ref{eqn:linear_system_case1} can be defined based on our method above and we can
obtain an augmented matrix coefficient matrix $(A|I) = \left( \begin{smallmatrix} 1&1&1&|&1 \\ 1&2&1/2& | & 1 \\ 4&1&1/2&|&1 \end{smallmatrix} \right)$. The matrix gives a relation between features $3 \times \mathbb{C}(X_{1}) = \mathbb{C}(X_{2}); 6 \times \mathbb{C}(X_{1}) = \mathbb{C}(X_{3})$, which satisfies the contribution distribution:
\begin{equation}
\left\{\begin{matrix}
\mathbb{C}(X_{1}) + \mathbb{C}(X_{2}) + \mathbb{C}(X_{3}) = 1\\
\mathbb{C}(X_{1}) +  2 \times \mathbb{C}(X_{2}) + 1/2 \times \mathbb{C}(X_{i}) = 1\\
4 \times \mathbb{C}(X_{1}) + \mathbb{C}(X_{2}) + 1/2 \times \mathbb{C}(X_{i}) = 1
\end{matrix}\right.
\label{eqn:linear_system_case1}
\end{equation}

%------------------------------------------------------------------------

\subsubsection*{Linear and Logistic Regression Supporting Information}

\begin{figure*}[]
\centering
\small
\includegraphics[width=.4\textwidth]{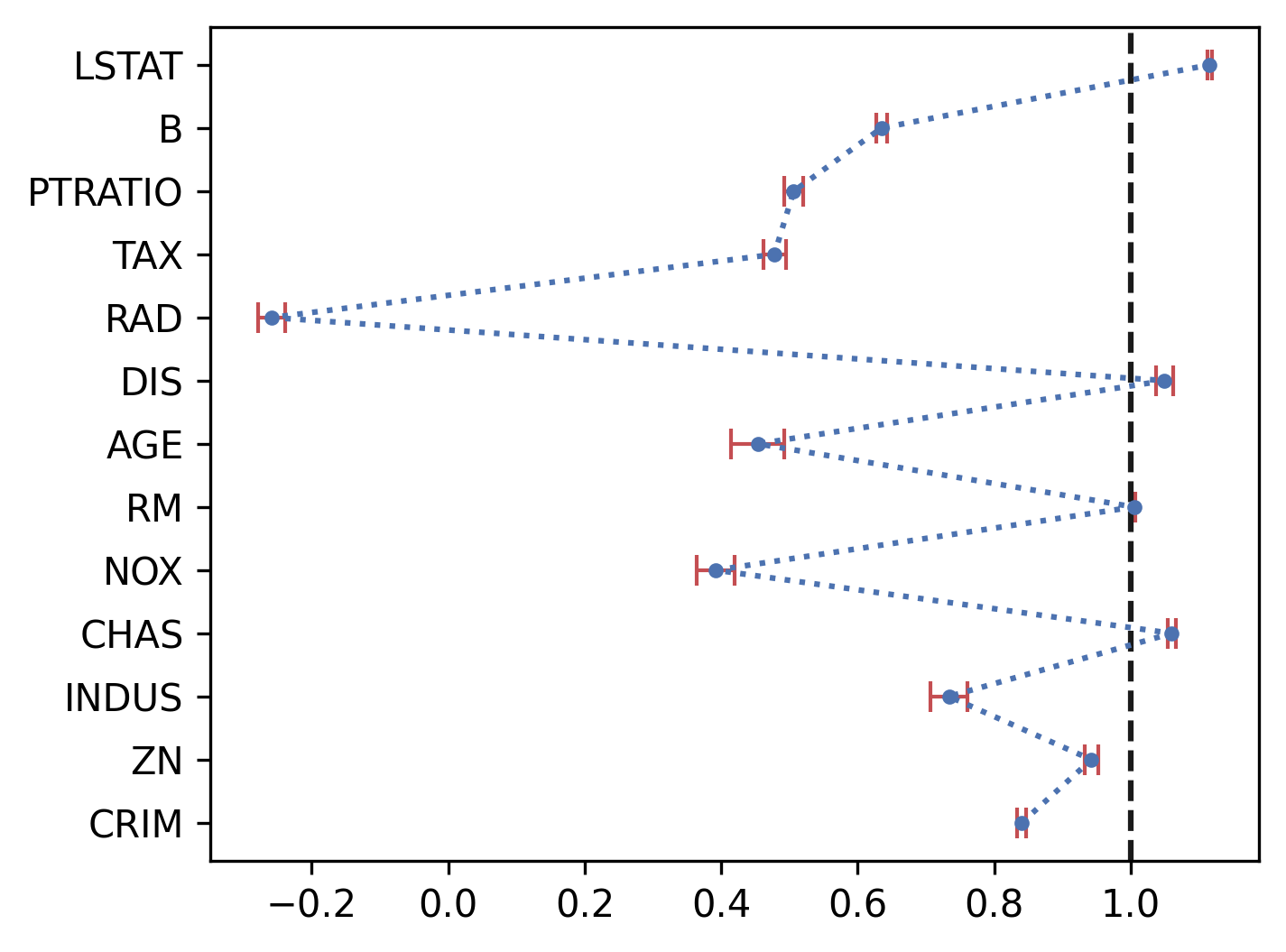}
\includegraphics[width=.485\textwidth]{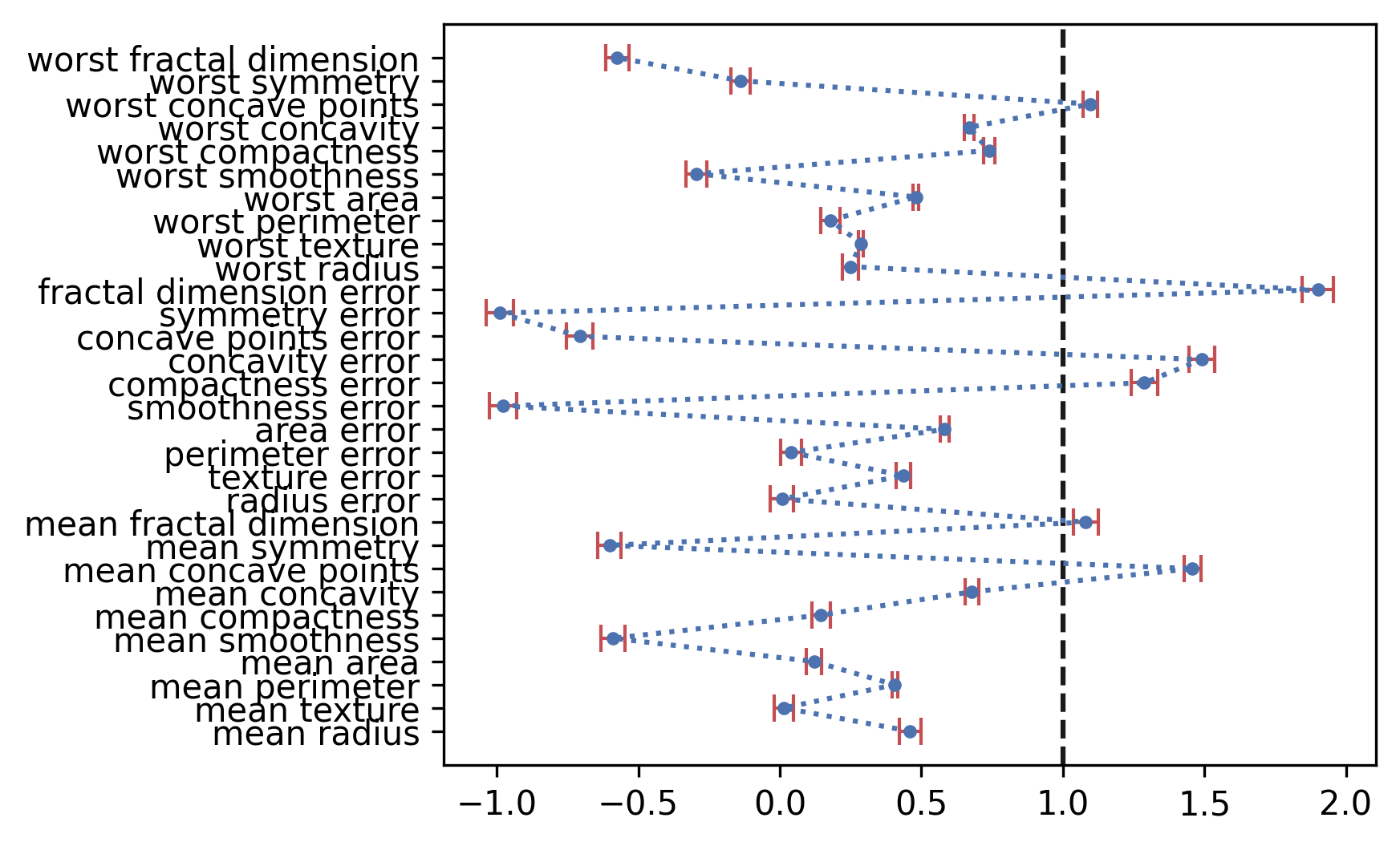}
\caption{The weights of linear regression and logistic regression are displayed in weight plot and the target 1 is marked as the line}
\label{fig:linear_base_model_weight_plot}
\end{figure*}

\begin{figure*}[]
\centering
\includegraphics[width=.4\textwidth]{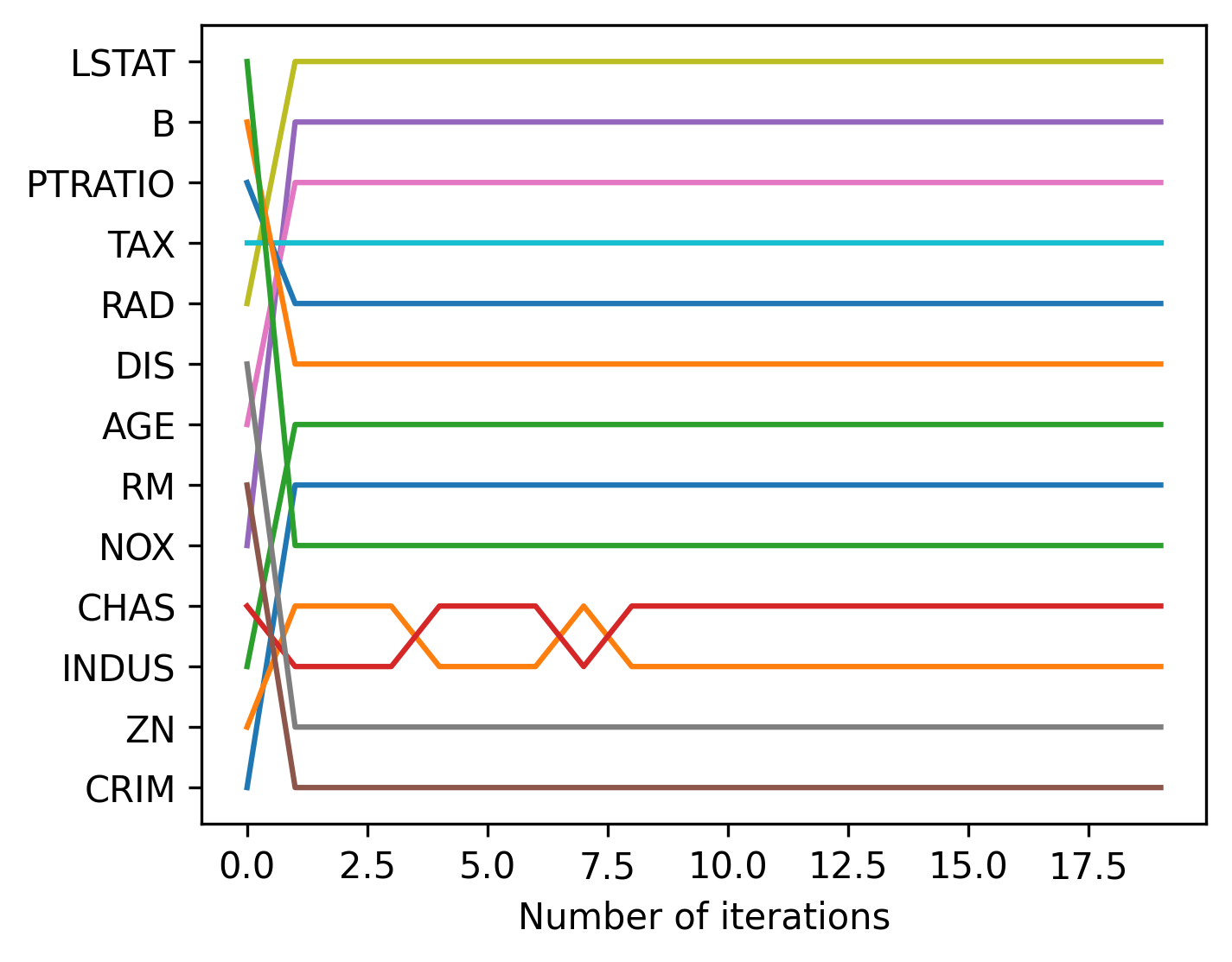}
\includegraphics[width=.485\textwidth]{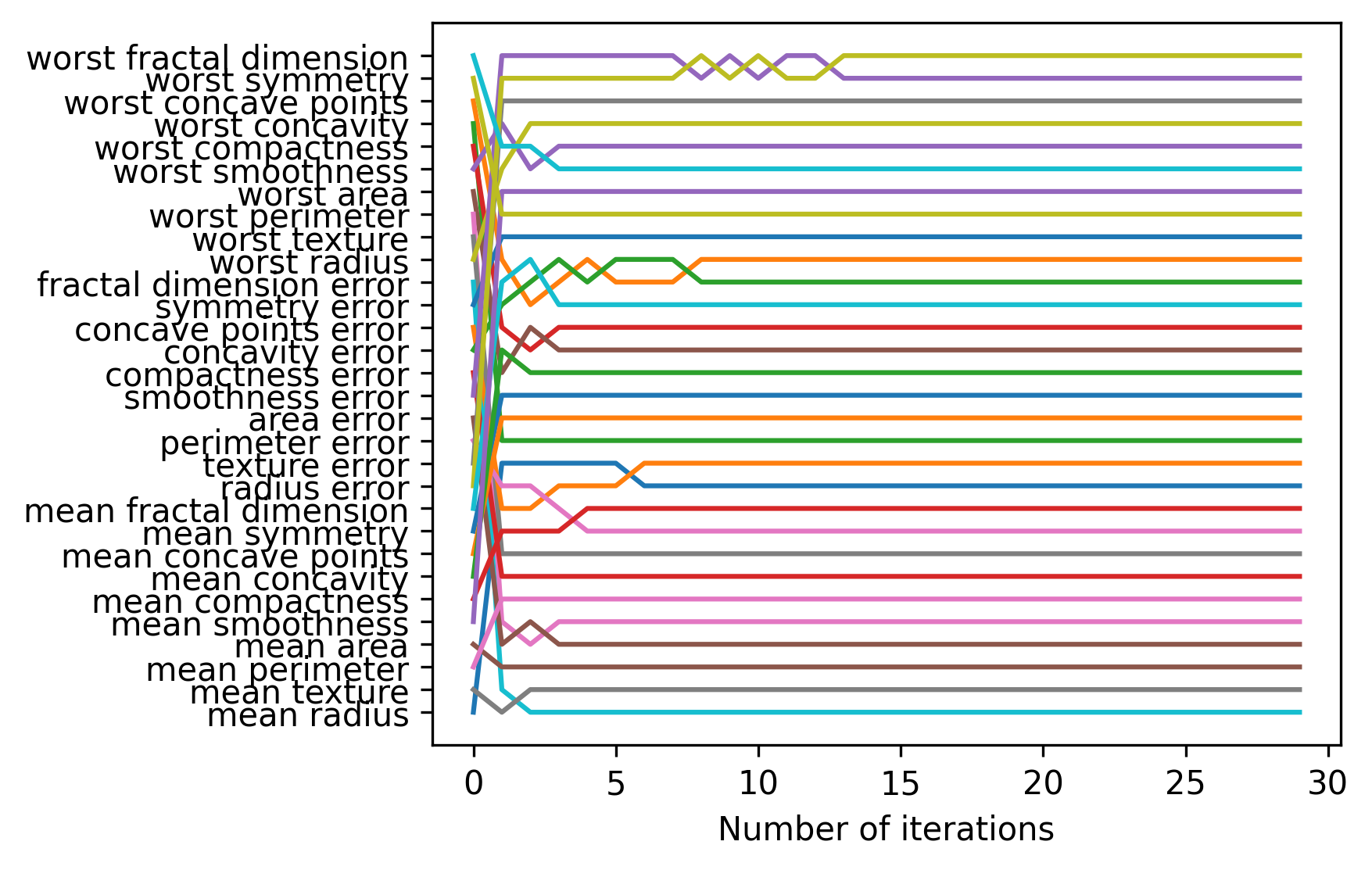}
\caption{Profile change of linear regression and logistic regression as the number of iterations increases.}
\label{fig:linear_base_model_converge}
\end{figure*}

\begin{figure*}[]
\centering
\includegraphics[width=.41\textwidth]{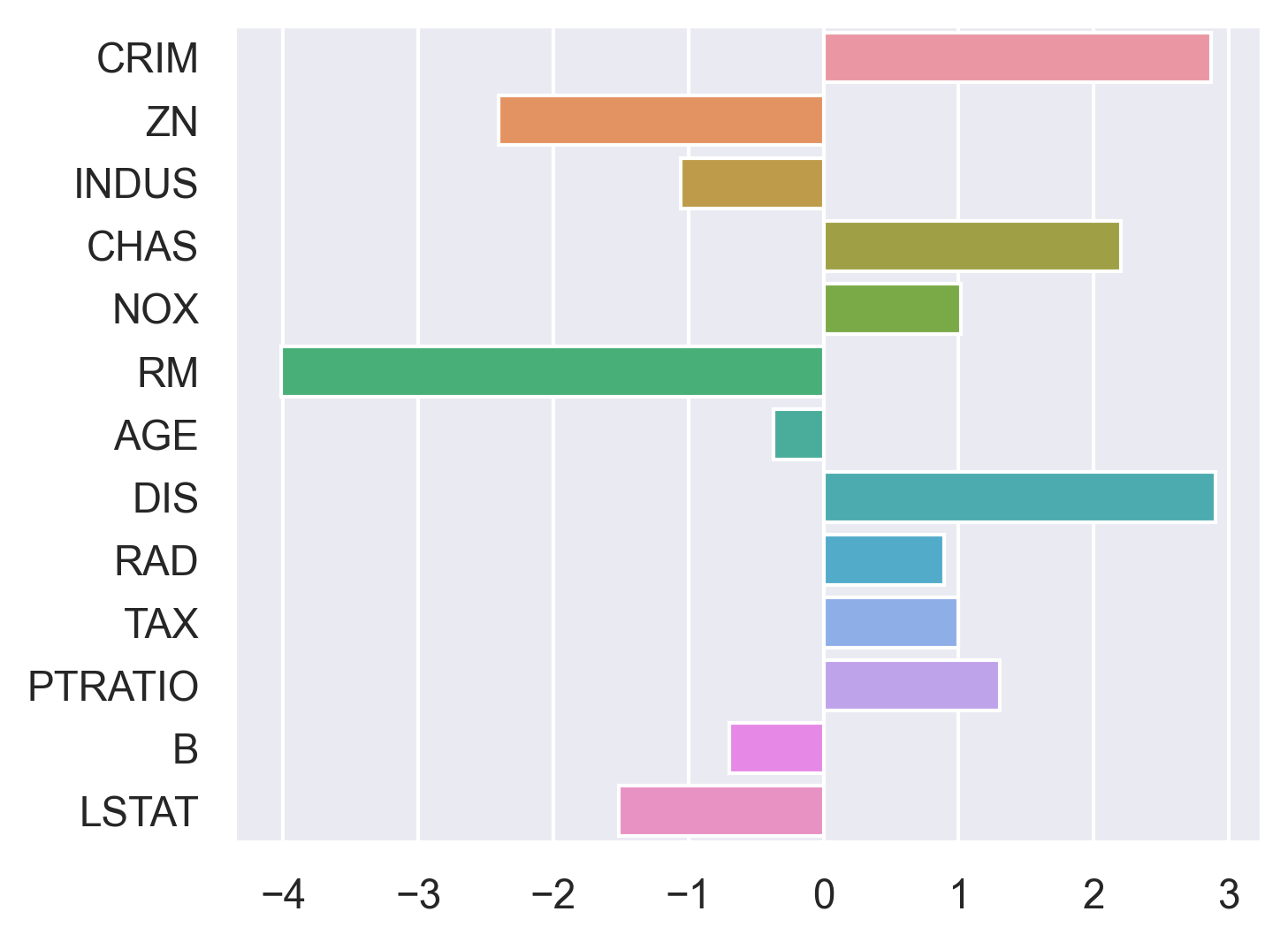}
\includegraphics[width=.485\textwidth]{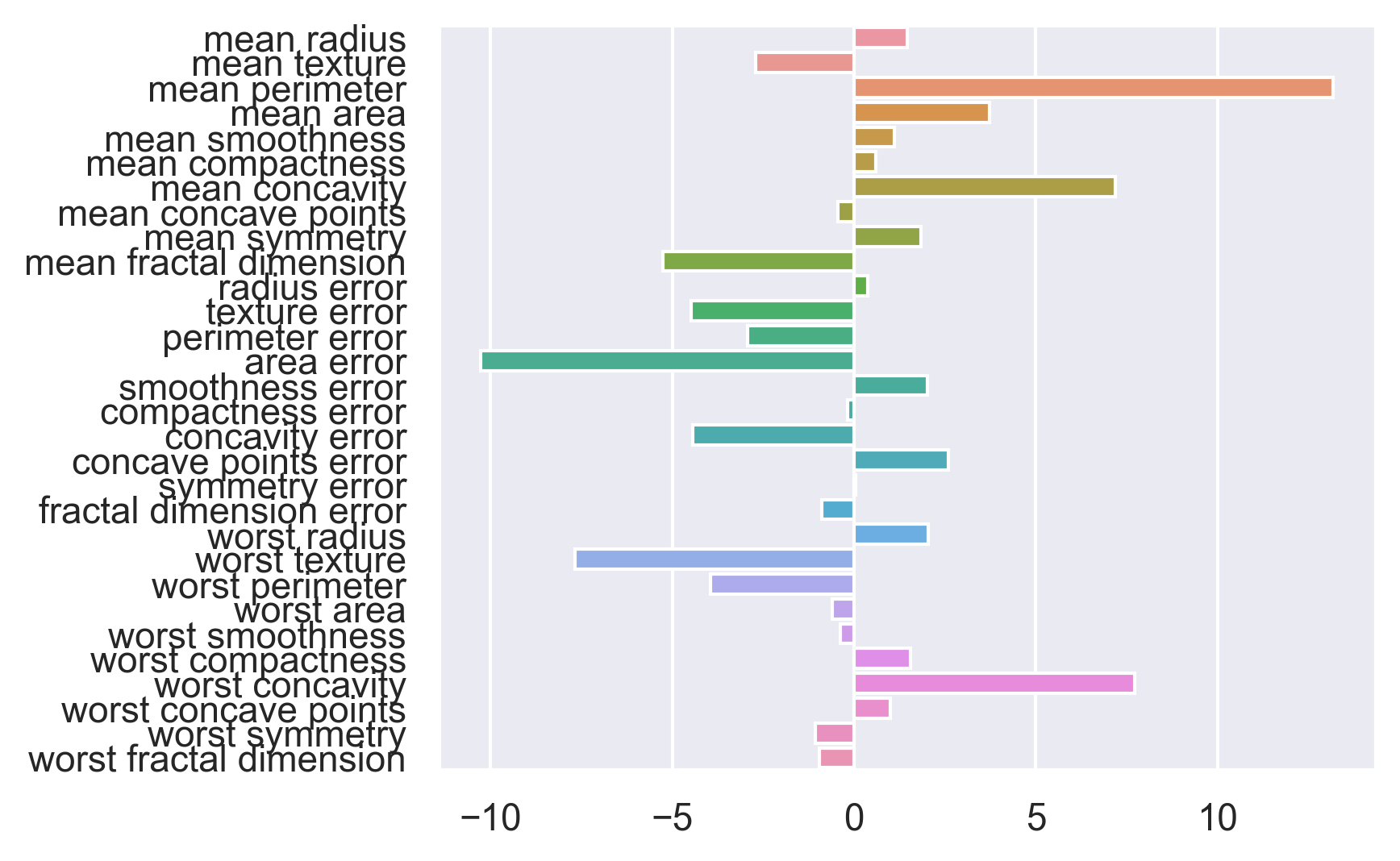}
\caption{Detailed contribution of each feature from CF.}
\label{fig:linear_contribution_calculation}
\end{figure*}

\begin{figure*}[]
\centering
(a)\includegraphics[width=.4\textwidth]{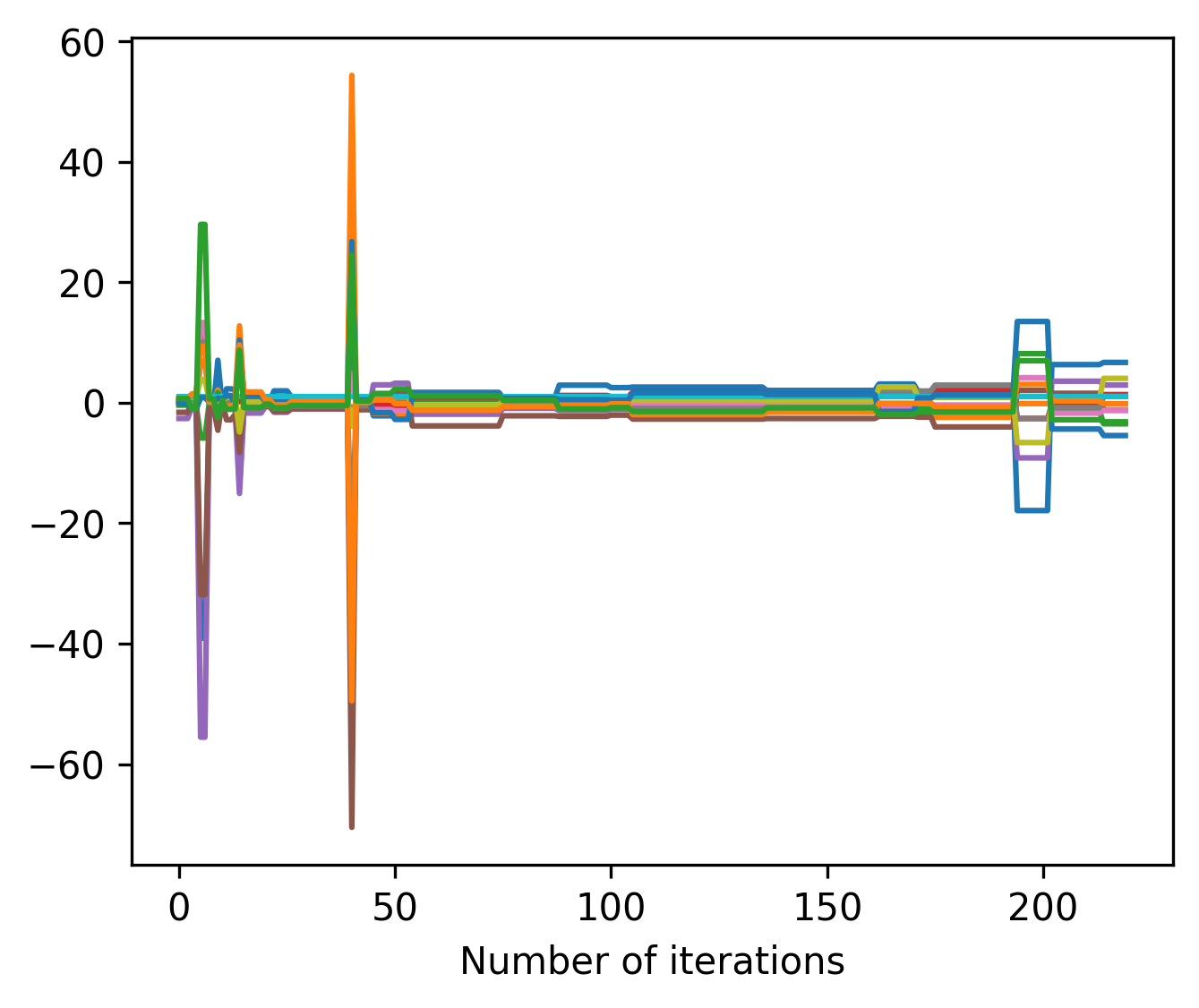}
(b)\includegraphics[width=.4\textwidth]{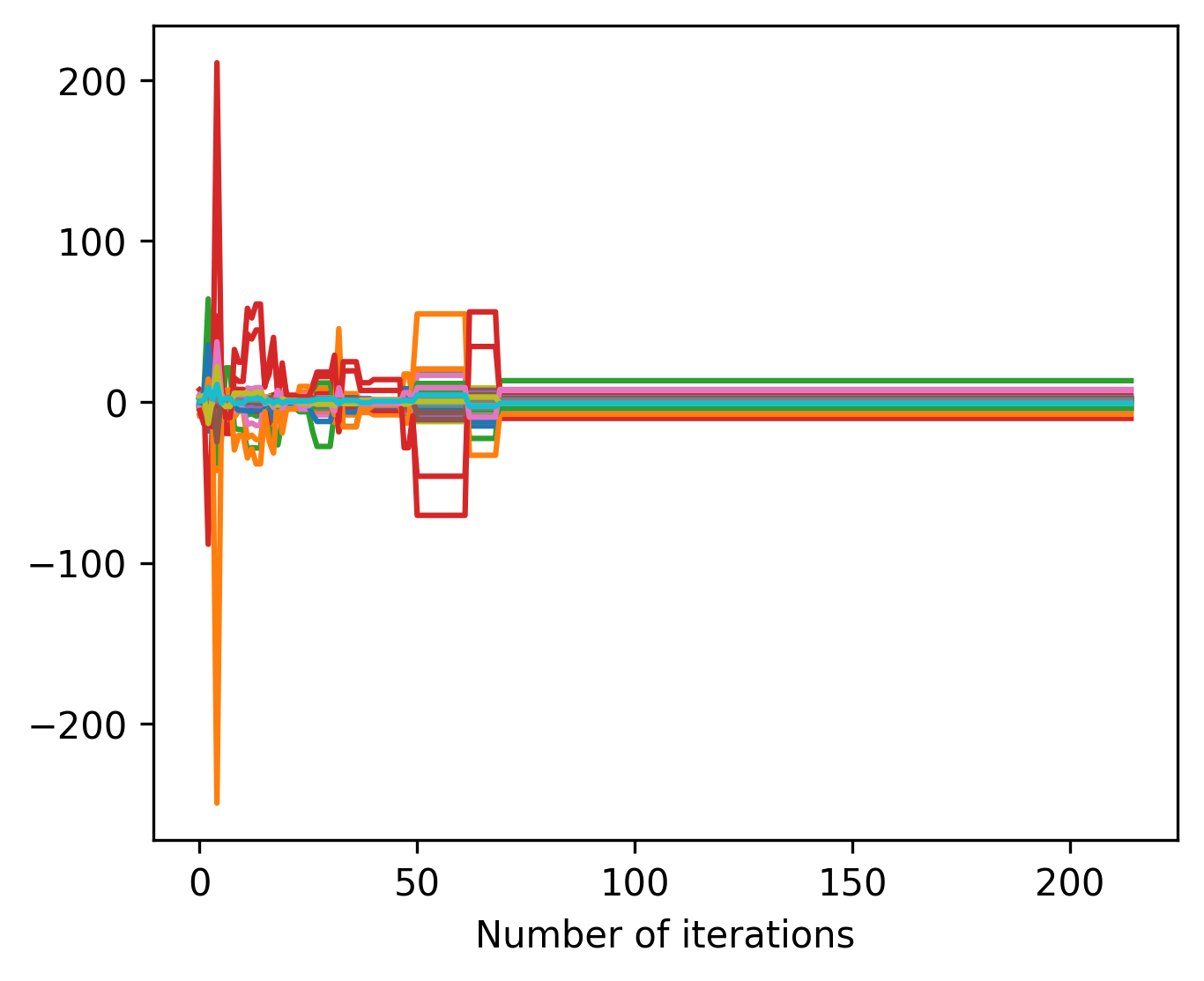}
\caption{Contribution distribution of linear regression and logistic regression with equations accumulated.}
\label{fig:linear_contribution_calculation_iteration}
\end{figure*}

\begin{figure}[]
\centering
\includegraphics[width=0.4\textwidth]{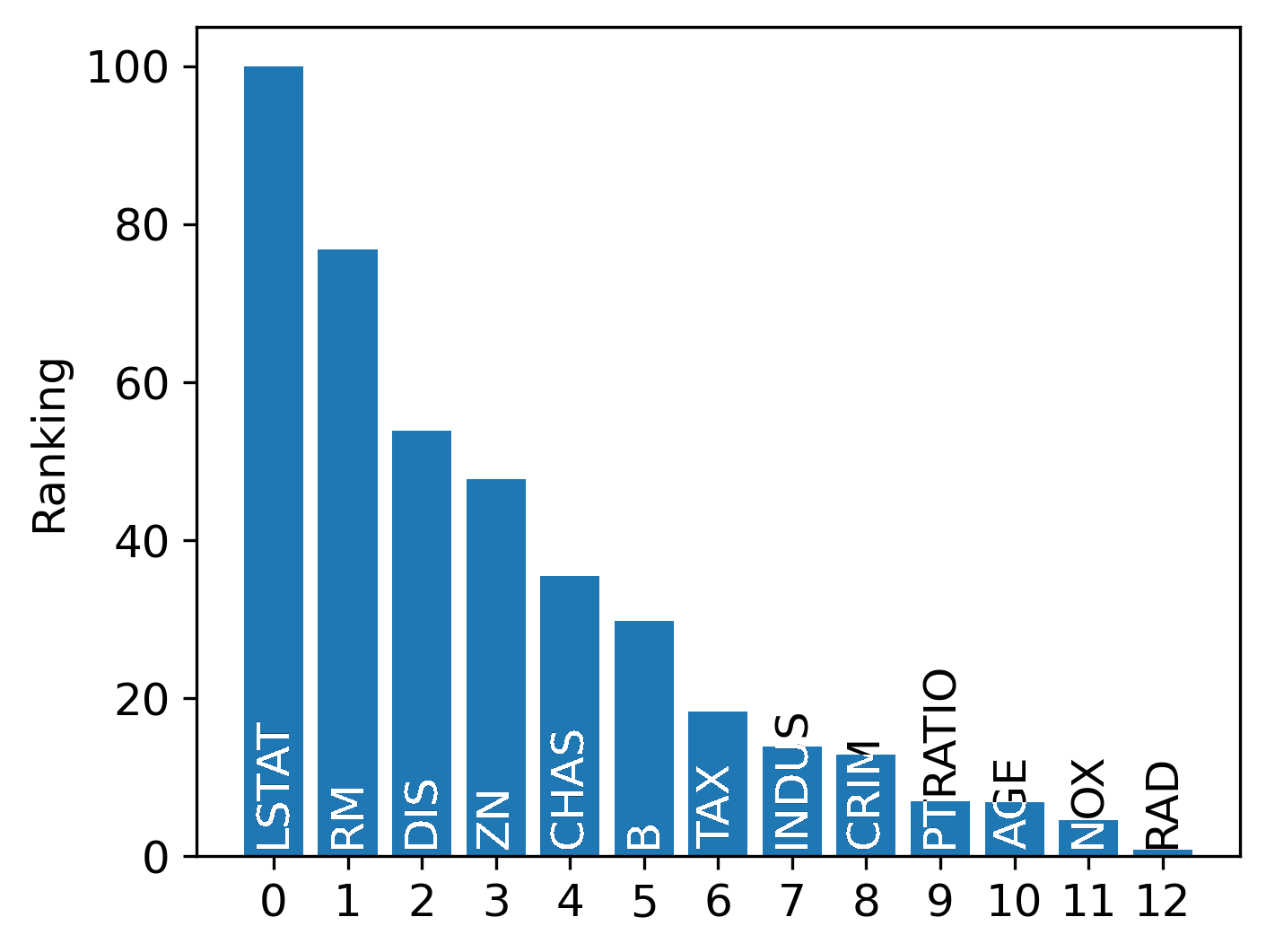}
\includegraphics[width=0.4\textwidth]{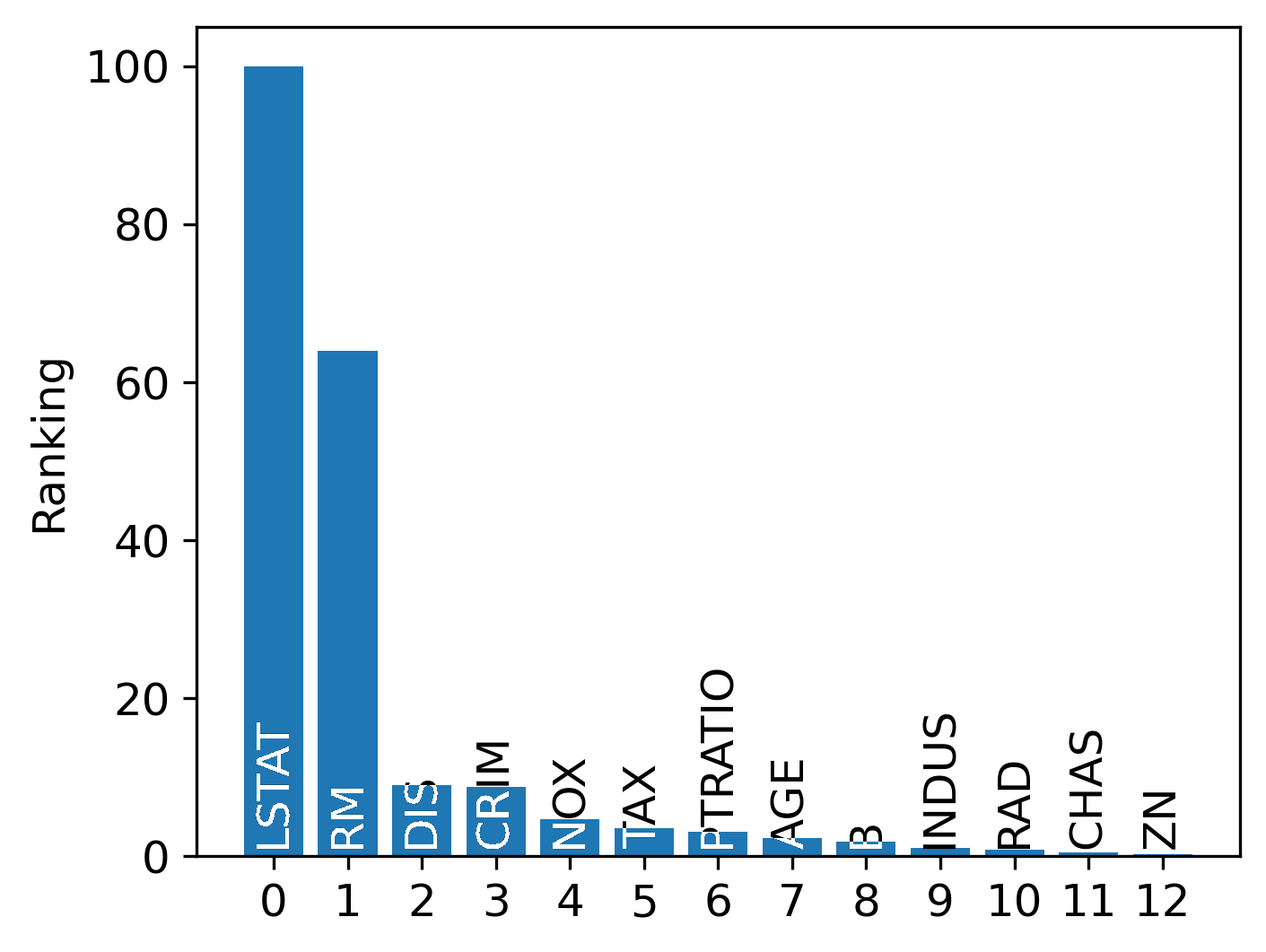}
(a)\includegraphics[width=0.4\textwidth]{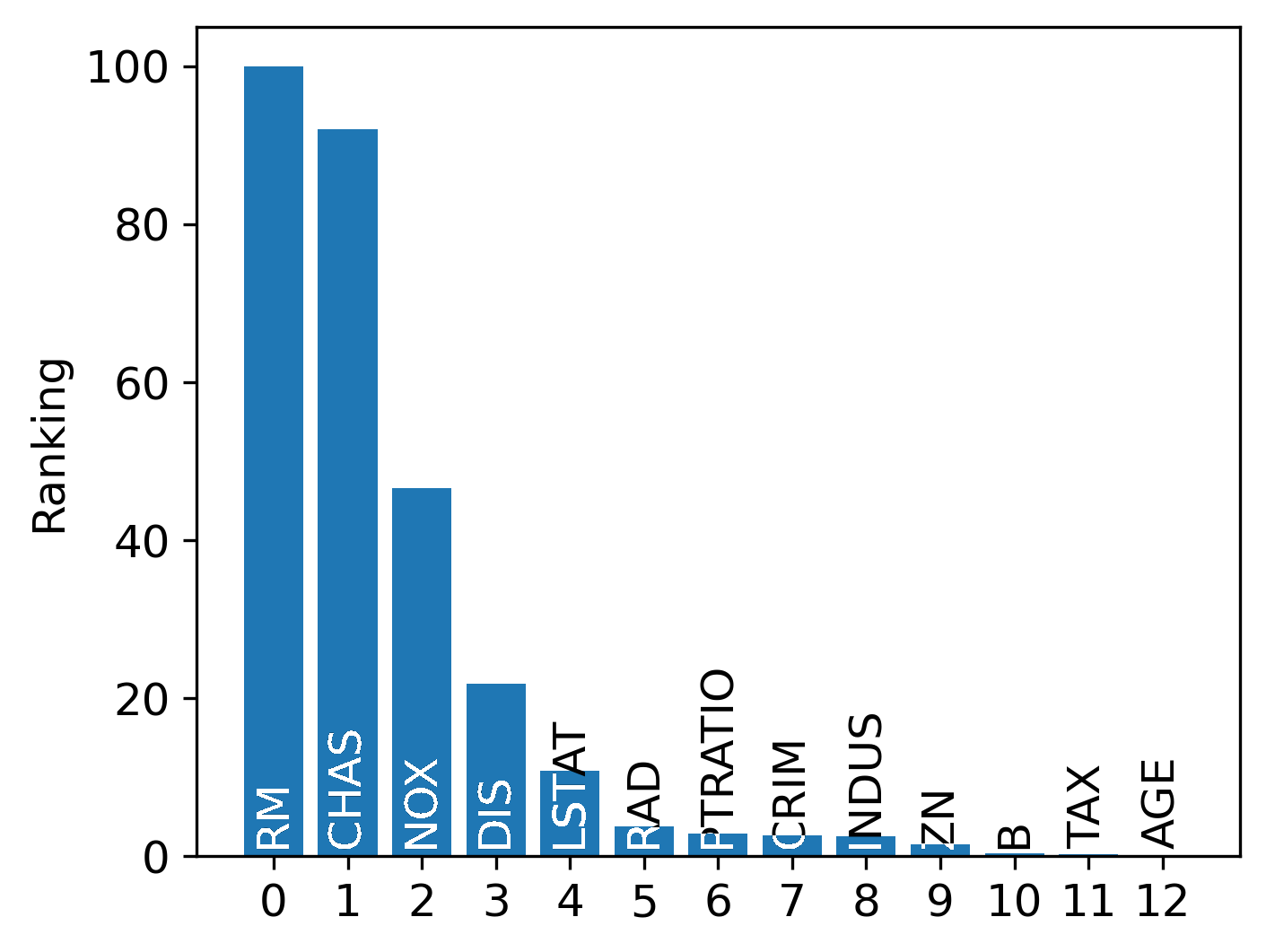}
\includegraphics[width=0.4\textwidth]{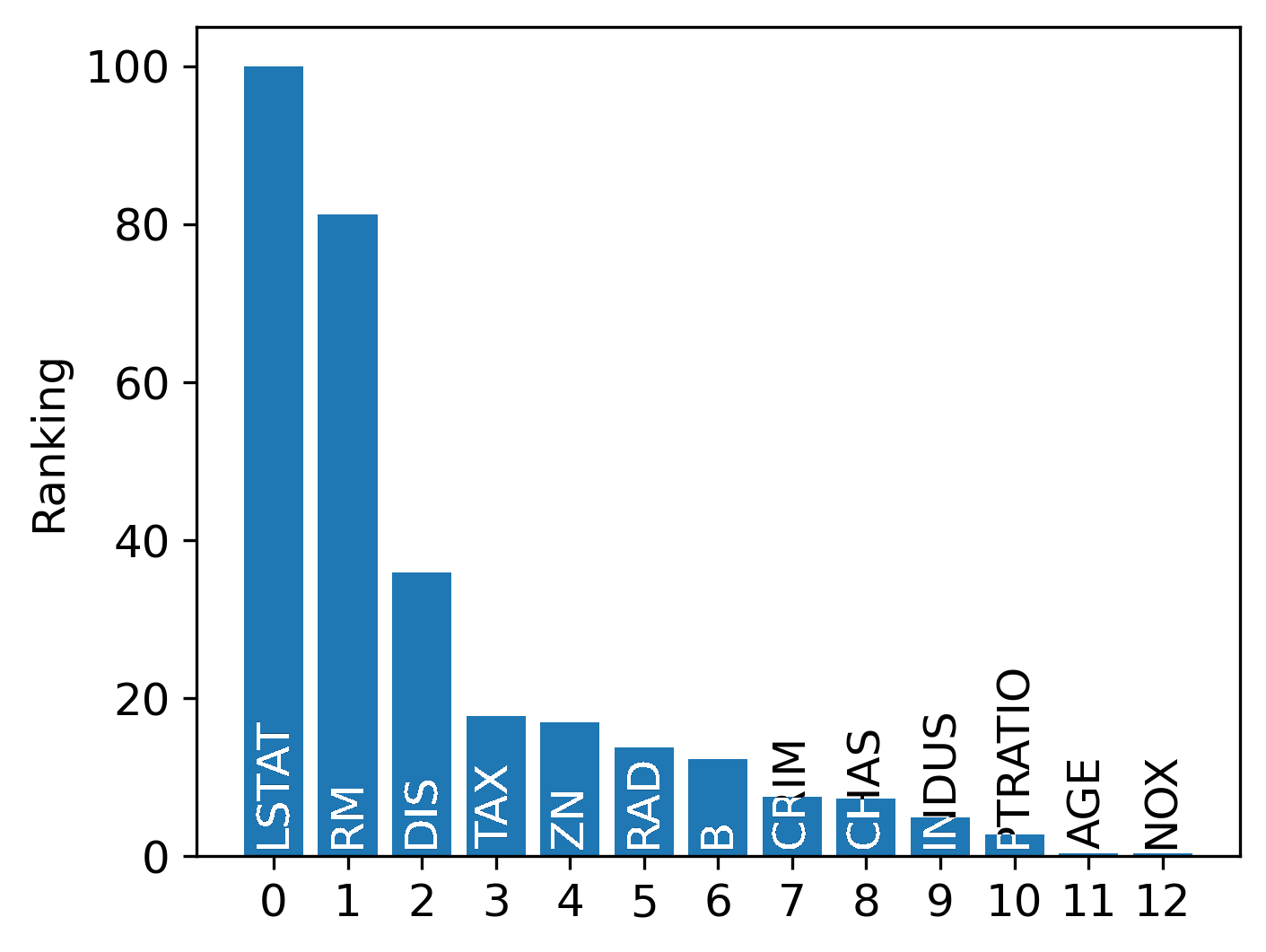}
\includegraphics[width=0.4\textwidth]{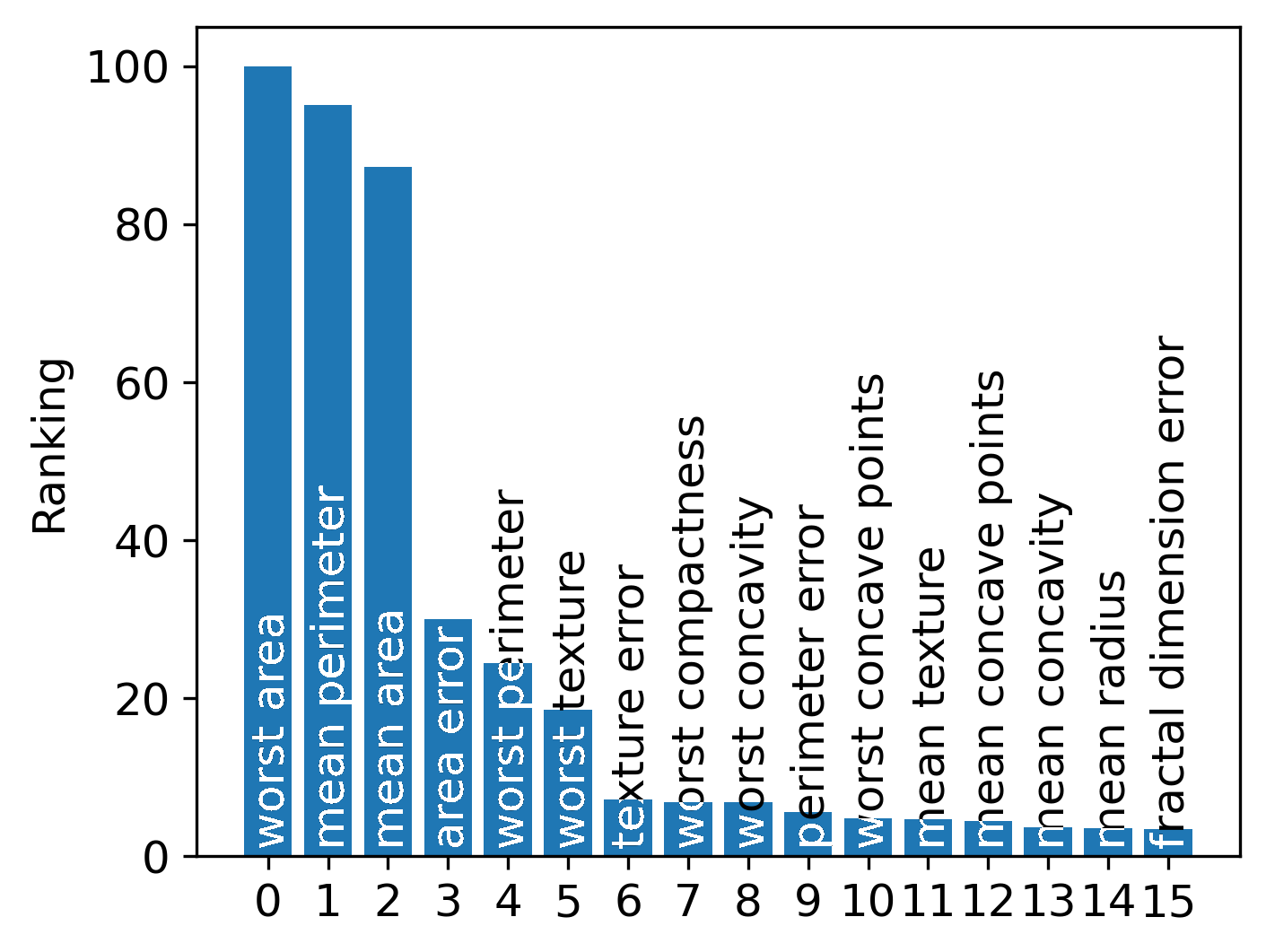}
\includegraphics[width=0.4\textwidth]{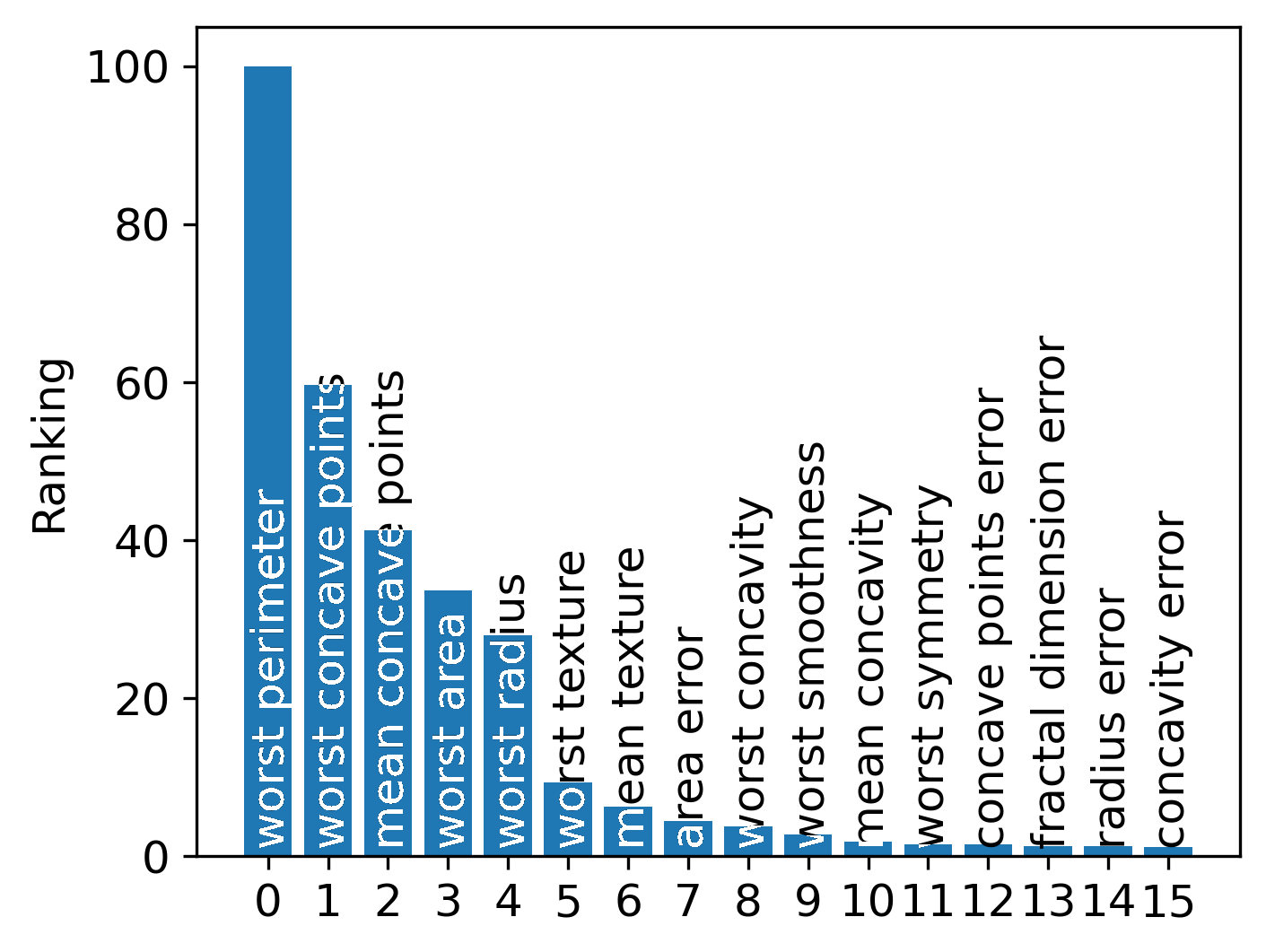}
(b)\includegraphics[width=0.4\textwidth]{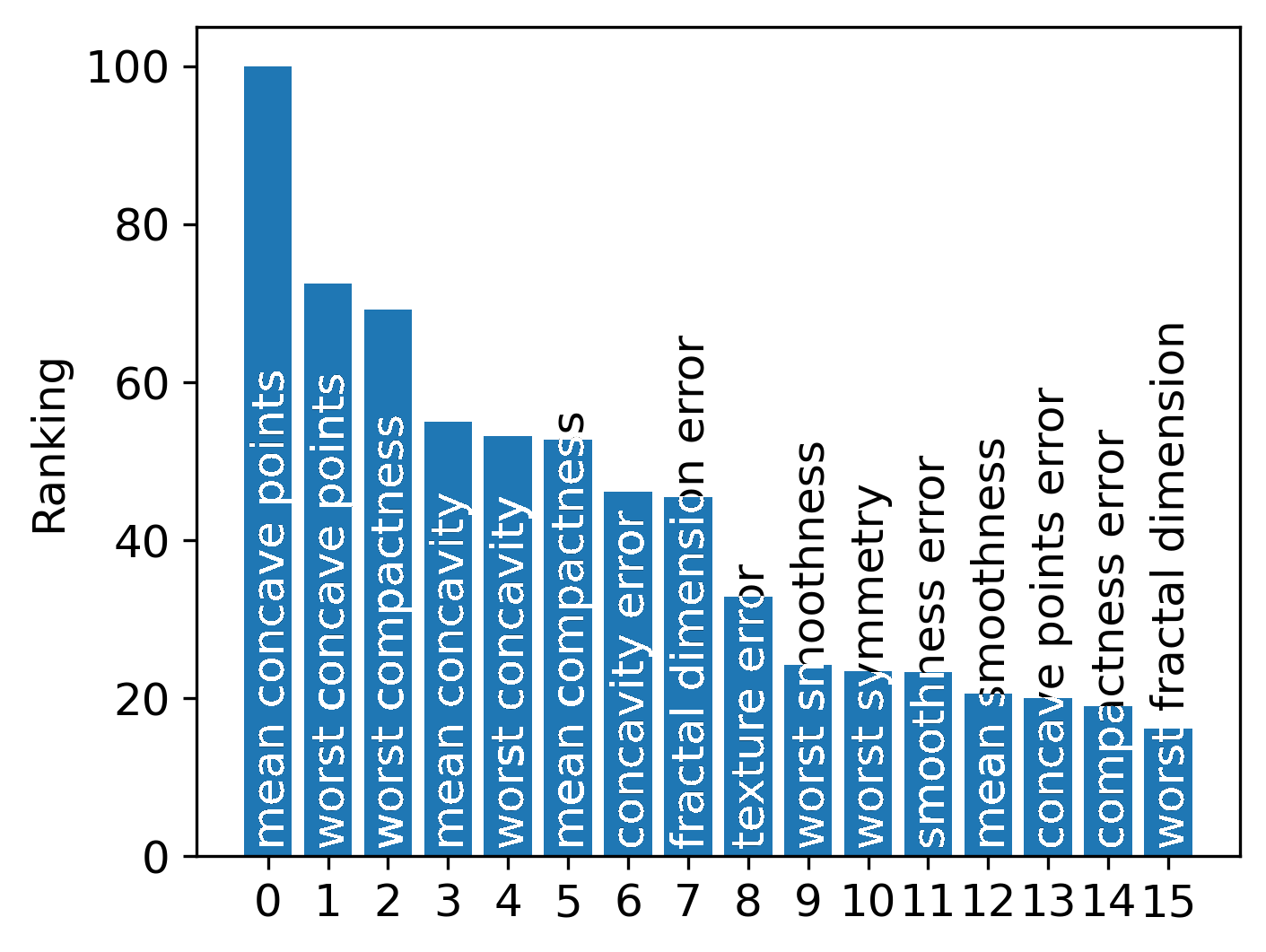}
\includegraphics[width=0.4\textwidth]{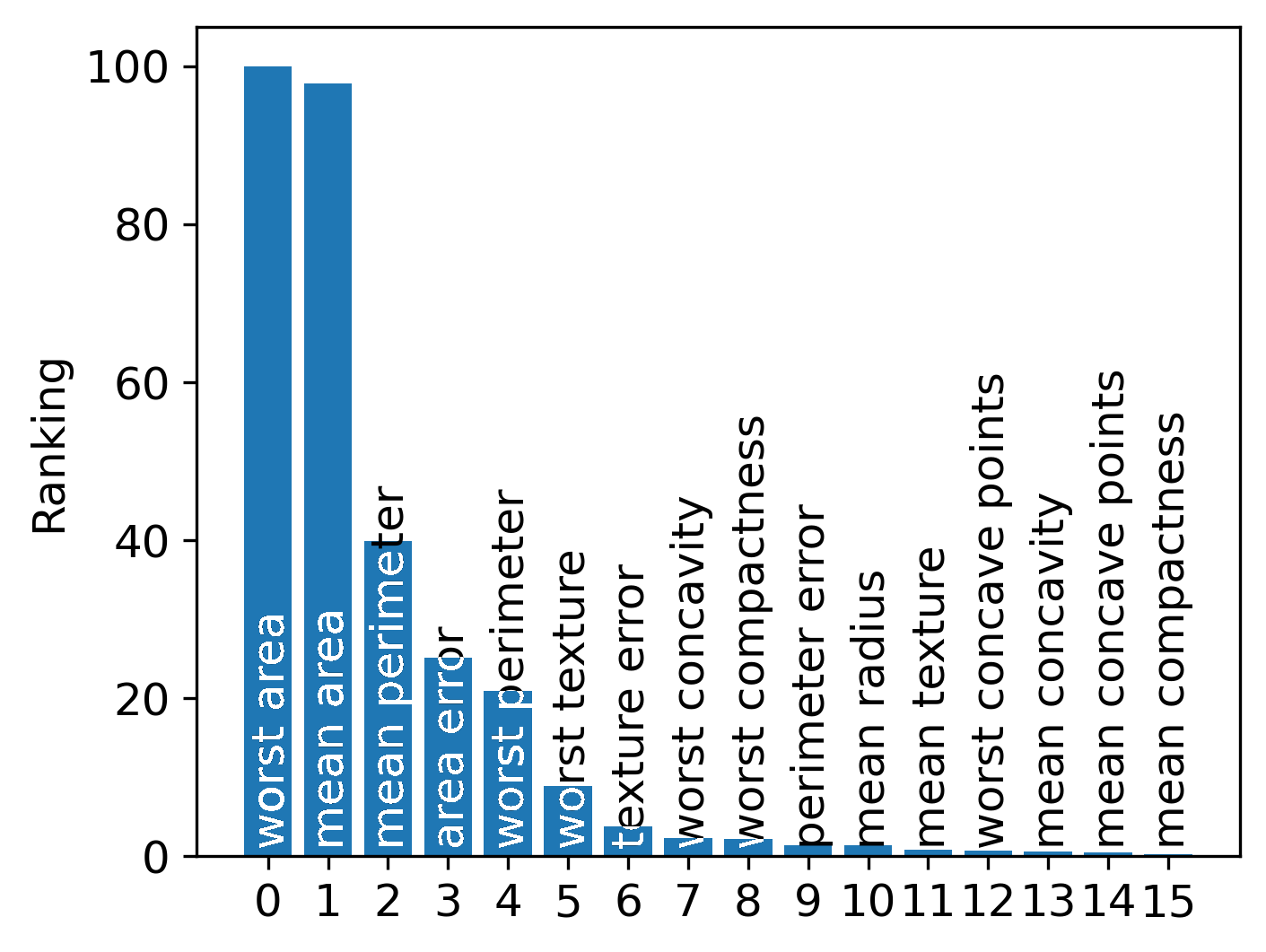}
\caption{The feature importance profile of Boston house price and Breast Cancer Wisconsin from KernalSHAP, random forest (RF), Connection weights, Permutation method (from left to right) (a) Linear regression, (b) Logistic regression.}
\label{fig:linear_FI_results}
\end{figure}

Supporting information for linear regression and logistic regression, including linear regressor and logistic regressor training settings in Table \ref{tb:linear_model}; linear and logistic base model and feature model structures in Table \ref{tb:linear_model_structure}; the training settings of independent model in Table \ref{tb:inde_model}; weights distribution of linear and logistic models in Fig. \ref{fig:linear_base_model_weight_plot}; profile change of linear regression and logistic regression as the number of iterations increases in Fig. \ref{fig:linear_base_model_converge}; the feature importance profile of Boston house price and Breast Cancer Wisconsin from different feature selection methods in Fig. \ref{fig:linear_FI_results}; important features and detailed contribution of each feature in Fig. \ref{fig:linear_contribution_calculation}, and CF of linear regression and logistic regression with equations accumulated in Fig. \ref{fig:linear_contribution_calculation_iteration}.

\begin{figure}[h!]
\centering
\includegraphics[width=0.4\textwidth]{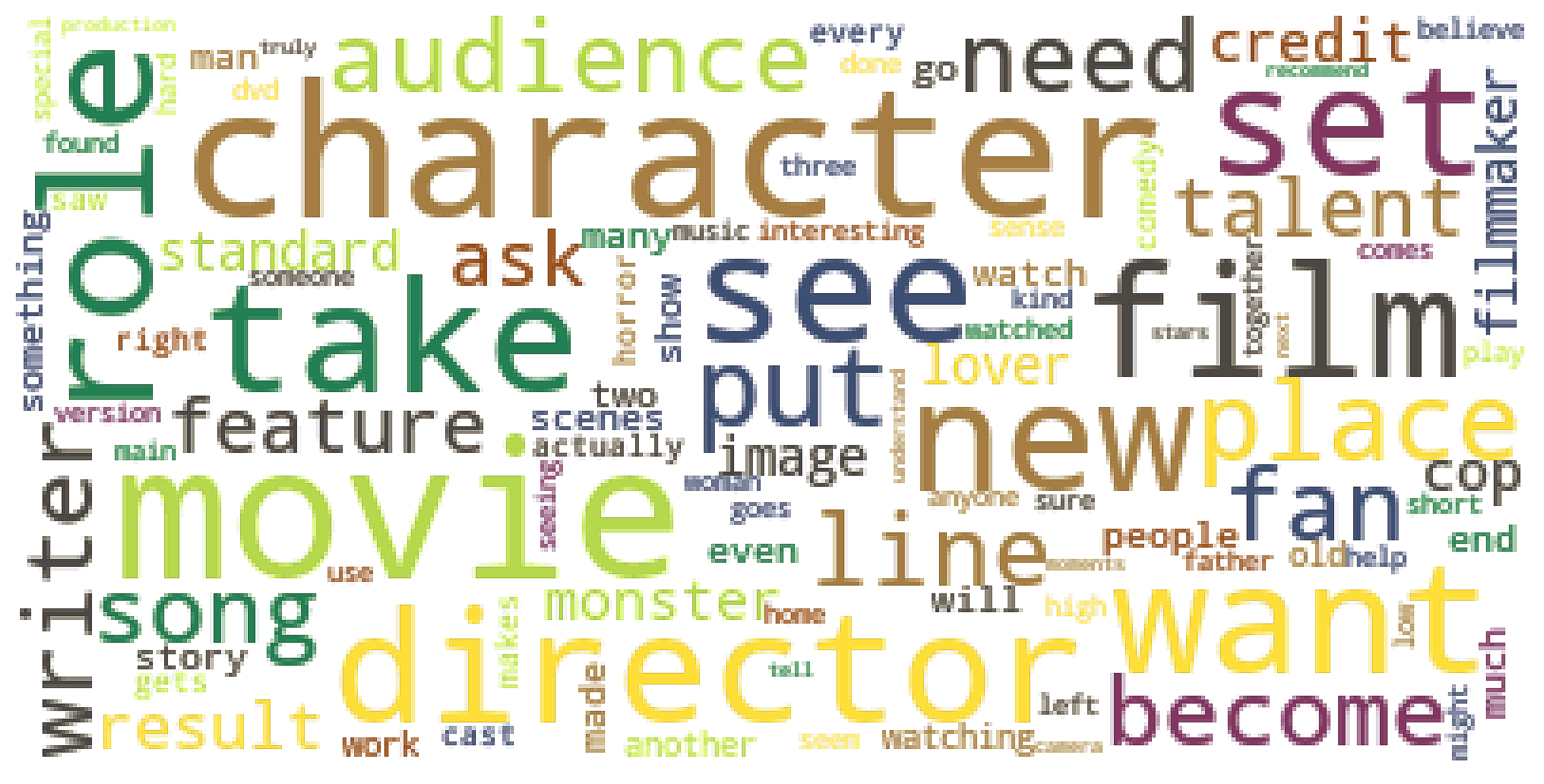}
\caption{Unimportant words.}
\label{fig:RNN_unimportant}
\end{figure}

\begin{table*}[]
\centering
\begin{tabular}{llllp{1.2cm}lp{0.5cm}}
 & Bulk & Surface  & Total & Condensed  & Expanded \\ \hline
MAE & 0.1109 & 0.1355 & 0.0981 & 0.0615 & 0.0529 \\
MSE & 0.0253 & 0.0421 & 0.0213 & 0.0087 & 0.0117 \\ \hline 
\end{tabular}
\caption{Results of 5 linear models trained on Bulk (B), Surface (S), Totals (T), Condensed (C) and Expanded (E), Gold feature sets.}
\label{tb:gold_results}
\end{table*}

%\begin{figure*}[ph!]
%\centering
%\includegraphics[width=.4\textwidth]{linear/linear_correlation_matrix}
%\includegraphics[width=.4\textwidth]{logistic/logistic_correlation_matrix}
%\caption{Correlation matrix for the Boston housing dataset and breast cancer Wisconsin dataset}
%\label{fig:linear_base_model_correlation_matrix}
%\end{figure*}

%\begin{table}[]
%\centering
%\begin{tabular}{ccccc}
%& Layer & Layer type  & Layer parameters  & Output size \\ \hline
%& L0 & Dense layer  & input\_dim, relu   & N, 32  \\
%& L1 & Dropout layer  & 0.1 & N, 30 \\
%& L2 & Dense layer  & relu & N, 32  \\
%& L3 & Dropout layer  & 0.1 & N, 30 \\
%& L4 & Dense layer  & sigmoid & N, 1 \\
%\end{tabular}
%\caption{The structure of independent model used to compare}
%\label{tb:independent_model_structure}
%\end{table}

\subsection*{Application Supporting Information}

\begin{figure*}[h!]
\centering
(a)\includegraphics[width=0.3\textwidth]{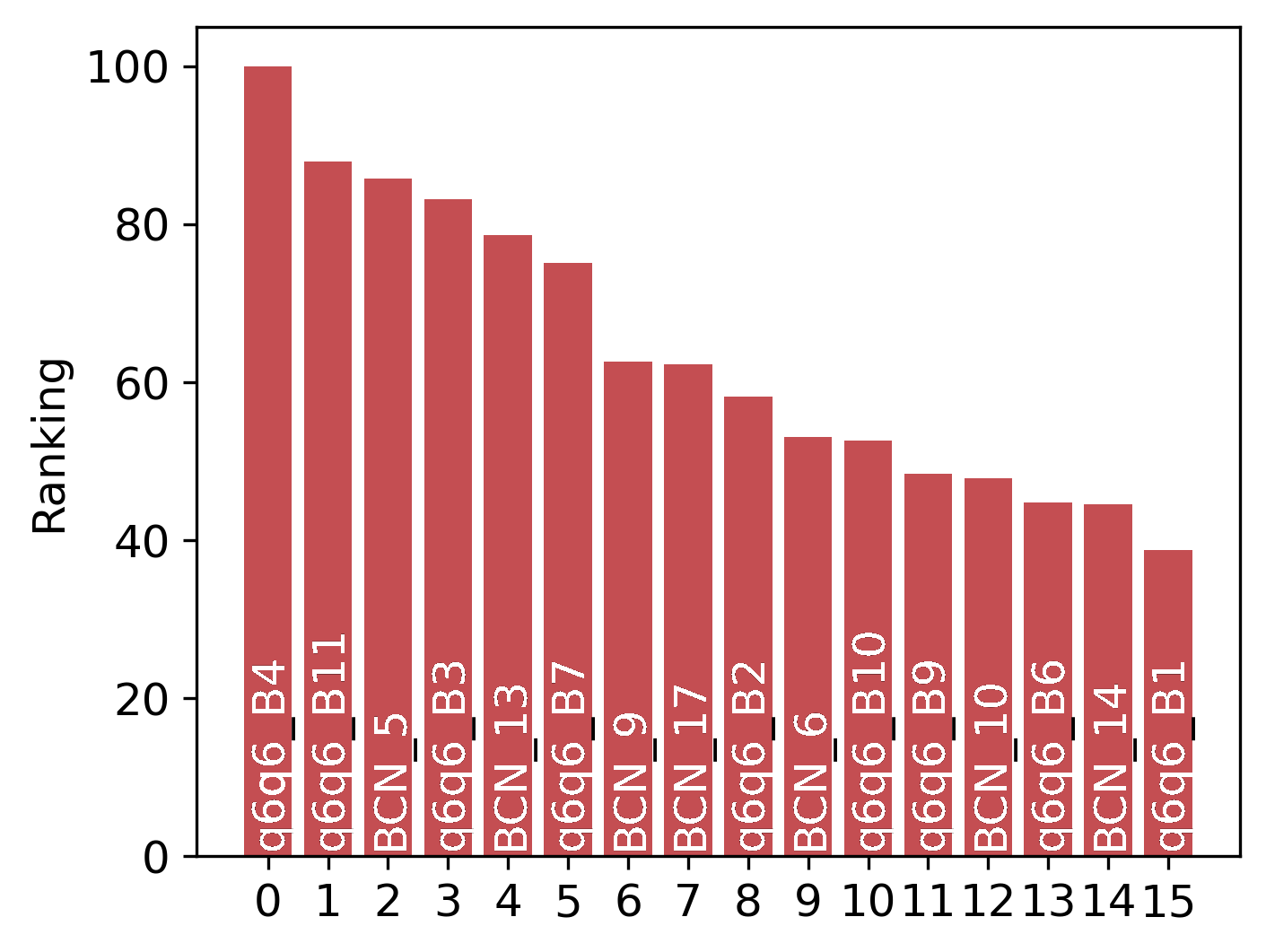}
\includegraphics[width=0.3\textwidth]{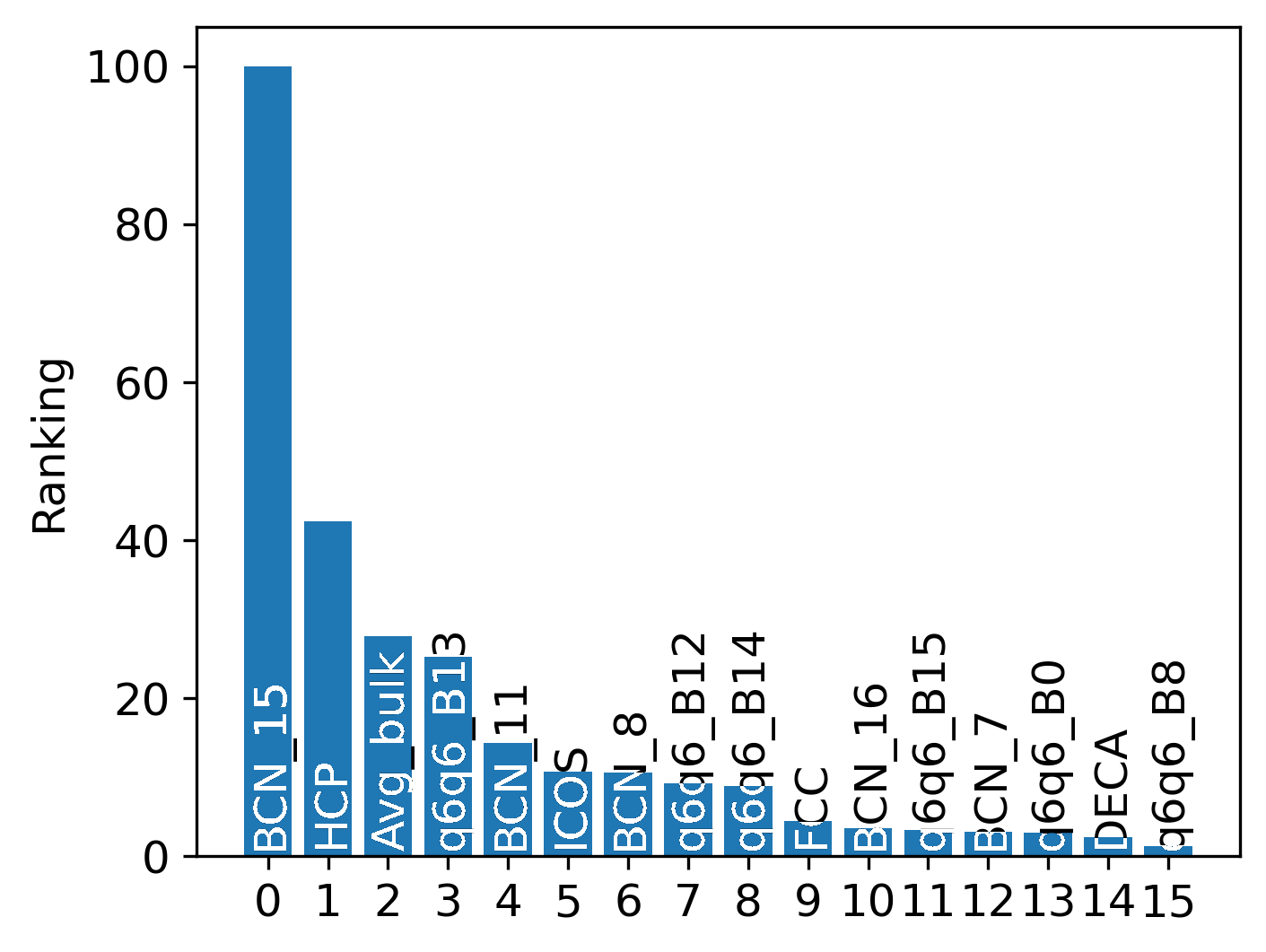}
\includegraphics[width=0.3\textwidth]{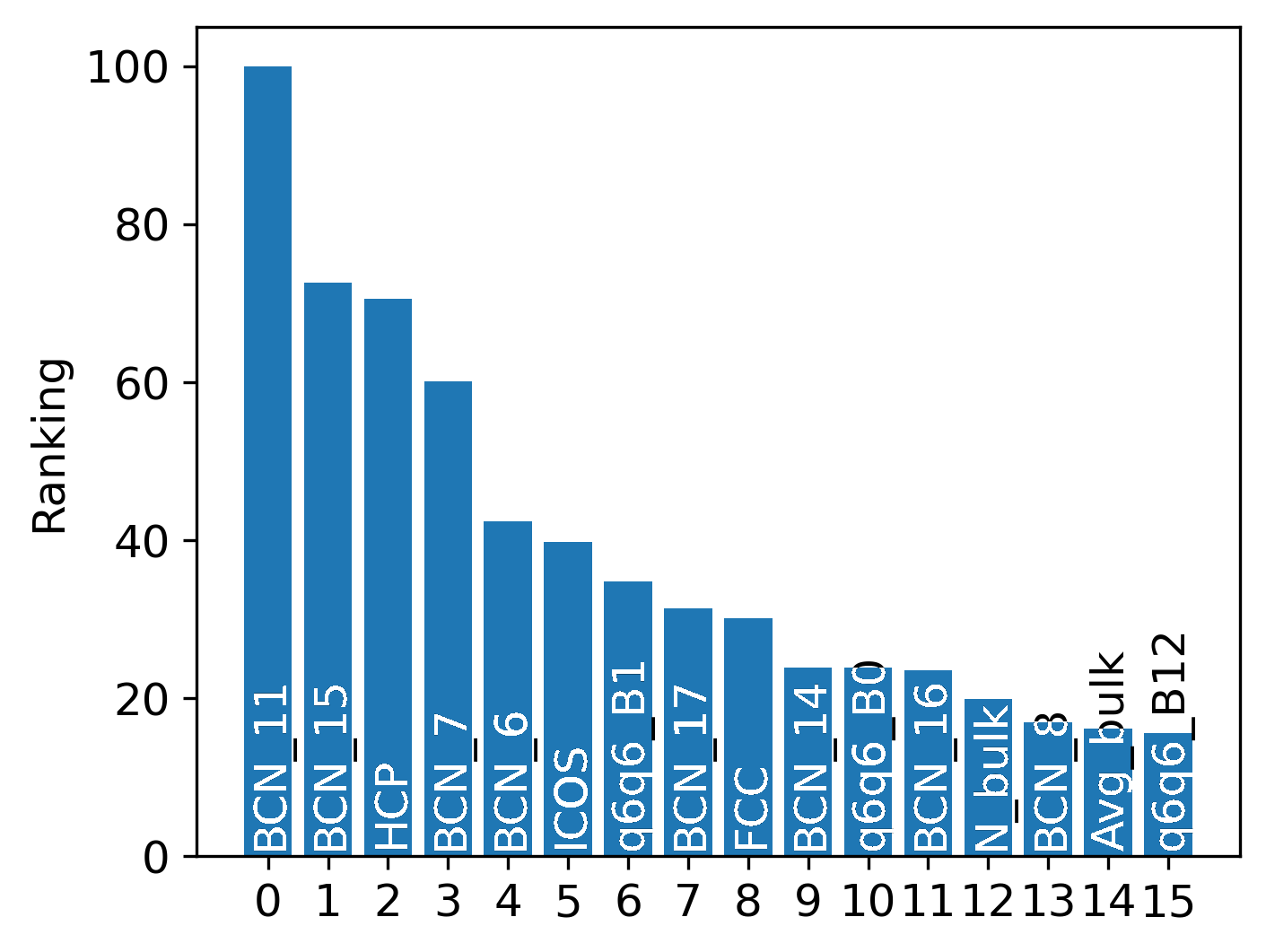}\\
(b)\includegraphics[width=0.3\textwidth]{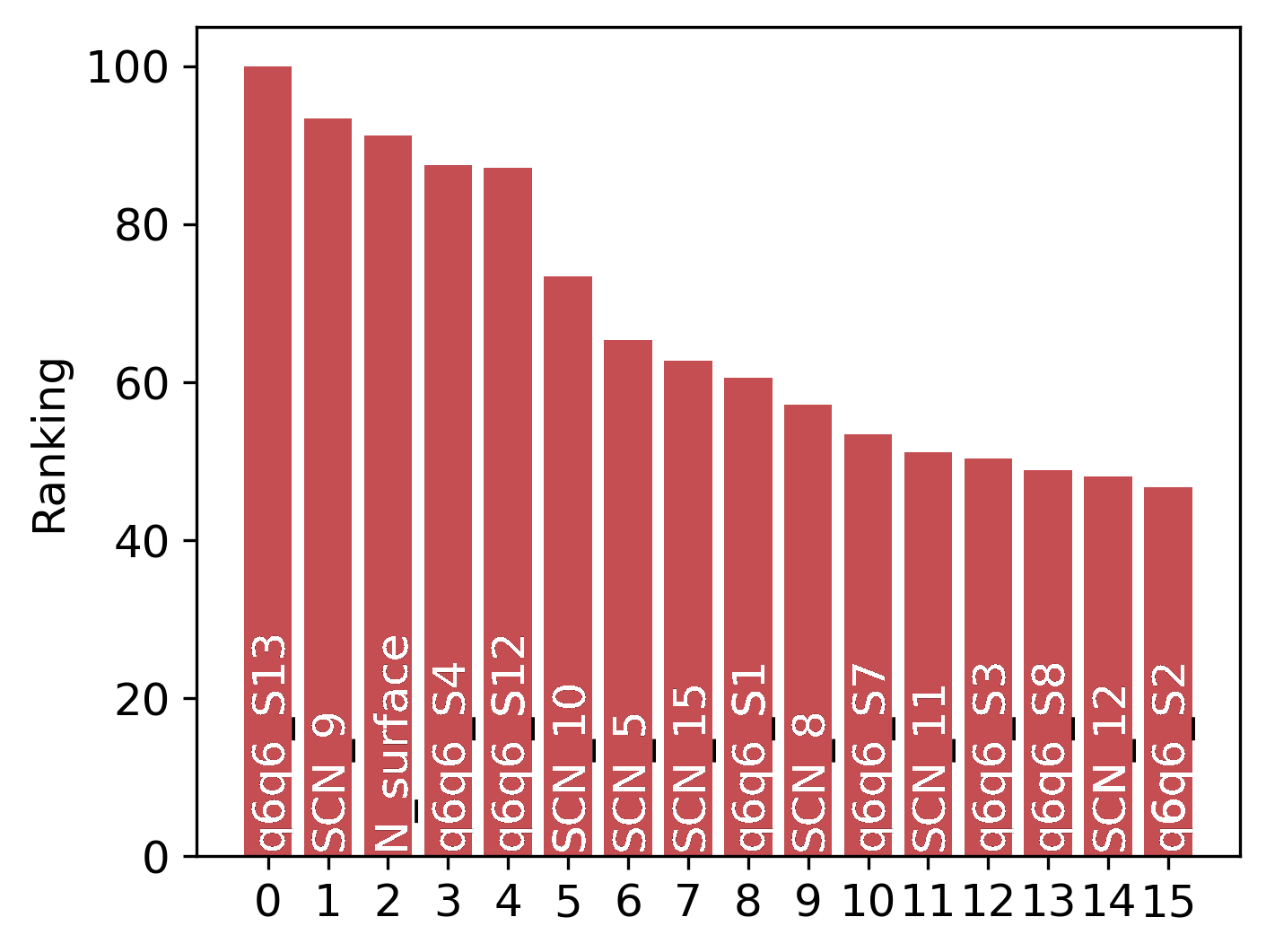}
\includegraphics[width=0.3\textwidth]{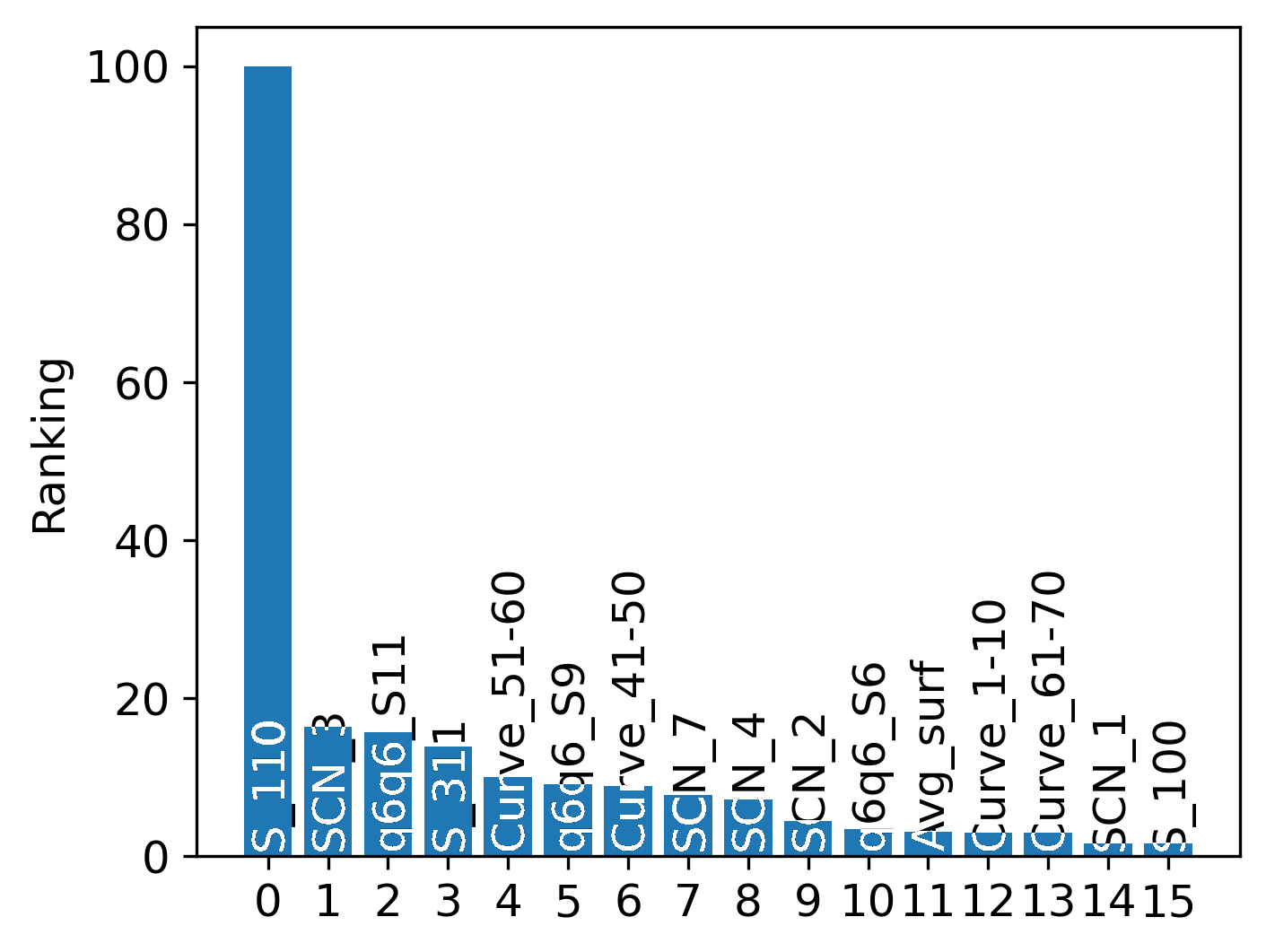}
\includegraphics[width=0.3\textwidth]{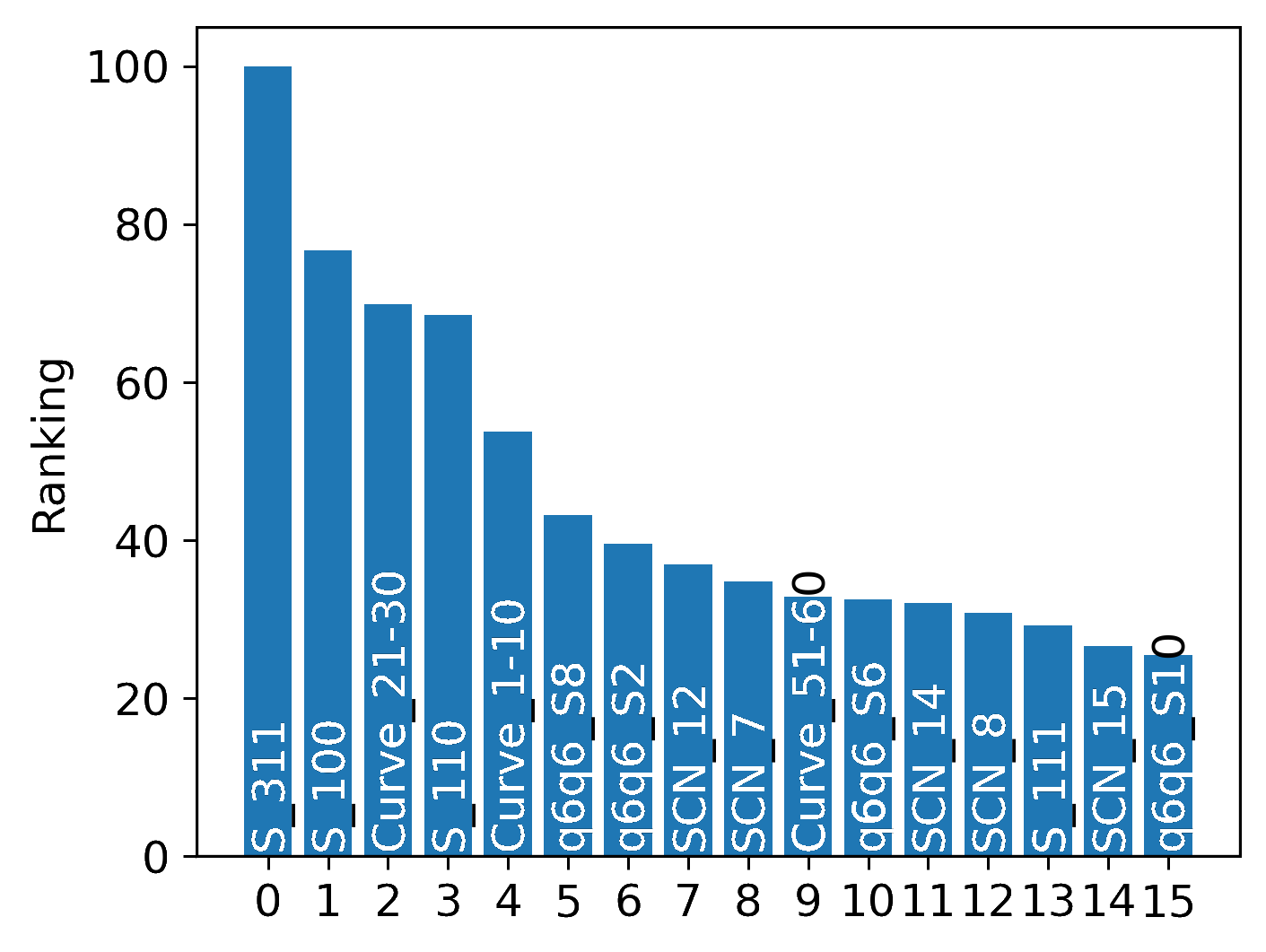}\\
(c)\includegraphics[width=0.3\textwidth]{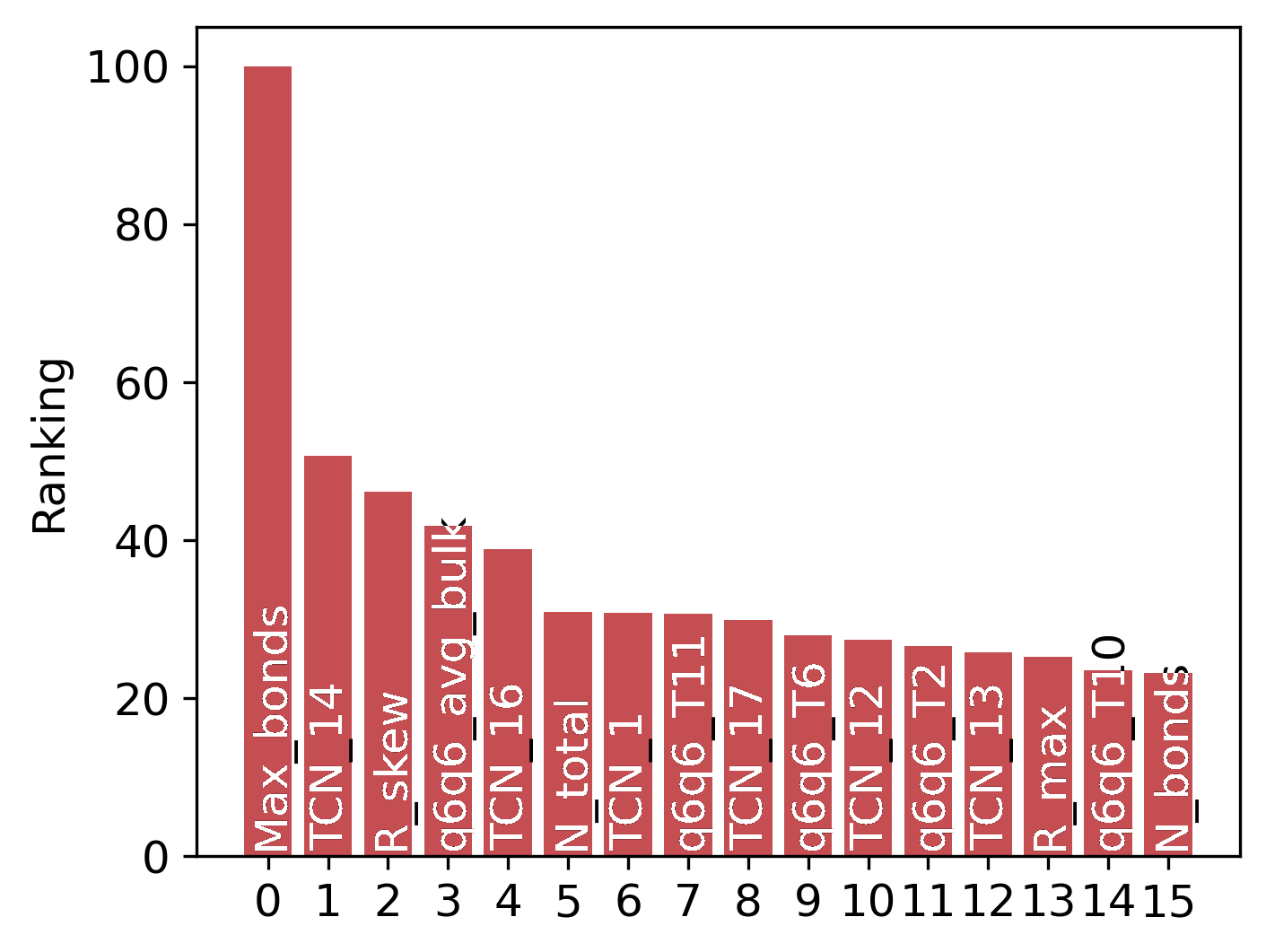}
\includegraphics[width=0.3\textwidth]{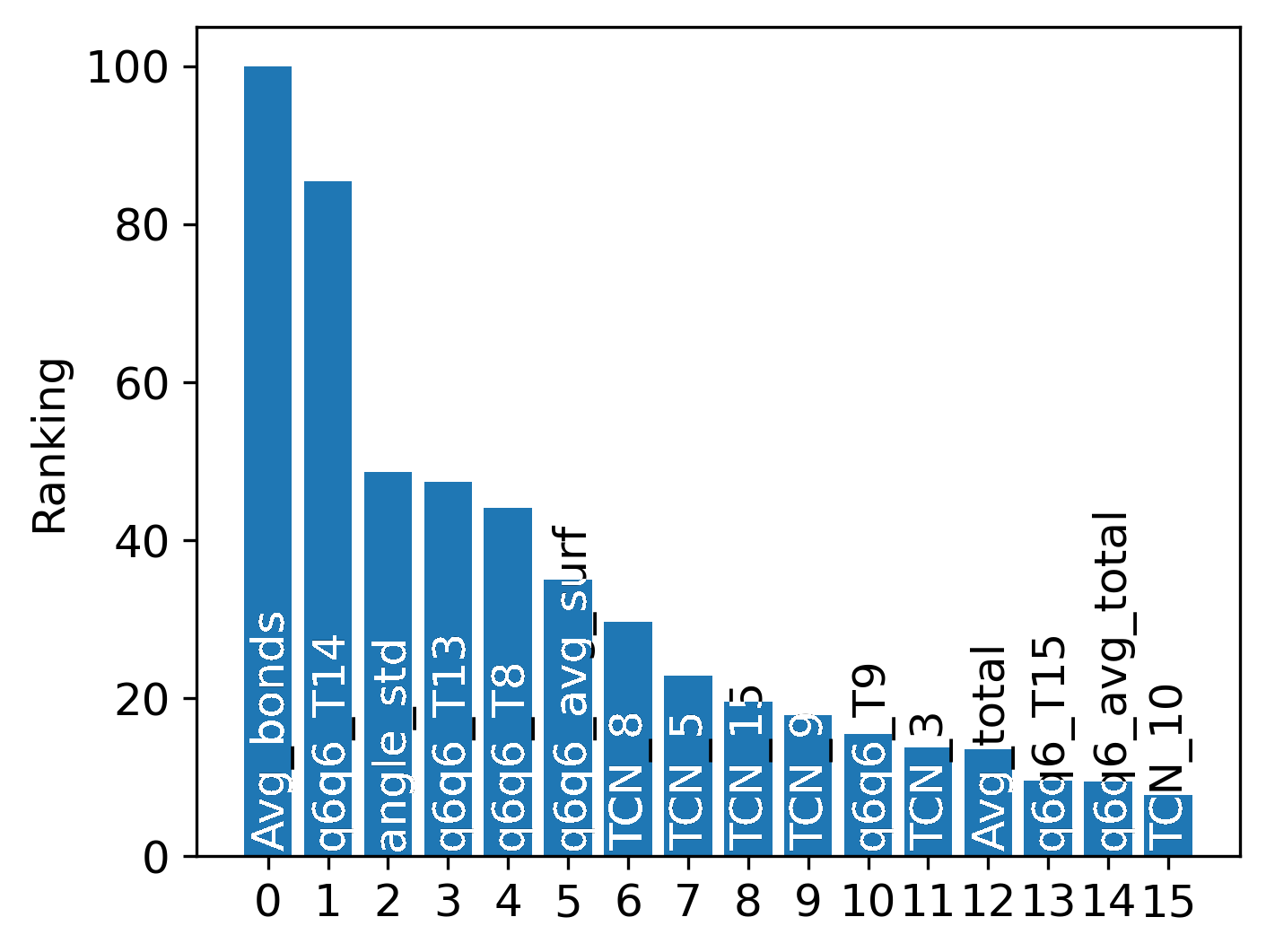}
\includegraphics[width=0.3\textwidth]{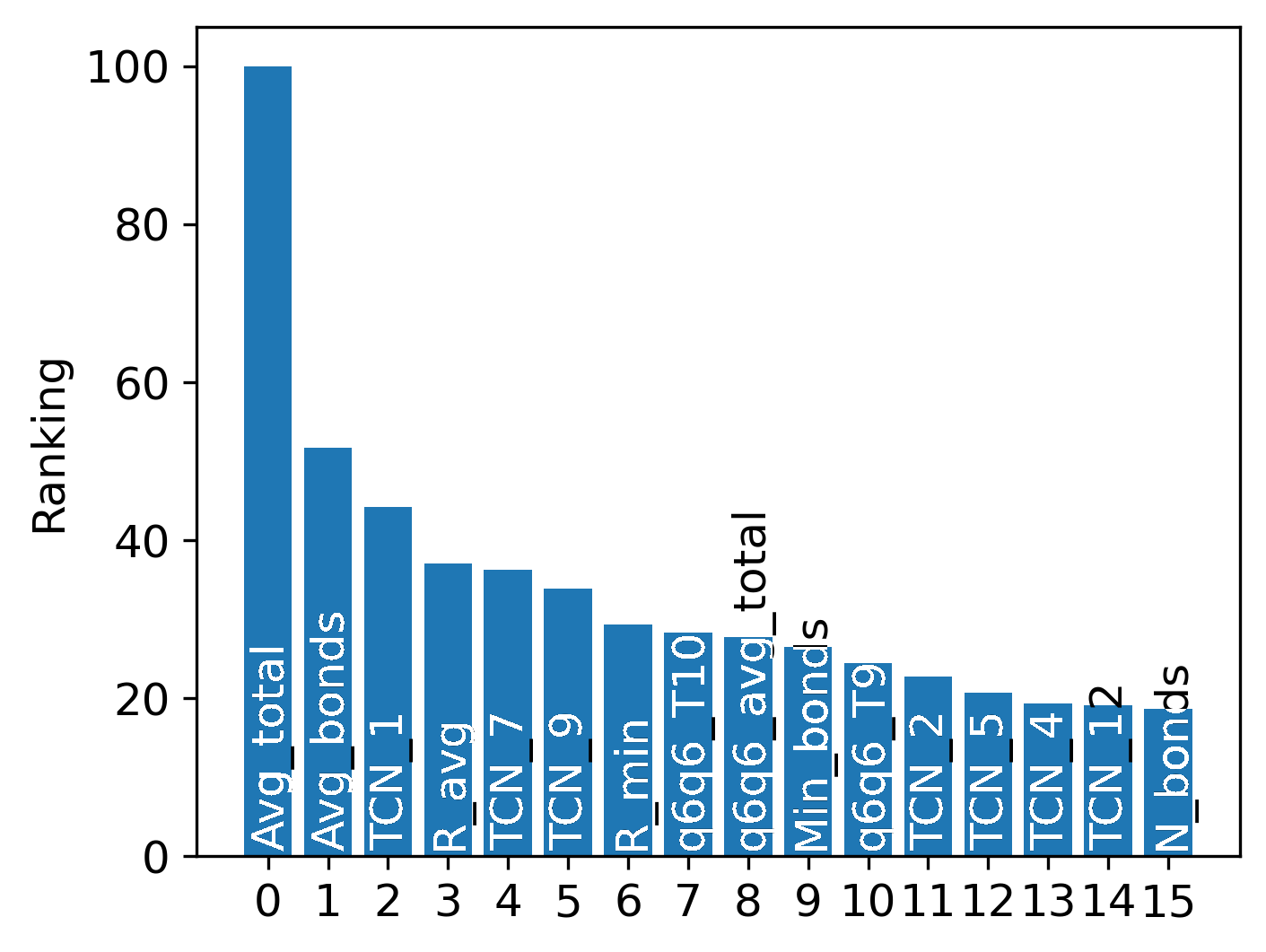}\\
(d)\includegraphics[width=0.3\textwidth]{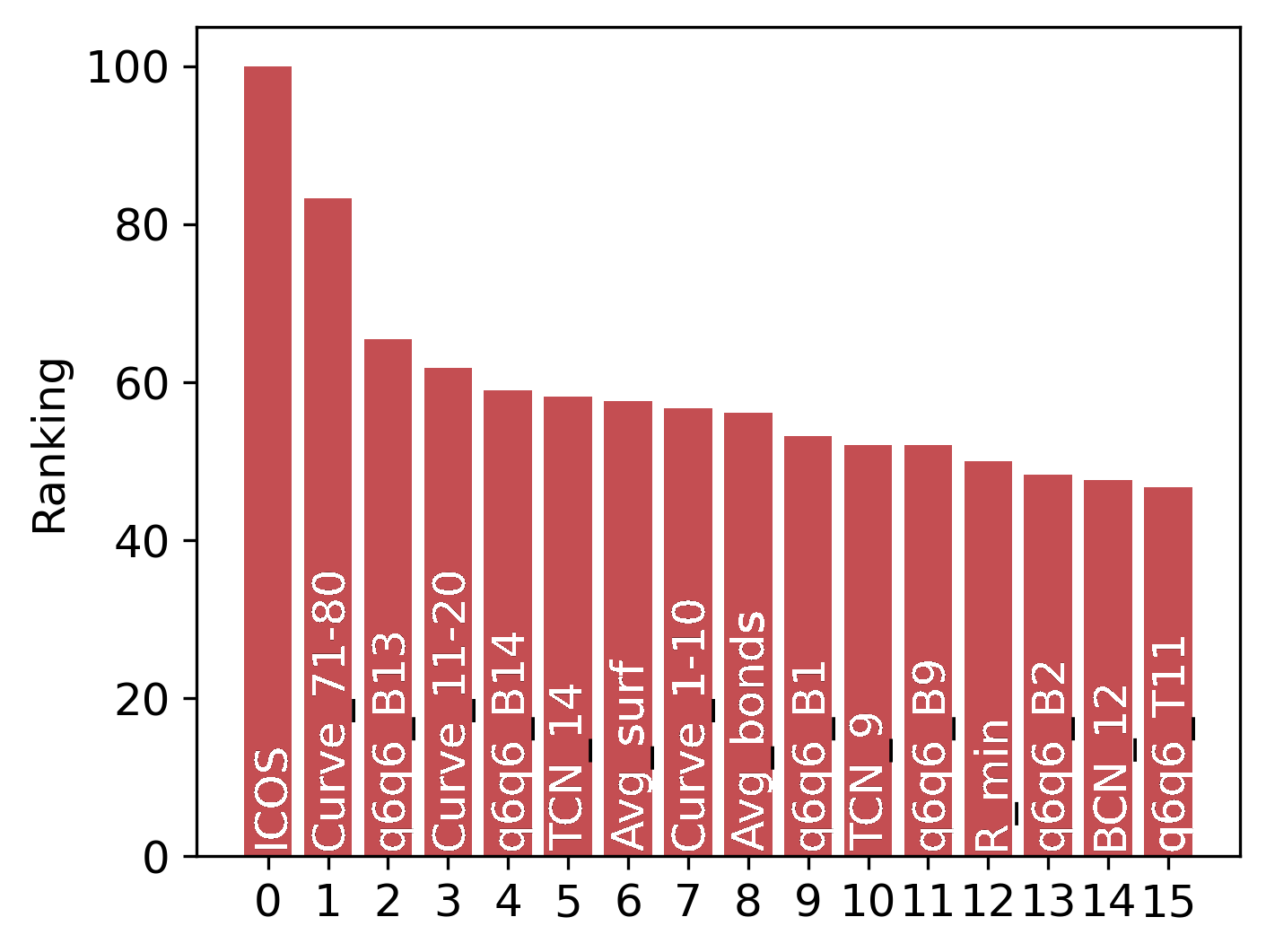}
\includegraphics[width=0.3\textwidth]{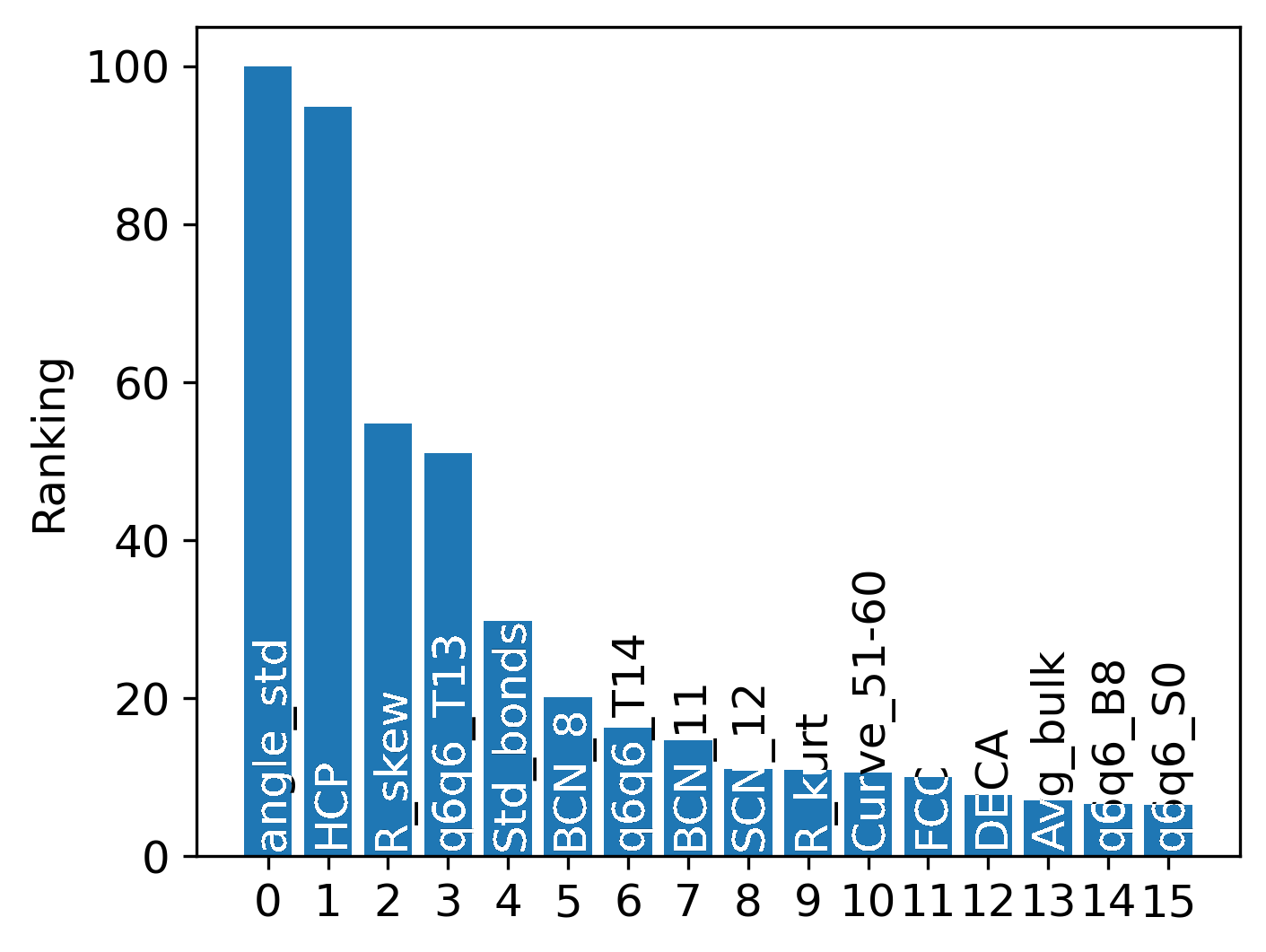}
\includegraphics[width=0.3\textwidth]{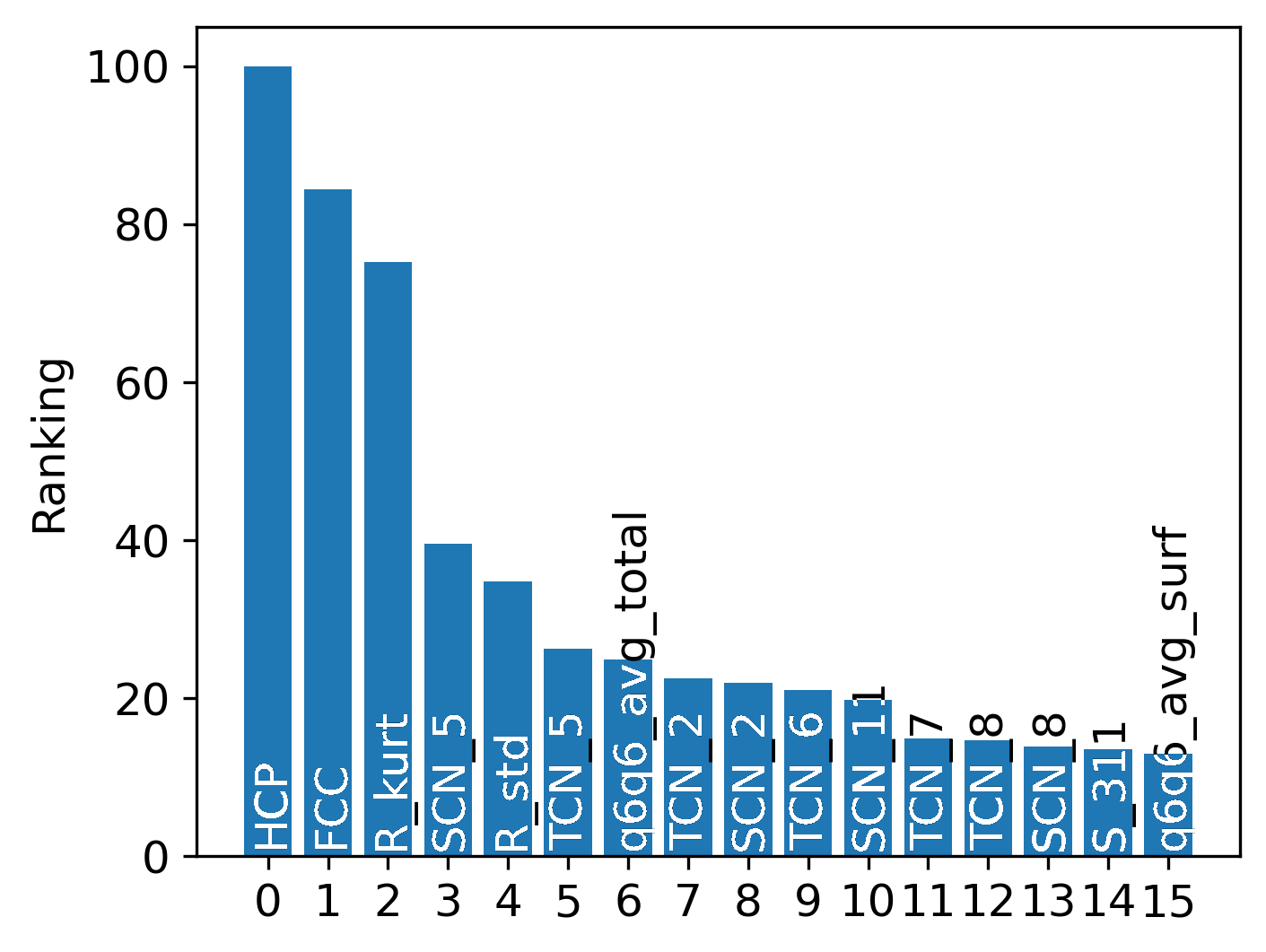}\\
(e)\includegraphics[width=0.3\textwidth]{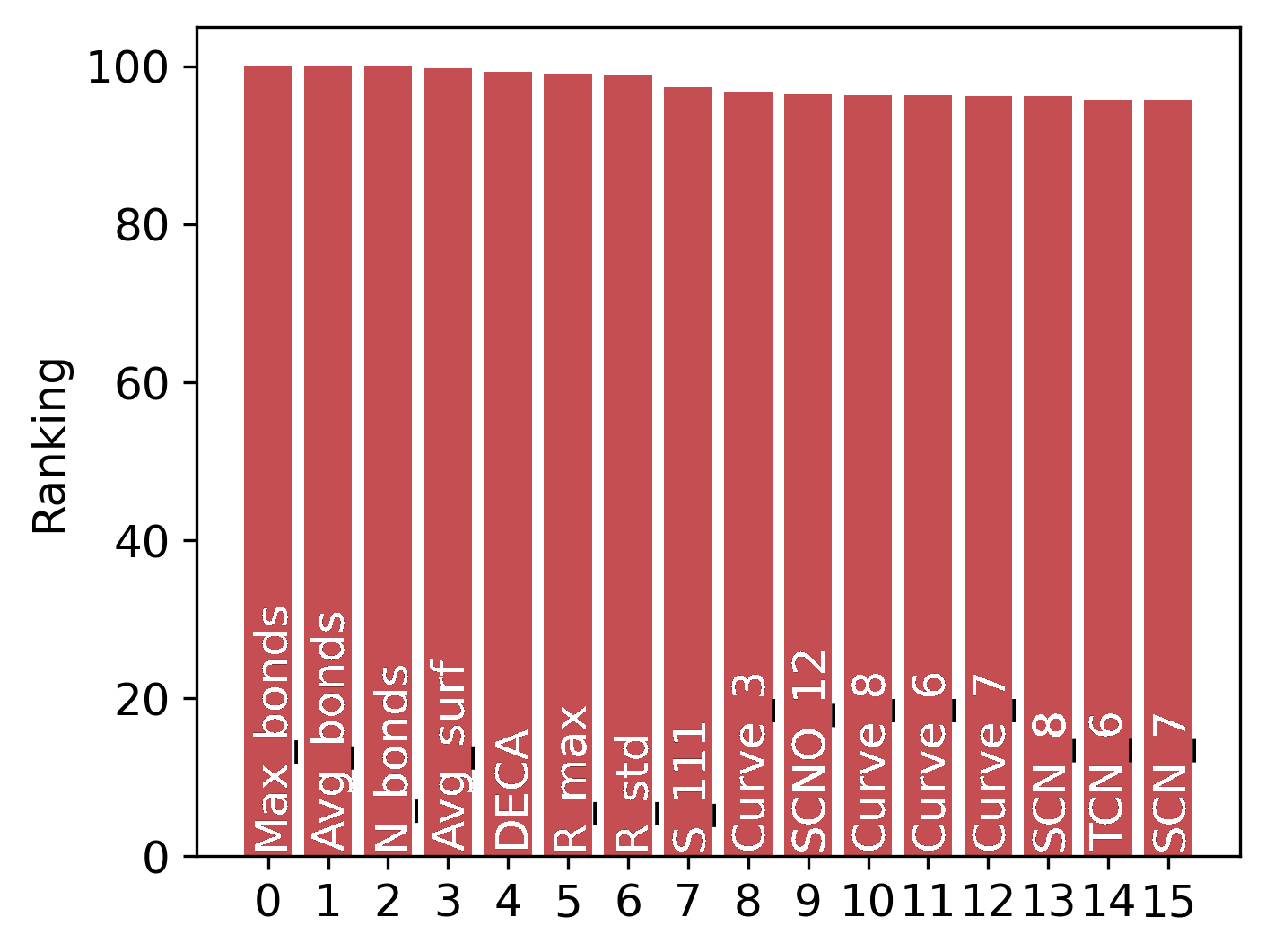}
\includegraphics[width=0.3\textwidth]{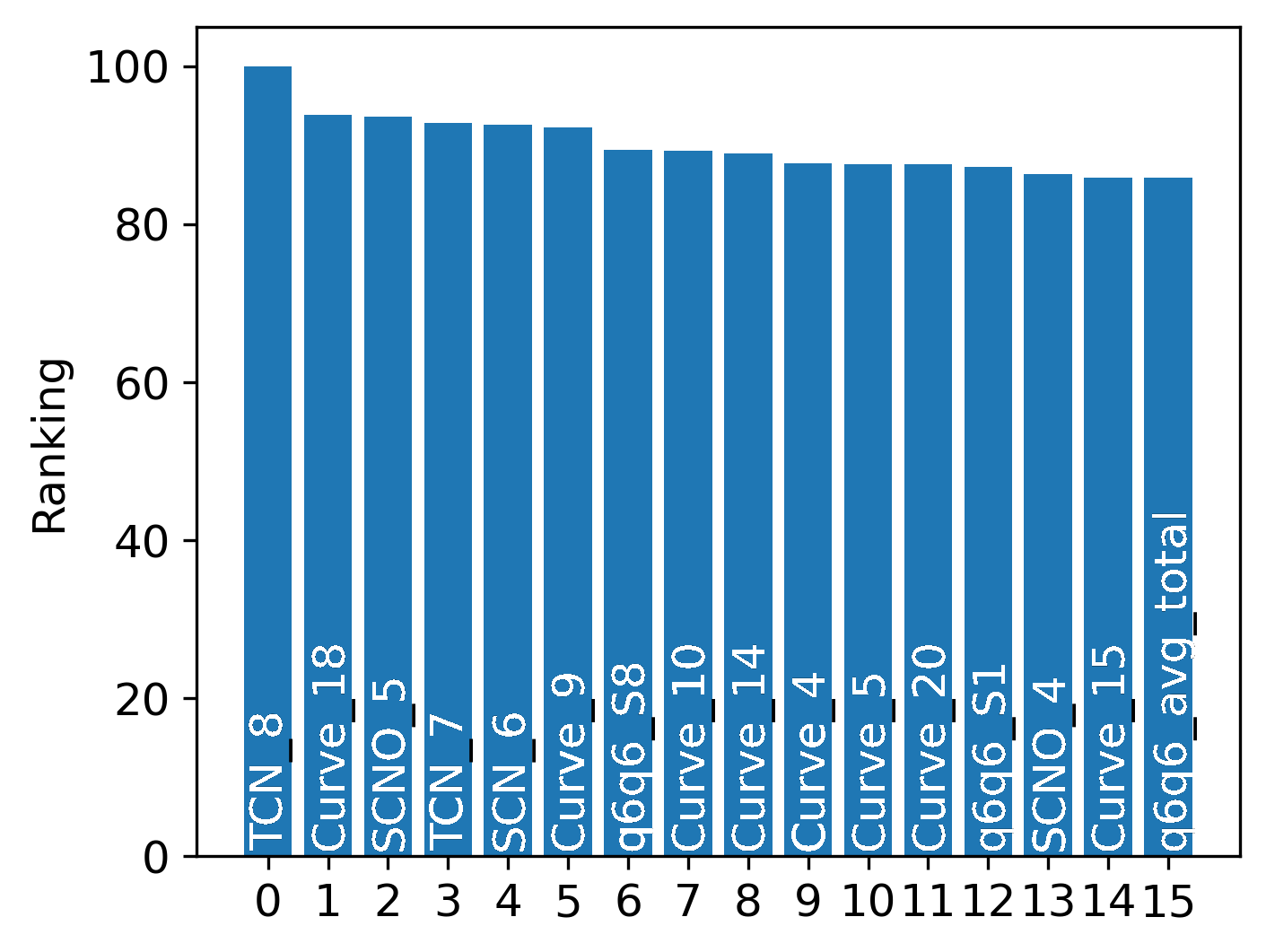}
\includegraphics[width=0.3\textwidth]{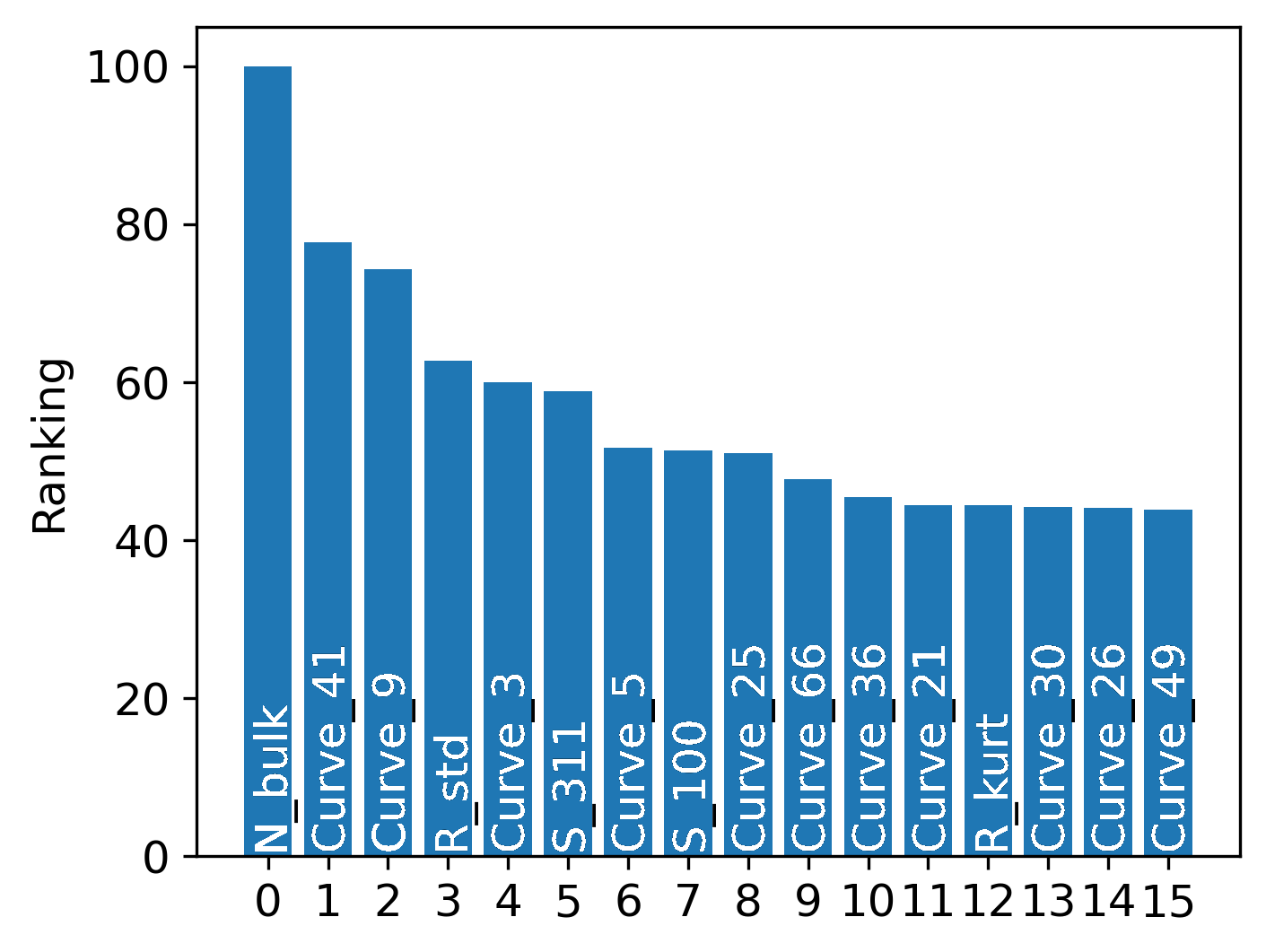}
\caption{Feature unimportance maps (left) from VTF and feature importance maps, including RVTW and CF (middle and right) of Gold feature sets on different feature sets. (a) Bulk, (b) Surface, (c) Total, (d) Condensed, (e) Expanded.}
\label{fig:gold_feature_results}
\end{figure*}

Supporting information for applications, including RNN base model and feature model structures in Fig. \ref{tb:RNN_model_structure}; RNN model training settings in Table \ref{tb:RNN_model}; Gold training settings in Table \ref{tb:RNN_model}; results of 5 models trained on different Gold feature sets in Table \ref{tb:gold_model}; Multilayer perceptron base model and feature model structures in Table \ref{tb:gold_results}, CNN base model and feature model structures in Table \ref{tb:CNN_model_structure}; CNN model training settings in Table \ref{tb:CNN_model}, and feature unimportance maps from different methods in Fig. \ref{fig:gold_feature_results}.

\begin{table*}[tph!]
\centering
\begin{tabular}{lp{2cm}lllp{1cm}p{2cm}}
              & Dataset            & Model            & Optimiser & Epoch & Batch size & Loss \\ \hline
Base model    & Boston house price & Linear regressor & ADAM      & 1000  & 10         & MSE  \\
Feature model & Boston house price & Linear regressor & ADAM      & -     & 10         & MSE  \\
Base model  & Breast Cancer Wisconsin Dataset  & Logistic classifier  & ADAM & 1000 & 10 & binary crossentropy \\
Feature model & Breast Cancer Wisconsin Dataset  & Logistic classifier  & ADAM & - & 10 & binary crossentropy \\ \hline
 \end{tabular}
 \caption{Linear regressor and logistic regressor training settings (seed is 3).}
  \label{tb:linear_model}
\end{table*}

\begin{table*}[pht!]
\centering
\begin{tabular}{ccccp{1.5cm}p{2cm}c}
&  & Layer & Layer type  & Layer parameters & Num of trainable parameters & Output size \\ \hline
  & Base model & L0 & Input layer  & - & 14 & N, 13  \\
& & L1 & Dense layer  & - & 14 & 13, 13 \\
& & L2 & Output layer & - & 14 & None, 1  \\
  & Feature model & L0    & Input layer & -  & 0  & N, 13       \\
& & L1 & Mask layer   & - & 13 & 13, 13 \\
& & L2 & Dense layer  & - & 13 & 13, 13 \\
& & L3 & Output layer & - & 13 & None, 1 \\ \hline
 \end{tabular}
 \caption{Linear base model and feature model structures.}
  \label{tb:linear_model_structure}
\end{table*}

\begin{table*}[]
\centering
\begin{tabular}{ccccp{1.5cm}p{2cm}c}
&  & Layer & Layer type  & Layer parameters & Num of trainable parameters & Output size \\ \hline
& Base model & L0 & Input layer  & -       & 31 & N, 30  \\
& & L1 & Dense layer  & sigmoid & 31 & 30, 30 \\
& & L2 & Output layer & -       & 31 & 30, 1  \\
&Feature model & L0    & Input layer & -                & 30                          & N, 30       \\
& & L1 & Mask layer   & -       & 30 & 30, 30 \\
& & L2 & Dense layer  & sigmoid & 30 & 30, 30 \\
& & L3 & Output layer & -       & 30 & 30, 1 \\ \hline
\end{tabular}
\caption{Logistic base model and feature model structures.}
\label{tb:logistic_model_structure}
\end{table*}

\begin{table*}[]
\centering
\begin{tabular}{llllll}
Dataset            & Model            & Optimiser & Epoch & Batch size & Loss \\ \hline
Boston house price & ANN & ADAM      & 150  & 100         & MSE  \\
Breast Cancer Wisconsin Dataset  & ANN  & ADAM & 150 & 100 & binary\_crossentropy \\
MNIST & CNN & ADAM      & 150  & 100         & binary\_crossentropy  \\ \hline
\end{tabular}
\caption{Independent model training settings (seed is 3)}
\label{tb:inde_model}
\end{table*}

\begin{table*}[]
\centering
\begin{tabular}{p{1.5cm}lllp{1.5cm}l}
Base Model structure & Layer & Layer type & Layer parameters & Num of trainable parameters & Output size \\ \hline
 & L0 & Embedding layer & - & 64000 & None, 200, 32 \\
 & L1 & LSTM layer & - & 53200 & None, 100 \\
 & L2 & Dense layer & sigmoid & 101 & None, 1 \\ \hline
Feature Model structure & L0 & Onehot layer & - & 0 & None, 200, 2000 \\
 & L1 & Mask layer & - & 2,000 & None, 200, 32 \\
 & L2 & LSTM layer & - & 53200 & None, 100 \\
 & L3 & Dense layer & sigmoid & 101 & None, 1 \\ \hline
\end{tabular}
\caption{RNN base model and feature model structures.}
\label{tb:RNN_model_structure}
\end{table*}

\begin{table*}[]
\centering
\begin{tabular}{p{1.5cm}llp{1.5cm}p{1.5cm}p{1.5cm}p{1.5cm}p{2cm}}
Base Model structure & Layer & Layer type & Layer parameters & Pool size & Kernel size & Num of trainable parameters & Output size \\ \hline
                        & L0 & Input layer         & -       & -   & -   & -     & None, 28, 28, 1  \\
                        & L1 & Convolutional layer & Relu    & -   & 3x3 & 640   & None, 26, 26, 64 \\
                        & L2 & Max pooling layer   & -       & 2x2 & -   & 0     & None, 13, 13, 64 \\
                        & L3 & Convolutional layer & Relu    & -   & 3x3 & 18464 & None, 11, 11, 32 \\
                        & L4 & Max pooling layer   & -       & 2x2 & -   & 0     & None, 5, 5, 32   \\
                        & L5 & Flatten layer       & -       & -   & -   & -     & None, 800        \\
                        & L6 & Drop out layer      & 0.5     & -   & -   & 0     & None, 800        \\
                        & L7 & Dense layer         & Softmax & -   & -   & 1602  & None, 2          \\ \hline

Feature Model structure & L0 & Input layer         & -       & -   & -   & -     & None, 28, 28, 1  \\
                        & L1 & Mask layer          & -       & -   & -   & 784   & None, 28, 28, 1  \\
                        & L2 & Convolutional layer & Relu    & -   & 3x3 & 0     & None, 26, 26, 64 \\
                        & L3 & Max pooling layer   & -       & 2x2 & -   & 0     & None, 13, 13, 64 \\
                        & L4 & Convolutional layer & Relu    & -   & 3x3 & 0     & None, 11, 11, 32 \\
                        & L5 & Max pooling layer   & -       & 2x2 & -   & 0     & None, 5, 5, 32   \\
                        & L6 & Flatten layer       & -       & -   & -   & -     & None, 800        \\
                        & L7 & Drop out layer      & 0.5     & -   & -   & 0     & None, 800        \\
                        & L8 & Dense layer         & Softmax & -   & -   & 0     & None, 2         \\ \hline
\end{tabular}
\caption{CNN base model and feature model structures.}
\label{tb:CNN_model_structure}
\end{table*}

\begin{table*}[]
\centering
\begin{tabular}{lllllll}
 & Dataset  & Model & Optimiser  & Epoch  & Batch size  & Loss  \\ \hline
Base model  & IMDB & RNN & ADAM & 500 & 64 & binary\_crossentropy \\
Feature model & IMDB & RNN & ADAM & - & 64 & binary\_crossentropy \\ \hline
\end{tabular}
\caption{RNN model training settings.}
\label{tb:RNN_model}
\end{table*}

\begin{table*}[]
\begin{center}
\begin{tabular}{p{2cm}cccp{2cm}p{2.5cm}c}
Base Model structure & Layer & Layer type & Layer parameters & Num of trainable parameters & Output size \\ \hline
 & L0 & Input layer & - & - & N, 10 \\
 & L1 & Dense layer & - & 1728 & 10, 32 \\
 & L2 & Dense layer & - & 528 & 10, 16 \\
 & L3 & Output layer & - & 17 & 10, 1 \\
Feature Model structure & L0 & Input layer & - & 0 & N, 53 \\
 & L1 & Mask layer & - & 53 & N, 53 \\
 & L3 & Output layer & - & 53 & 10, 1 \\ \hline
\end{tabular}
\end{center}
\caption{Multilayer perceptron base model and feature model structures.}
\label{tb:mpl_model_structure}
\end{table*}

\begin{table*}[]
\centering
\begin{tabular}{lllllll}
 & Dataset  & Model & Optimiser  & Epoch  & Batch size  & Loss  \\ \hline
Base model  & MNIST & CNN & ADAM & 1000 & 128 & binary\_crossentropy \\
Feature model & MNIST & CNN & ADAM & - & 128 & binary\_crossentropy \\ \hline
\end{tabular}
\caption{CNN model training settings}
\label{tb:CNN_model}
\end{table*}

\begin{table*}[]
\centering
\begin{tabular}{lllllll}
              & Dataset            & Model            & Optimiser & Epoch & Batch size & Loss \\ \hline
Base model    & Gold nanoparticles & Linear regressor & ADAM      & 1000  & 10         & Mse  \\
Feature model & Gold nanoparticles & Linear regressor & ADAM      & -     & 10         & Mse \\ \hline
\end{tabular}
\caption{Gold training settings.}
\label{tb:gold_model}
\end{table*}

\begin{table*}[]
\centering
\begin{tabular}{lllllll}
              & Dataset            & Model            & Optimiser & Epoch & Batch size & Loss \\ \hline
Base model    & $IGC_{50}$ & Smiles2vec & ADAM      & 200  & 128         & Mse  \\
Feature model & $IGC_{50}$ & Smiles2vec & ADAM      & -     & 128         & Mse \\ \hline
\end{tabular}
\caption{$IGC_{50}$ training settings.}
\label{tb:igc_model}
\end{table*}

